%% file: main.tex
\documentclass[10pt,twocolumn,letterpaper]{article}

\usepackage{cvpr}              % To produce the CAMERA-READY version
\usepackage{tikz}
\usepackage{algorithm}
\usepackage{algorithmic}
\usepackage{wrapfig}


\newcommand{\rulesep}{\unskip\ \vrule\ }

\definecolor{cvprblue}{rgb}{0.21,0.49,0.74}
\usepackage[pagebackref,breaklinks,colorlinks,citecolor=cvprblue]{hyperref}

\def\paperID{15956} % *** Enter the Paper ID here
\def\confName{CVPR}
\def\confYear{2024}

\title{Federated Unsupervised Visual Representation Learning via Exploiting General Content and Personal Style}

\author{Yuewei Yang\\
Duke University\\
\and
Jingwei Sun\\
Duke University\\
\and
Ang Li\\
Duke University\\
\and
Hai Li\\
Duke University\\
\and
Yiran Chen\\
Duke University\\
}

\begin{document}
\maketitle
\input{sec/0_abstract}    
\input{sec/1_intro}
\input{sec/2_relatedwork}
\input{sec/3_fedstyle}
\input{sec/4_experiments}
\input{sec/5_evaluations}
\input{sec/6_conclusions}

{
    \small
    \bibliographystyle{ieeenat_fullname}
    \bibliography{main}
}

\clearpage
\appendix
\onecolumn
\input{sec/X_suppl}

\end{document}

%% file: sec/0_abstract.tex
\begin{abstract}
  Discriminative unsupervised learning methods such as contrastive learning have demonstrated the ability to learn generalized visual representations on centralized data. It is nonetheless challenging to adapt such methods to a distributed system with unlabeled, private, and heterogeneous client data due to user styles and preferences. Federated learning enables multiple clients to collectively learn a global model without provoking any privacy breach between local clients. On the other hand, another direction of federated learning studies personalized methods to address the local heterogeneity. However, work on solving both generalization and personalization without labels in a decentralized setting remains unfamiliar. In this work, we propose a novel method, FedStyle, to learn a more generalized global model by infusing local style information with local content information for contrastive learning, and to learn more personalized local models by inducing local style information for downstream tasks. The style information is extracted by contrasting original local data with strongly augmented local data (Sobel filtered images). Through extensive experiments with linear evaluations in both IID and non-IID settings, we demonstrate that FedStyle outperforms both the generalization baseline methods and personalization baseline methods in a stylized decentralized setting. Through comprehensive ablations, we demonstrate our design of style infusion and stylized personalization improve performance significantly. 
\end{abstract}

%% file: sec/1_intro.tex
\section{Introduction}
Deep learning models are growing exponentially over the demand on endless data. In a centralized setting, all data is accessible to train a deep learning model at a massive scale. However, with the emerging of edge devices, it is more desirable to train a small-sized deep learning model on local training data and aggregate local models to form a global model for a new deployment. \citet{mcmahan2017communication} proposes federated learning that enables multiple local clients to collectively train a global model in a iterative updating process without sharing local training data between clients. By labelling each example generated on a local device, federated learning has demonstrated competitive performances on various computer vision tasks \cite{li2019privacy,liu2020fedvision,long2015fully}. However, manually labeling all data from numerous local devices is increasingly expensive and concerns users with privacy issue. This substantial requirement leaves most of data on local devices unlabeled. Additionally, data collected and accessible locally is very user-centered, which induces high data heterogeneity across devices due to distinctive user style and preference. Figure \ref{fig:objective} depicts the learning process on user-centered unlabeled data in a distributed environment.

\begin{figure}
    \centering
    \resizebox{!}{0.2\textheight}{
    \input{images/objective.tikz}}
    \caption{Federated unsupervised representation learning with stylized user data. Distinctive user style and preference introduce data heterogeneity and lead to deteriorated global generalization and local personalizations.}
    \label{fig:objective}
\end{figure}

% \begin{figure}
%      \centering
%      \begin{subfigure}[b]{0.45\textwidth}
%       \centering
%         \resizebox{!}{0.2\textheight}{
%          \input{images/objective.tikz}}
%          \caption{}
%          \label{fig:objective}
%      \end{subfigure}
%      \quad
%      \begin{subfigure}[b]{0.45\textwidth}
%      \centering
%          \resizebox{!}{0.14\textheight}{
%          \input{images/intuition.tikz}}
%          \caption{}
%          \label{fig:intuition}
%      \end{subfigure}
%      \caption{(a).Federated unsupervised representation learning with stylized user data. Distinctive user style and preference introduce data heterogeneity and lead to deteriorated global generalization and local personalizations.(b).Intuition of style infusion to improve representations via contrastive learning.}
%      \vspace{-5mm}
% \end{figure}

In the absence of labeled training data, Self-Supervised Learning (SSL) has been developed to pretrain a model with unlabeled data and learn visual representations that are well-generalized to downstream tasks. Discriminative and generative self-supervised methods received growing attentions in recent years \cite{chen2020simple,chen2020improved,donahue2019large,gidaris2018unsupervised,jaiswal2020survey,zhang2016colorful,he2022masked}. Contrastive learning has become the most popular discriminative SSL method that either computes InfoNCE \cite{oord2018representation} on positive and negative examples \cite{chen2020simple,chen2020improved} or predicts projected latent of only positive examples \cite{grill2020bootstrap,chen2021exploring,zbontar2021barlow}. The success of the contrastive learning on visual datasets engenders research and study about applying such a method to other formats of data, such as audio data \cite{saeed2021contrastive} and medical data \cite{mohsenvand2020contrastive}. However, the effectiveness of the method in a distributed environment seeks further study.

Applying contrastive learning in a distributed system with stylized local data poses two core challenges. First, without any supervision, learning a generalized global model is more difficult when all local data is diverse in distribution and limited in volume. Second, while securing a generalized global model, customizing personalized local models is also harder when local data is dispersed not only in class distribution but also in style. These two obstacles require completely different treatments to resolve. Generalization involves a strong regularization to align different local models whilst personalization weakens such regularization to promote customized fit on local dispersed data.

\begin{figure}[htbp]
    \centering
    \resizebox{!}{0.2\textwidth}{
    \input{images/intuition.tikz}}
    \caption{Intuition of style infusion to improve representations via contrastive learning.}
    \label{fig:intuition}
\end{figure}

In this work, we address both generalization and personalization of the federated unsupervised representation learning with the utilization of local style information. \citet{von2021self} assumes that an image is a product of a content factor ($c$) and a style factor ($s$). The authors further prove that, under a self-supervised learning framework, image augmentations are applied so that the feature extractor learns content-identifiable representation that is robust to the image distortions (upper part of Figure \ref{fig:intuition}). Based on this argument, our intuition is that a locally extracted image feature is a product of a generalized content feature ($h_c$) and a personalized style feature ($h_s$). During self-supervised learning, the content feature is infused with locally extracted style features so that the learned representation is also robust to this latent distortion (lower part of Figure \ref{fig:intuition}). Since every client has its own style and preference, extracting local style features to strengthen the local generalized content features and facilitate personalized features in a decentralized setting become feasible to improve both generalization and personalization. We propose Federated with Style (\textbf{FedStyle}) to extract local style features via training a local style feature extractor and the local style features serve 2 functions. \textbf{First} local style features are infused with local content features so the local content models are robust to local latent distortion in addition to other image-level distortions applied for contrastive learning \cite{chen2020simple}. With local models robust to local distortions, a generalized global model robust to various distortions can be aggregated. \textbf{Second} the local \textit{personalized} style features are incorporated with local \textit{generalized} content features to promote personalization when deploying on downstream tasks.

We propose \textbf{FedStyle} to improve both \textit{generalization} and \textit{personalization} in a decentralized framework via following contributions:
\begin{itemize}
    \item We extract style representations locally by contrasting local style with Sobel filtered style.
    \item By infusing style and content hidden features in the latent space, we learn more robust local models that aggregate a more generlized global model.
    \item By generating stylized content representation that is better customized to local data, we improve the personalization of local models in the downstream task.
\end{itemize}

With extensive experiments, we evaluate the effectiveness of FedStyle in terms of generalization and personalization under both IID and non-IID data distributions. To simulate user-centered data style, we evaluate FedStyle on three different data types: 1. Digit consisting MNIST \cite{deng2012mnist}, USPS \cite{uspsdataset}, and SVHN \cite{netzer2011reading}; 2. Office Home \cite{venkateswara2017deep} consisting Art, Clipart, Product, and Real World styles; 3. Adaptiope \cite{Ringwald_2021_WACV} consisting Synthetic, Product, Real Life styles. FedStyle delivers a more generalized global model than the FedAvg baseline and achieves higher personalized accruacy than personalized federated algorithms: FedRep \cite{collins2021exploiting}, FedPer \cite{arivazhagan2019federated}, FedProx \cite{li2020federated}, APFL \cite{deng2020adaptive}, and Ditto \cite{li2021ditto}.

%% file: images/objective.tikz
\tikzset{every picture/.style={line width=0.75pt}} %set default line width to 0.75pt        

\begin{tikzpicture}[x=0.75pt,y=0.75pt,yscale=-1,xscale=1]
%uncomment if require: \path (0,784); %set diagram left start at 0, and has height of 784

%Rounded Rect [id:dp6672499526670588] 
\draw  [draw opacity=0][fill={rgb, 255:red, 184; green, 233; blue, 134  }  ,fill opacity=0.2 ] (440,60) .. controls (440,33.49) and (461.49,12) .. (488,12) -- (1022,12) .. controls (1048.51,12) and (1070,33.49) .. (1070,60) -- (1070,204) .. controls (1070,230.51) and (1048.51,252) .. (1022,252) -- (488,252) .. controls (461.49,252) and (440,230.51) .. (440,204) -- cycle ;
%Rounded Rect [id:dp9262589976227917] 
\draw  [draw opacity=0][fill={rgb, 255:red, 74; green, 144; blue, 226 }  ,fill opacity=0.5 ] (650,82) .. controls (650,64.33) and (664.33,50) .. (682,50) -- (1018,50) .. controls (1035.67,50) and (1050,64.33) .. (1050,82) -- (1050,178) .. controls (1050,195.67) and (1035.67,210) .. (1018,210) -- (682,210) .. controls (664.33,210) and (650,195.67) .. (650,178) -- cycle ;
%Rounded Rect [id:dp042926930769702665] 
\draw  [draw opacity=0][fill={rgb, 255:red, 184; green, 233; blue, 134  }  ,fill opacity=0.2 ] (30,289) .. controls (30,253.65) and (58.65,225) .. (94,225) -- (286,225) .. controls (321.35,225) and (350,253.65) .. (350,289) -- (350,481) .. controls (350,516.35) and (321.35,545) .. (286,545) -- (94,545) .. controls (58.65,545) and (30,516.35) .. (30,481) -- cycle ;
%Image [id:dp1935135460602977] 
\draw (194.94,389.95) node [rotate=-269.96] {\includegraphics[width=170pt,height=150pt]{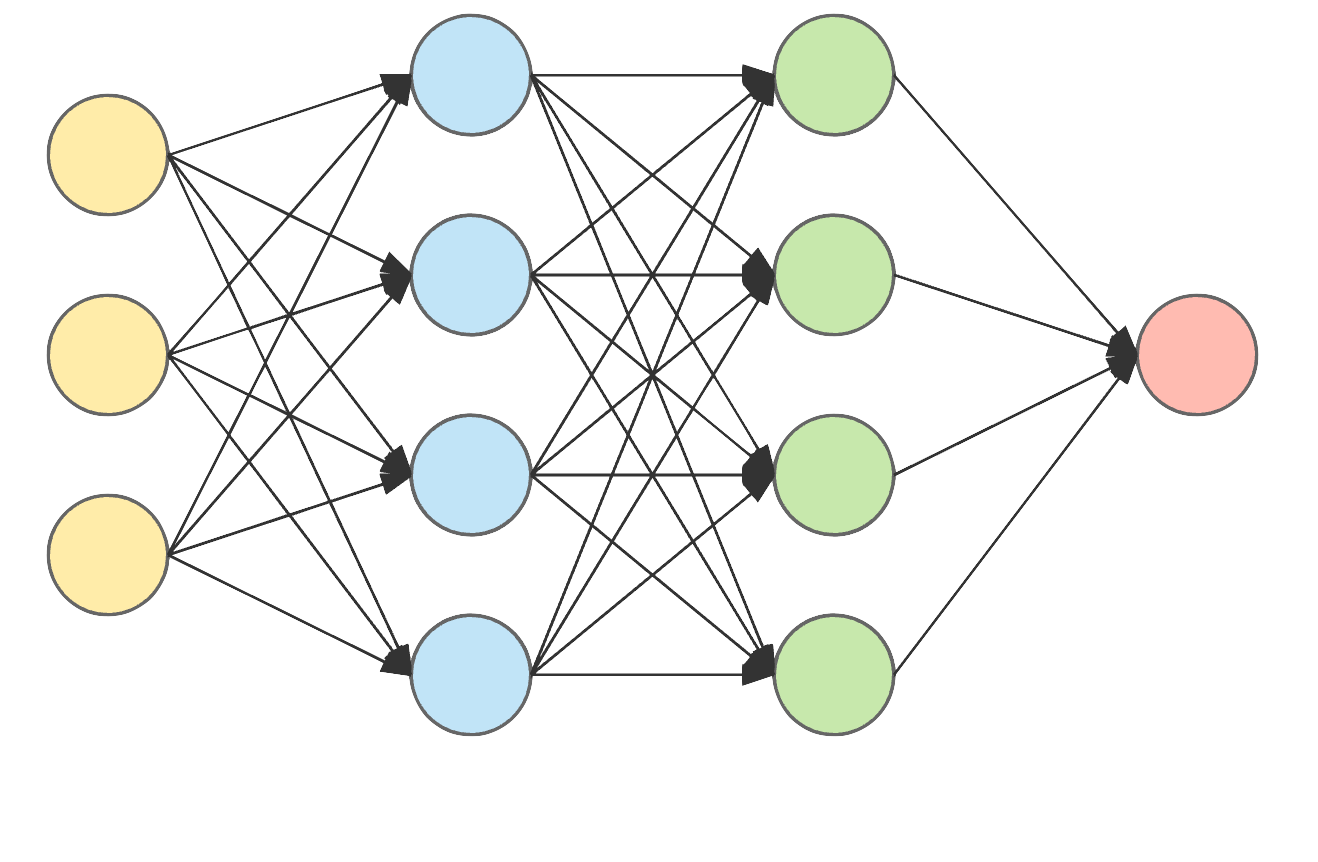}};
%Image [id:dp9619445196782008] 
\draw (539.96,110) node [rotate=-269.96] {\includegraphics[width=90pt,height=75pt]{images/deep_learning_model.png}};
%Image [id:dp7004536742980869] 
\draw (720,130) node  {\includegraphics[width=75pt,height=75pt]{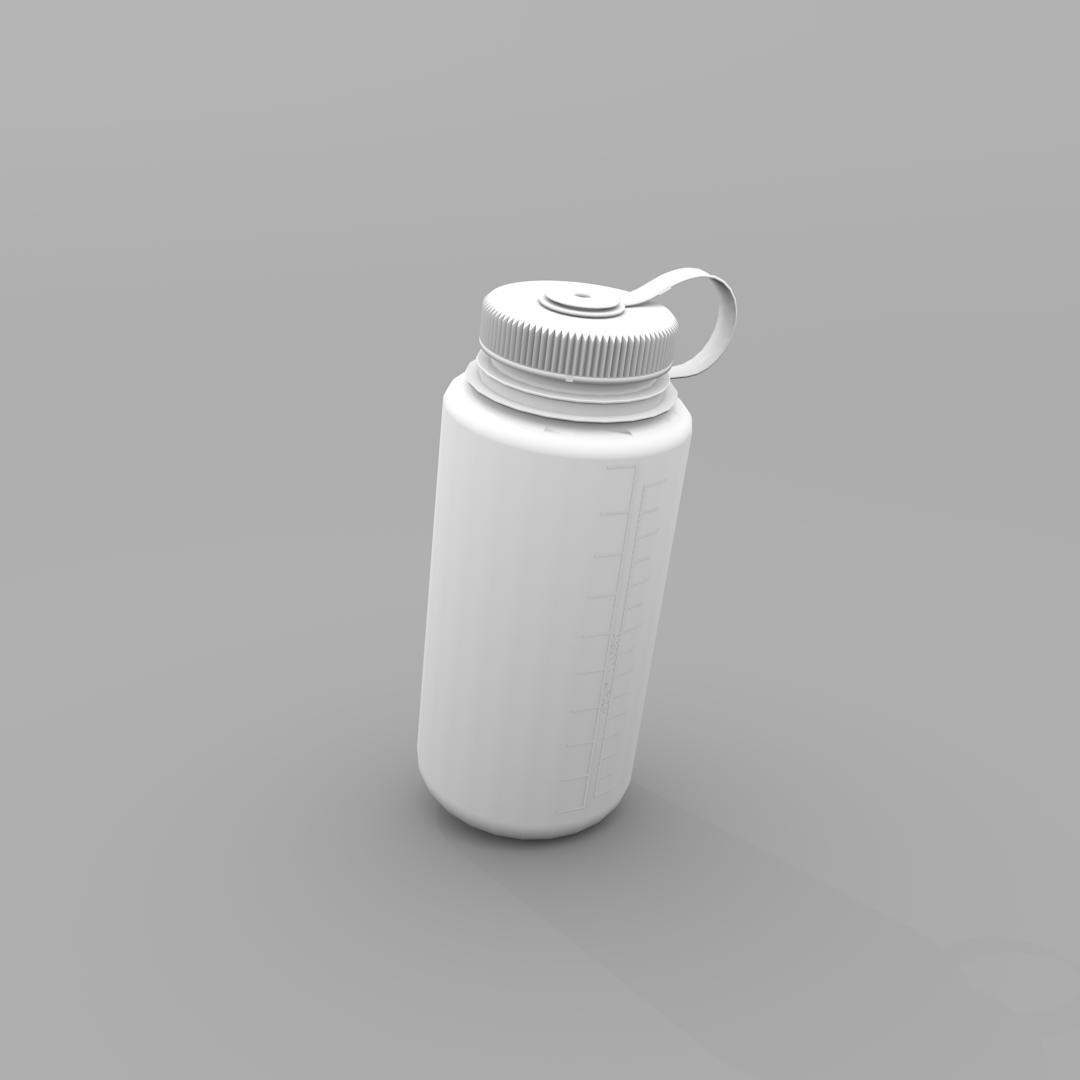}};
%Image [id:dp1251045912834623] 
\draw (850,130) node  {\includegraphics[width=75pt,height=75pt]{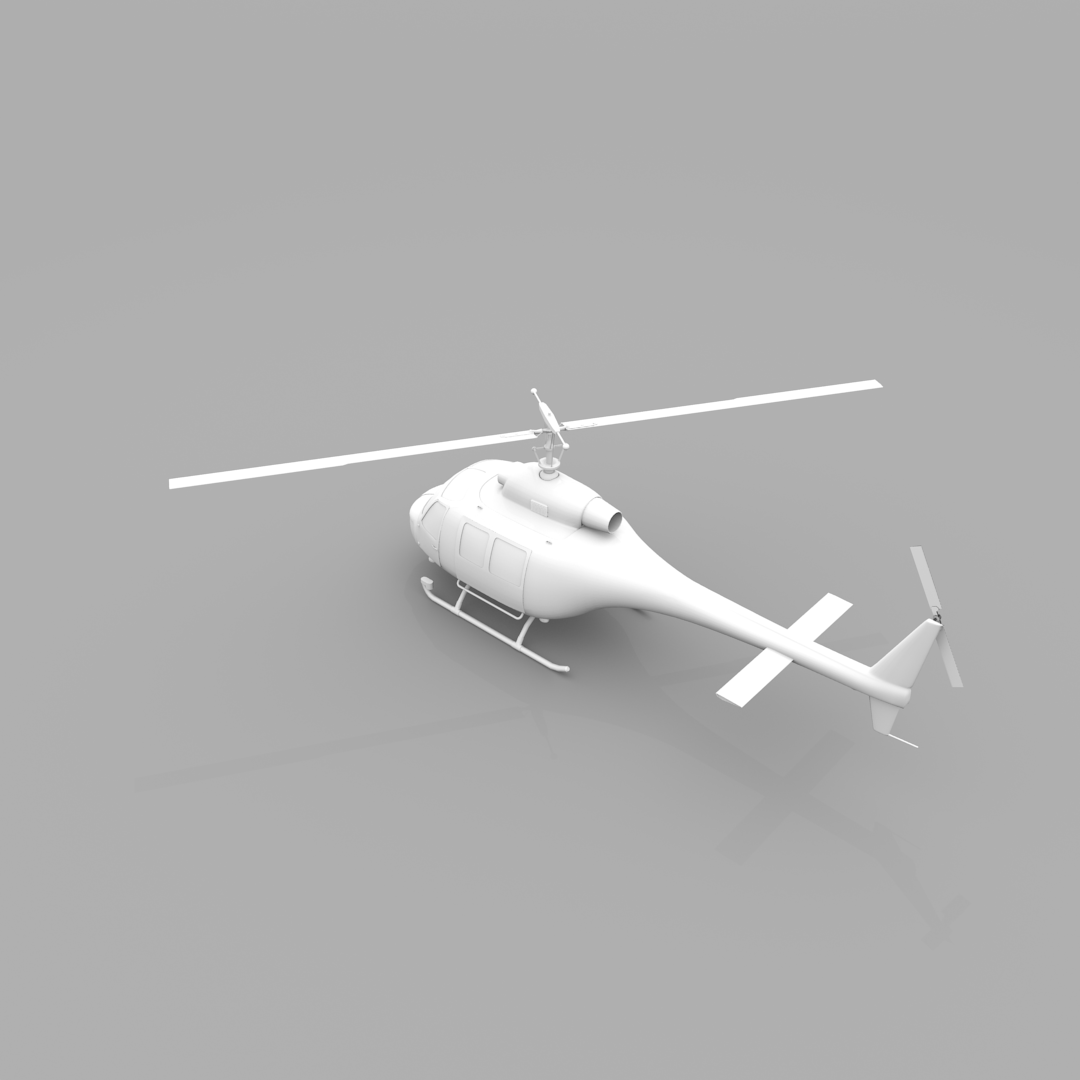}};
%Image [id:dp848506565799177] 
\draw (980,130) node  {\includegraphics[width=75pt,height=75pt]{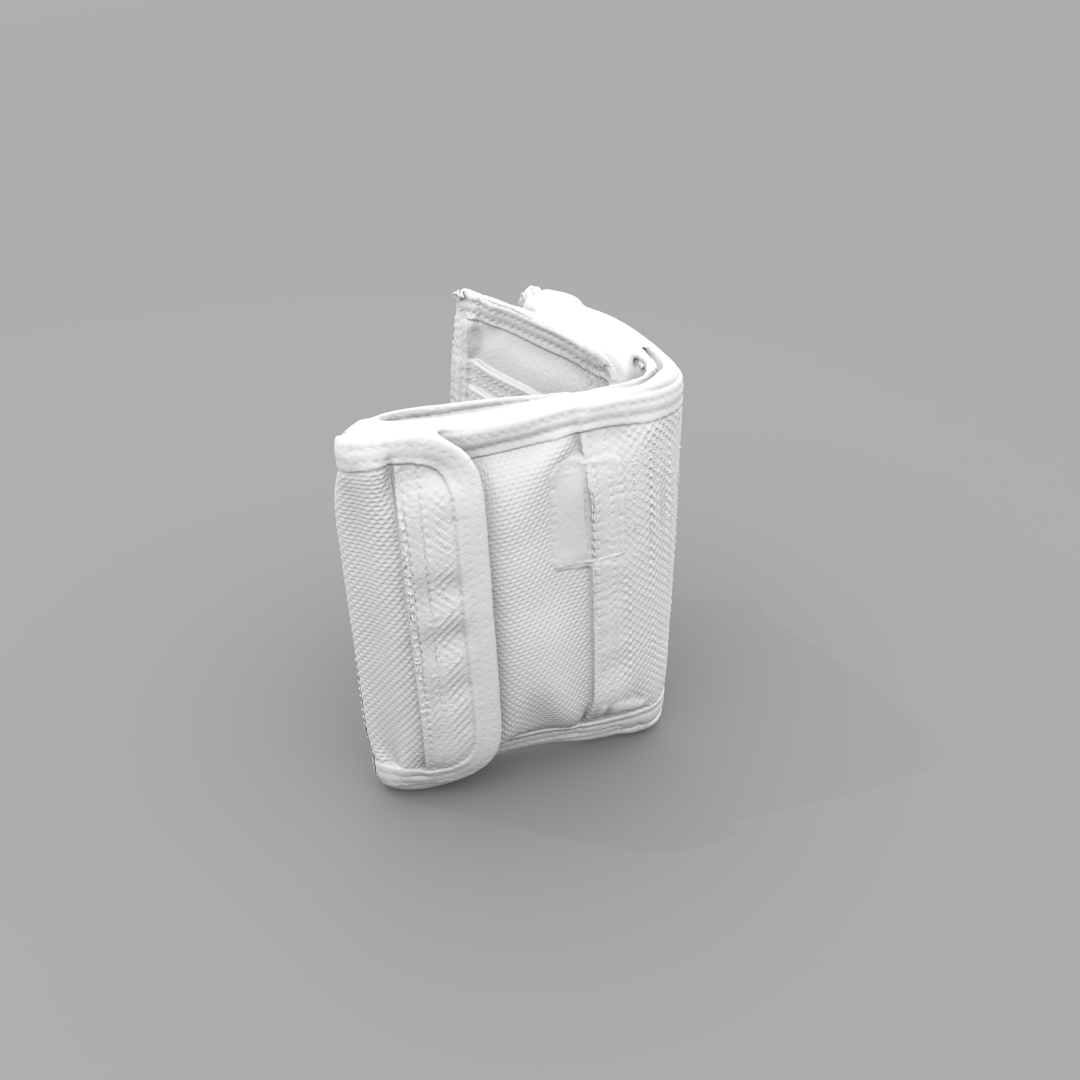}};
%Straight Lines [id:da4339502941546034] 
\draw [line width=2.25]    (650,130) -- (578.5,129.53) ;
\draw [shift={(574.5,129.5)}, rotate = 0.38] [color={rgb, 255:red, 0; green, 0; blue, 0 }  ][line width=2.25]    (27.98,-8.42) .. controls (17.79,-3.57) and (8.47,-0.77) .. (0,0) .. controls (8.47,0.77) and (17.79,3.57) .. (27.98,8.42)   ;
%Rounded Rect [id:dp5440476635566296] 
\draw  [draw opacity=0][fill={rgb, 255:red, 184; green, 233; blue, 134 }  ,fill opacity=0.2 ] (440,318) .. controls (440,291.49) and (461.49,270) .. (488,270) -- (1022,270) .. controls (1048.51,270) and (1070,291.49) .. (1070,318) -- (1070,462) .. controls (1070,488.51) and (1048.51,510) .. (1022,510) -- (488,510) .. controls (461.49,510) and (440,488.51) .. (440,462) -- cycle ;
%Rounded Rect [id:dp9557430652170193] 
\draw  [draw opacity=0][fill={rgb, 255:red, 208; green, 2; blue, 27 }  ,fill opacity=0.5 ] (650,340) .. controls (650,322.33) and (664.33,308) .. (682,308) -- (1018,308) .. controls (1035.67,308) and (1050,322.33) .. (1050,340) -- (1050,436) .. controls (1050,453.67) and (1035.67,468) .. (1018,468) -- (682,468) .. controls (664.33,468) and (650,453.67) .. (650,436) -- cycle ;
%Image [id:dp25263813103445365] 
\draw (539.96,370) node [rotate=-269.96] {\includegraphics[width=90pt,height=75pt]{images/deep_learning_model.png}};
%Image [id:dp7601366595472239] 
\draw (720,388) node  {\includegraphics[width=75pt,height=75pt]{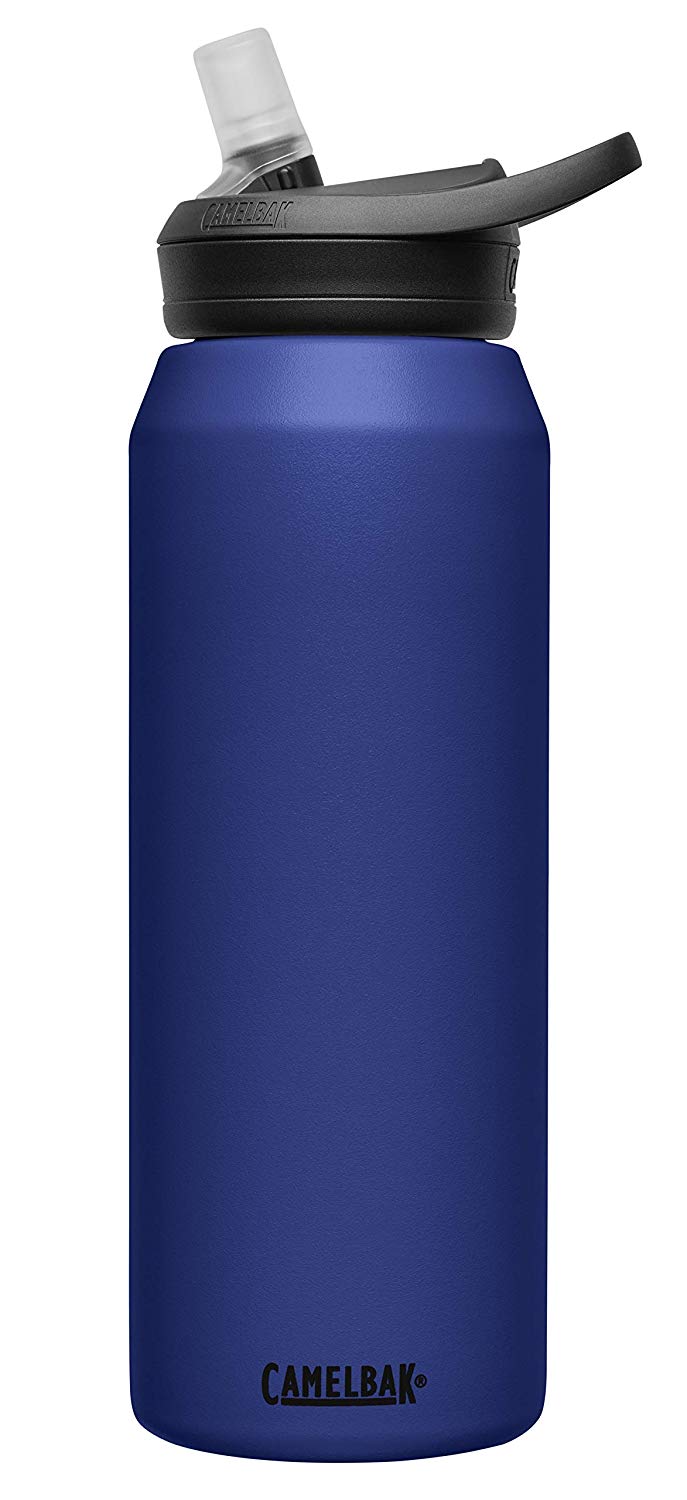}};
%Image [id:dp4522105389931166] 
\draw (850,388) node  {\includegraphics[width=75pt,height=75pt]{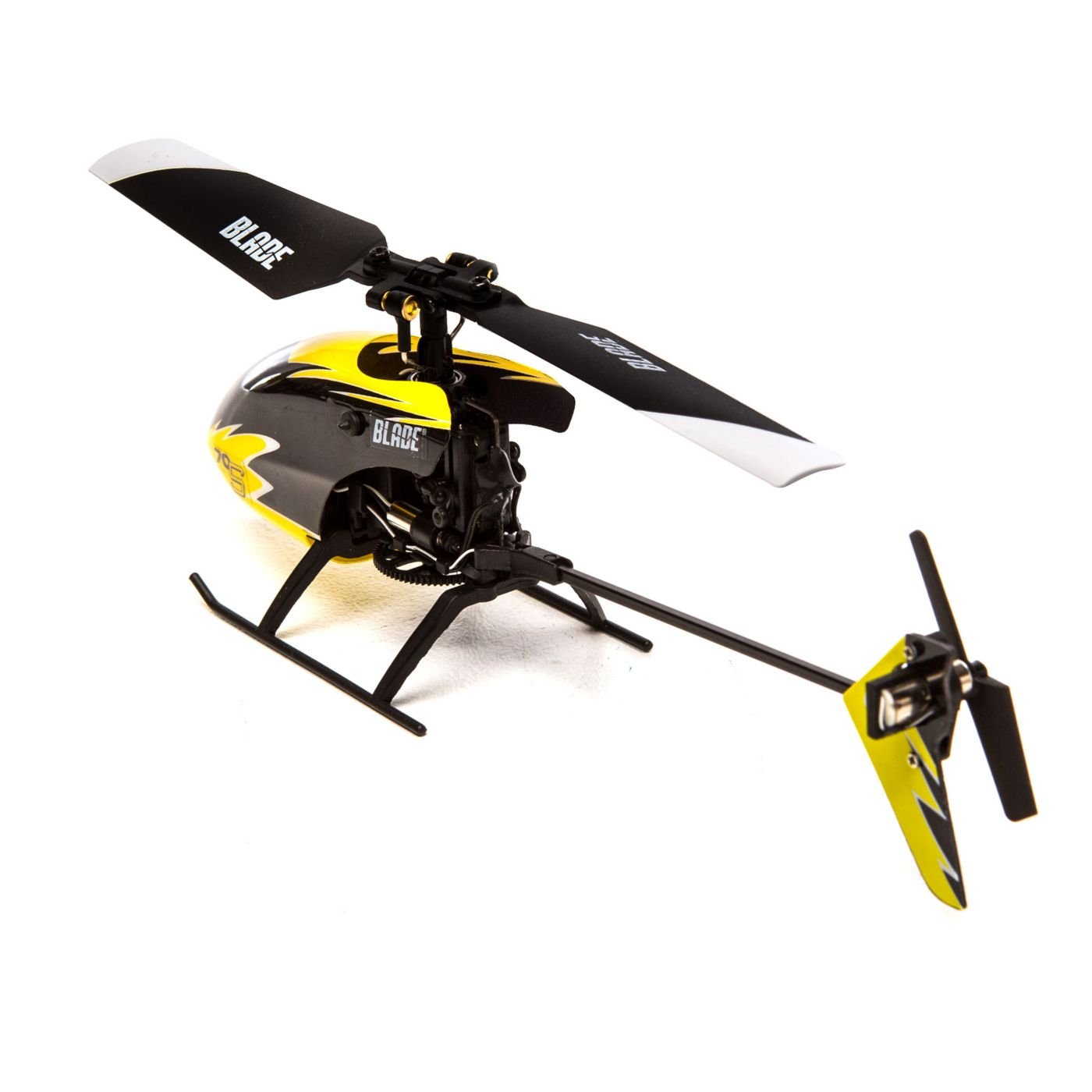}};
%Image [id:dp4543484174868424] 
\draw (980,388) node  {\includegraphics[width=75pt,height=75pt]{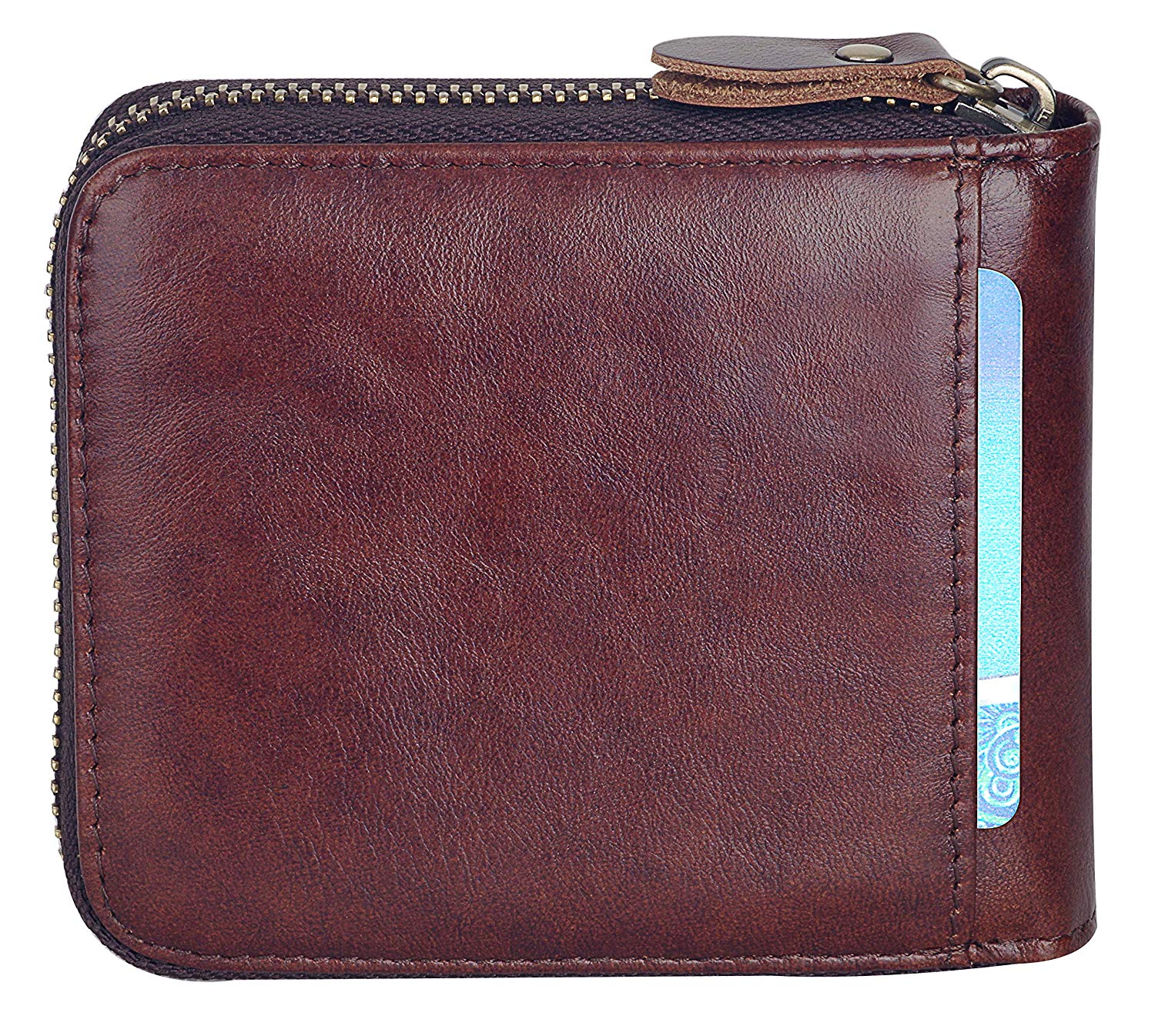}};
%Straight Lines [id:da9077037497399543] 
\draw [line width=2.25]    (650,390) -- (578.5,389.53) ;
\draw [shift={(574.5,389.5)}, rotate = 0.38] [color={rgb, 255:red, 0; green, 0; blue, 0 }  ][line width=2.25]    (27.98,-8.42) .. controls (17.79,-3.57) and (8.47,-0.77) .. (0,0) .. controls (8.47,0.77) and (17.79,3.57) .. (27.98,8.42)   ;
%Rounded Rect [id:dp10569563289145045] 
\draw  [draw opacity=0][fill={rgb, 255:red, 184; green, 233; blue, 134  }  ,fill opacity=0.2 ] (440,578) .. controls (440,551.49) and (461.49,530) .. (488,530) -- (1022,530) .. controls (1048.51,530) and (1070,551.49) .. (1070,578) -- (1070,722) .. controls (1070,748.51) and (1048.51,770) .. (1022,770) -- (488,770) .. controls (461.49,770) and (440,748.51) .. (440,722) -- cycle ;
%Rounded Rect [id:dp7585744936253973] 
\draw  [draw opacity=0][fill={rgb, 255:red, 126; green, 211; blue, 33 }  ,fill opacity=0.5 ] (650,600) .. controls (650,582.33) and (664.33,568) .. (682,568) -- (1018,568) .. controls (1035.67,568) and (1050,582.33) .. (1050,600) -- (1050,696) .. controls (1050,713.67) and (1035.67,728) .. (1018,728) -- (682,728) .. controls (664.33,728) and (650,713.67) .. (650,696) -- cycle ;
%Image [id:dp9799815583379592] 
\draw (539.96,630) node [rotate=-269.96] {\includegraphics[width=90pt,height=75pt]{images/deep_learning_model.png}};
%Image [id:dp7963269888676607] 
\draw (720,648) node  {\includegraphics[width=75pt,height=75pt]{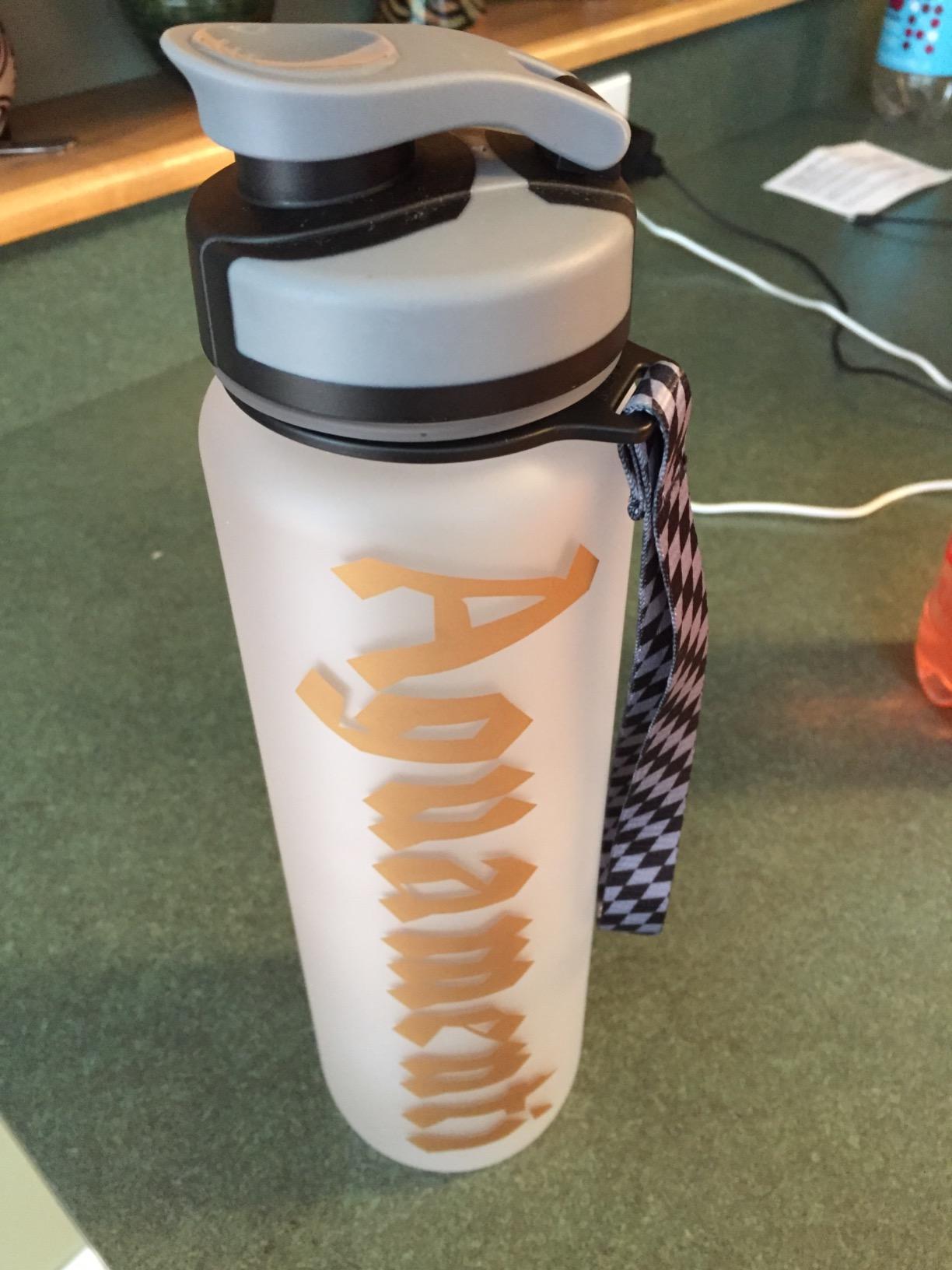}};
%Image [id:dp9637702440481526] 
\draw (850,648) node  {\includegraphics[width=75pt,height=75pt]{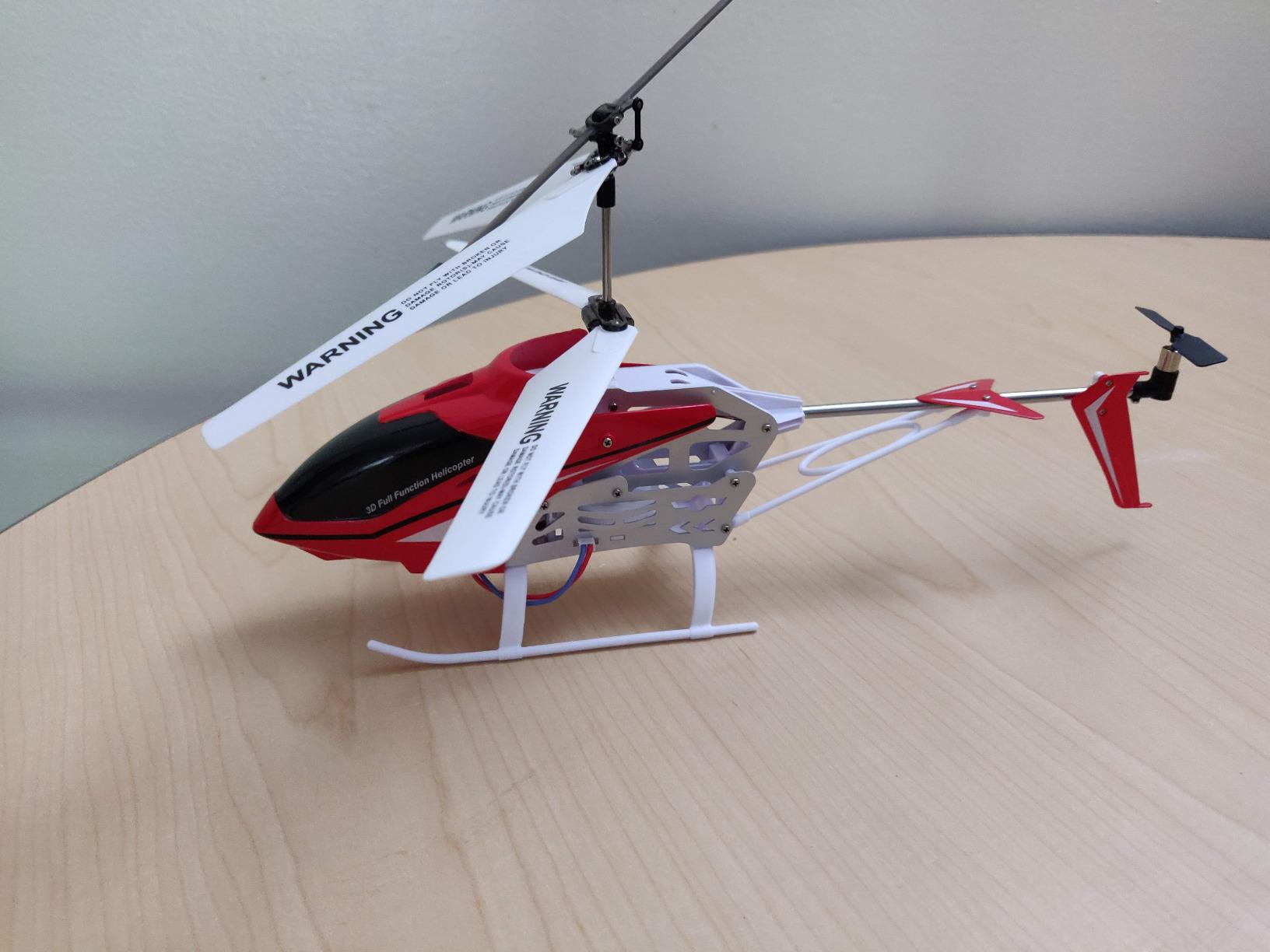}};
%Image [id:dp44368282563047834] 
\draw (980,648) node  {\includegraphics[width=75pt,height=75pt]{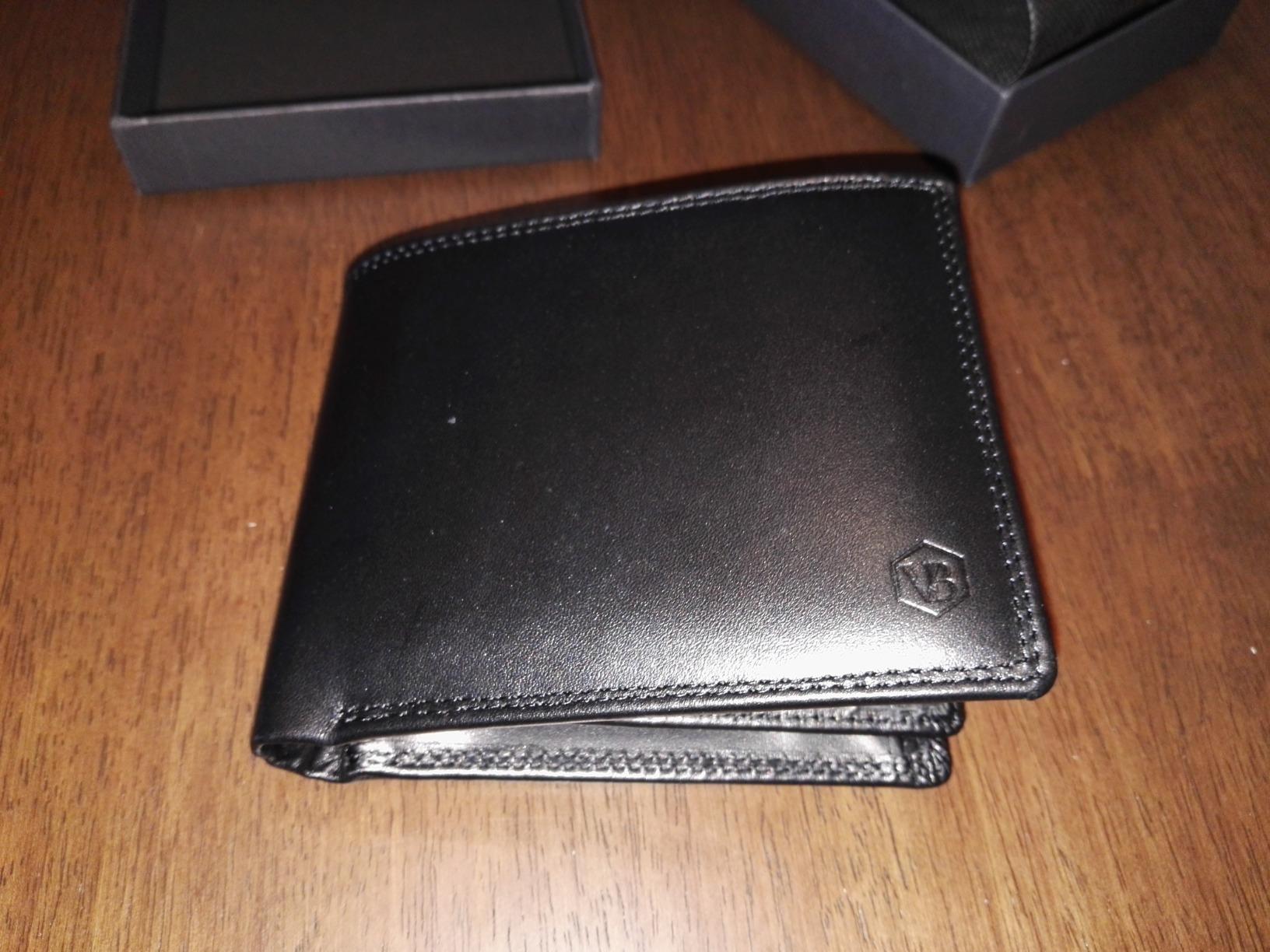}};
%Straight Lines [id:da009867849604060286] 
\draw [line width=2.25]    (650,650) -- (578.5,649.53) ;
\draw [shift={(574.5,649.5)}, rotate = 0.38] [color={rgb, 255:red, 0; green, 0; blue, 0 }  ][line width=2.25]    (27.98,-8.42) .. controls (17.79,-3.57) and (8.47,-0.77) .. (0,0) .. controls (8.47,0.77) and (17.79,3.57) .. (27.98,8.42)   ;
%Curve Lines [id:da26172464480421564] 
\draw [color={rgb, 255:red, 0; green, 180; blue, 0 }  ,draw opacity=1 ][line width=2.25]    (490,130) .. controls (463.22,128.41) and (348.82,136.98) .. (269.3,162.84) .. controls (226.82,176.66) and (194.29,195.41) .. (190.39,220.18) ;
\draw [shift={(190,225)}, rotate = 270.38] [fill={rgb, 255:red, 0; green, 180; blue, 0 }  ,fill opacity=1 ][line width=0.08]  [draw opacity=0] (14.29,-6.86) -- (0,0) -- (14.29,6.86) -- cycle    ;
%Curve Lines [id:da8811816192539428] 
\draw [color={rgb, 255:red, 0; green, 180; blue, 0 }  ,draw opacity=1 ][line width=2.25]    (490,650) .. controls (448.84,647.55) and (200.39,621.41) .. (190.32,549.46) ;
\draw [shift={(190,545)}, rotate = 89.87] [fill={rgb, 255:red, 0; green, 180; blue, 0 }  ,fill opacity=1 ][line width=0.08]  [draw opacity=0] (14.29,-6.86) -- (0,0) -- (14.29,6.86) -- cycle    ;
%Straight Lines [id:da24014269749579098] 
\draw [color={rgb, 255:red, 0; green, 180; blue, 0 }  ,draw opacity=1 ][line width=2.25]    (490,390) -- (354,390) ;
\draw [shift={(350,390)}, rotate = 360] [color={rgb, 255:red, 0; green, 180; blue, 0 }  ,draw opacity=1 ][line width=2.25]    (17.49,-5.26) .. controls (11.12,-2.23) and (5.29,-0.48) .. (0,0) .. controls (5.29,0.48) and (11.12,2.23) .. (17.49,5.26)   ;
%Curve Lines [id:da6591855339486958] 
\draw [color={rgb, 255:red, 0; green, 0; blue, 180 }  ,draw opacity=1 ][line width=2.25]    (160,225) .. controls (160.65,111.97) and (419.97,109.07) .. (486.19,109.94) ;
\draw [shift={(490,110)}, rotate = 180.97] [color={rgb, 255:red, 0; green, 0; blue, 180 }  ,draw opacity=1 ][line width=2.25]    (17.49,-5.26) .. controls (11.12,-2.23) and (5.29,-0.48) .. (0,0) .. controls (5.29,0.48) and (11.12,2.23) .. (17.49,5.26)   ;
%Curve Lines [id:da5181147674018867] 
\draw [color={rgb, 255:red, 0; green, 0; blue, 180 }  ,draw opacity=1 ][line width=2.25]    (160,550) .. controls (160.65,669.89) and (419.33,669.08) .. (485.53,669.94) ;
\draw [shift={(489.33,670)}, rotate = 180.97] [color={rgb, 255:red, 0; green, 0; blue, 180 }  ,draw opacity=1 ][line width=2.25]    (17.49,-5.26) .. controls (11.12,-2.23) and (5.29,-0.48) .. (0,0) .. controls (5.29,0.48) and (11.12,2.23) .. (17.49,5.26)   ;
%Straight Lines [id:da2182969187703967] 
\draw [color={rgb, 255:red, 0; green, 0; blue, 180 }  ,draw opacity=1 ][fill={rgb, 255:red, 0; green, 0; blue, 180 }  ,fill opacity=1 ][line width=2.25]    (348.67,360) -- (382,360) -- (420,360) -- (480,360) ;
\draw [shift={(490,360)}, rotate = 180] [color={rgb, 255:red, 0; green, 0; blue, 180 }  ,draw opacity=1 ][line width=2.25]    (17.49,-5.26) .. controls (11.12,-2.23) and (5.29,-0.48) .. (0,0) .. controls (5.29,0.48) and (11.12,2.23) .. (17.49,5.26)   ;
%Straight Lines [id:da622054320878553] 
\draw [color={rgb, 255:red, 0; green, 0; blue, 180 }  ,draw opacity=1 ][fill={rgb, 255:red, 0; green, 0; blue, 180 }  ,fill opacity=1 ][line width=2.25]    (250,25) -- (320,25) ;
\draw [shift={(320,25)}, rotate = 180] [color={rgb, 255:red, 0; green, 0; blue, 180 }  ,draw opacity=1 ][line width=2.25]    (17.49,-5.26) .. controls (11.12,-2.23) and (5.29,-0.48) .. (0,0) .. controls (5.29,0.48) and (11.12,2.23) .. (17.49,5.26)   ;
%Straight Lines [id:da4163394045063835] 
\draw [color={rgb, 255:red, 0; green, 180; blue, 0 }  ,draw opacity=1 ][line width=2.25]    (320,75) -- (390,75) ;
\draw [shift={(390,75)}, rotate = 180] [color={rgb, 255:red, 0; green, 180; blue, 0 }  ,draw opacity=1 ][line width=2.25]    (17.49,-5.26) .. controls (11.12,-2.23) and (5.29,-0.48) .. (0,0) .. controls (5.29,0.48) and (11.12,2.23) .. (17.49,5.26)   ;

% Text Node
\draw (90,505) node [anchor=north west][inner sep=0.75pt]   [align=left] {{\Huge Global Network}};
% Text Node
\draw (120,240) node [anchor=north west][inner sep=0.75pt]   [align=left] {\textbf{{\fontsize{30}{40}\selectfont Server}}};
% Text Node
\draw (465,180) node [anchor=north west][inner sep=0.75pt]   [align=left] {{\Huge Local Network}};
% Text Node
\draw (765,180) node [anchor=north west][inner sep=0.75pt]   [align=left] {{\Huge \textbf{Unlabled Data}}};
% Text Node
\draw (805,55) node [anchor=north west][inner sep=0.75pt]   [align=left] {\textbf{{\Huge \textcolor[rgb]{0.29,0.56,0.89}{Style A}}}};
% Text Node
\draw (700,15) node [anchor=north west][inner sep=0.75pt]   [align=left] {{\fontsize{30}{40}\selectfont \textbf{Client A}}};
% Text Node
\draw (465,440) node [anchor=north west][inner sep=0.75pt]   [align=left] {{\Huge Local Network}};
% Text Node
\draw (765,440) node [anchor=north west][inner sep=0.75pt]   [align=left] {{\Huge \textbf{Unlabled Data}}};
% Text Node
\draw (805,310) node [anchor=north west][inner sep=0.75pt]   [align=left] {\textbf{{\Huge \textcolor[rgb]{0.82,0.01,0.11}{Style B}}}};
% Text Node
\draw (700,275) node [anchor=north west][inner sep=0.75pt]   [align=left] {{\fontsize{30}{40}\selectfont \textbf{Client B}}};
% Text Node
\draw (465,700) node [anchor=north west][inner sep=0.75pt]   [align=left] {{\Huge Local Network}};
% Text Node
\draw (765,700) node [anchor=north west][inner sep=0.75pt]   [align=left] {{\Huge \textbf{Unlabled Data}}};
% Text Node
\draw (805,570) node [anchor=north west][inner sep=0.75pt]   [align=left] {\textbf{{\Huge \textcolor[rgb]{0.49,0.83,0.13}{Style C}}}};
% Text Node
\draw (700,535) node [anchor=north west][inner sep=0.75pt]   [align=left] {{\fontsize{30}{40}\selectfont \textbf{Client C}}};
% Text Node
\draw (5,10) node [anchor=north west][inner sep=0.75pt]   [align=left] {\textcolor[rgb]{0,0,0.71}{\textbf{{\fontsize{30}{40}\selectfont Local Update}}}};
% Text Node
\draw (5,55) node [anchor=north west][inner sep=0.75pt]   [align=left] {{\fontsize{30}{40}\selectfont \textbf{\textcolor[rgb]{0,0.71,0}{Global Aggregate}}}};

\end{tikzpicture}

%% file: images/intuition.tikz
\tikzset{every picture/.style={line width=0.75pt}} %set default line width to 0.75pt        

\begin{tikzpicture}[x=0.75pt,y=0.75pt,yscale=-1,xscale=1]
%uncomment if require: \path (0,371); %set diagram left start at 0, and has height of 371

%Straight Lines [id:da259914181405573] 
\draw [color={rgb, 255:red, 74; green, 144; blue, 226 }  ,draw opacity=1 ][line width=2.25]    (240,120) -- (296,120) ;
\draw [shift={(300,120)}, rotate = 180] [color={rgb, 255:red, 74; green, 144; blue, 226 }  ,draw opacity=1 ][line width=2.25]    (17.49,-5.26) .. controls (11.12,-2.23) and (5.29,-0.48) .. (0,0) .. controls (5.29,0.48) and (11.12,2.23) .. (17.49,5.26)   ;
%Straight Lines [id:da4619063571863695] 
\draw [color={rgb, 255:red, 74; green, 144; blue, 226 }  ,draw opacity=1 ][line width=2.25]    (240,50) -- (296,50) ;
\draw [shift={(300,50)}, rotate = 180] [color={rgb, 255:red, 74; green, 144; blue, 226 }  ,draw opacity=1 ][line width=2.25]    (17.49,-5.26) .. controls (11.12,-2.23) and (5.29,-0.48) .. (0,0) .. controls (5.29,0.48) and (11.12,2.23) .. (17.49,5.26)   ;
%Straight Lines [id:da9316614258695564] 
\draw [color={rgb, 255:red, 74; green, 144; blue, 226 }  ,draw opacity=1 ][line width=2.25]    (335,50) -- (391,50) ;
\draw [shift={(395,50)}, rotate = 180] [color={rgb, 255:red, 74; green, 144; blue, 226 }  ,draw opacity=1 ][line width=2.25]    (17.49,-5.26) .. controls (11.12,-2.23) and (5.29,-0.48) .. (0,0) .. controls (5.29,0.48) and (11.12,2.23) .. (17.49,5.26)   ;
%Straight Lines [id:da7452893543593335] 
\draw [color={rgb, 255:red, 74; green, 144; blue, 226 }  ,draw opacity=1 ][line width=2.25]    (335,120) -- (391,120) ;
\draw [shift={(395,120)}, rotate = 180] [color={rgb, 255:red, 74; green, 144; blue, 226 }  ,draw opacity=1 ][line width=2.25]    (17.49,-5.26) .. controls (11.12,-2.23) and (5.29,-0.48) .. (0,0) .. controls (5.29,0.48) and (11.12,2.23) .. (17.49,5.26)   ;
%Straight Lines [id:da07930644258355524] 
\draw [color={rgb, 255:red, 74; green, 144; blue, 226 }  ,draw opacity=1 ][line width=2.25]    (160,50) -- (196,50) ;
\draw [shift={(200,50)}, rotate = 180] [color={rgb, 255:red, 74; green, 144; blue, 226 }  ,draw opacity=1 ][line width=2.25]    (17.49,-5.26) .. controls (11.12,-2.23) and (5.29,-0.48) .. (0,0) .. controls (5.29,0.48) and (11.12,2.23) .. (17.49,5.26)   ;
%Straight Lines [id:da2023725794987401] 
\draw [color={rgb, 255:red, 74; green, 144; blue, 226 }  ,draw opacity=1 ][line width=2.25]    (160,120) -- (196,120) ;
\draw [shift={(200,120)}, rotate = 180] [color={rgb, 255:red, 74; green, 144; blue, 226 }  ,draw opacity=1 ][line width=2.25]    (17.49,-5.26) .. controls (11.12,-2.23) and (5.29,-0.48) .. (0,0) .. controls (5.29,0.48) and (11.12,2.23) .. (17.49,5.26)   ;
%Straight Lines [id:da34167568201258147] 
\draw [color={rgb, 255:red, 74; green, 144; blue, 226 }  ,draw opacity=1 ][line width=2.25]    (40,84.5) -- (106.41,51.77) ;
\draw [shift={(110,50)}, rotate = 153.76] [color={rgb, 255:red, 74; green, 144; blue, 226 }  ,draw opacity=1 ][line width=2.25]    (17.49,-5.26) .. controls (11.12,-2.23) and (5.29,-0.48) .. (0,0) .. controls (5.29,0.48) and (11.12,2.23) .. (17.49,5.26)   ;
%Straight Lines [id:da6518407386227685] 
\draw [color={rgb, 255:red, 74; green, 144; blue, 226 }  ,draw opacity=1 ][line width=2.25]    (40,84.5) -- (106.43,118.19) ;
\draw [shift={(110,120)}, rotate = 206.89] [color={rgb, 255:red, 74; green, 144; blue, 226 }  ,draw opacity=1 ][line width=2.25]    (17.49,-5.26) .. controls (11.12,-2.23) and (5.29,-0.48) .. (0,0) .. controls (5.29,0.48) and (11.12,2.23) .. (17.49,5.26)   ;
%Rounded Rect [id:dp3322348751784556] 
\draw  [fill={rgb, 255:red, 184; green, 233; blue, 134 }  ,fill opacity=0.5 ][line width=3]  (200,18) .. controls (200,13.58) and (203.58,10) .. (208,10) -- (232,10) .. controls (236.42,10) and (240,13.58) .. (240,18) -- (240,152) .. controls (240,156.42) and (236.42,160) .. (232,160) -- (208,160) .. controls (203.58,160) and (200,156.42) .. (200,152) -- cycle ;
%Shape: Circle [id:dp4684886534582584] 
\draw  [line width=3]  (110,50) .. controls (110,36.19) and (121.19,25) .. (135,25) .. controls (148.81,25) and (160,36.19) .. (160,50) .. controls (160,63.81) and (148.81,75) .. (135,75) .. controls (121.19,75) and (110,63.81) .. (110,50) -- cycle ;
%Curve Lines [id:da38958374585715694] 
\draw [color={rgb, 255:red, 249; green, 6; blue, 36 }  ,draw opacity=1 ][line width=2.25]    (267,97.5) .. controls (271,193.5) and (260,175.5) .. (228,177.5) .. controls (196,179.5) and (43.5,173.88) .. (33,178.5) .. controls (22.87,182.96) and (28.57,196.51) .. (28.08,230.7) ;
\draw [shift={(28,234.5)}, rotate = 271.55] [color={rgb, 255:red, 249; green, 6; blue, 36 }  ,draw opacity=1 ][line width=2.25]    (17.49,-5.26) .. controls (11.12,-2.23) and (5.29,-0.48) .. (0,0) .. controls (5.29,0.48) and (11.12,2.23) .. (17.49,5.26)   ;
%Straight Lines [id:da09414291890410253] 
\draw [color={rgb, 255:red, 74; green, 144; blue, 226 }  ,draw opacity=1 ][line width=2.25]    (160,215) -- (300,215) ;
\draw [shift={(300,215)}, rotate = 180] [color={rgb, 255:red, 74; green, 144; blue, 226 }  ,draw opacity=1 ][line width=2.25]    (17.49,-5.26) .. controls (11.12,-2.23) and (5.29,-0.48) .. (0,0) .. controls (5.29,0.48) and (11.12,2.23) .. (17.49,5.26)   ;
%Straight Lines [id:da4566726553164755] 
\draw [color={rgb, 255:red, 74; green, 144; blue, 226 }  ,draw opacity=1 ][line width=2.25]    (160,290) -- (300,290) ;
\draw [shift={(300,290)}, rotate = 180] [color={rgb, 255:red, 74; green, 144; blue, 226 }  ,draw opacity=1 ][line width=2.25]    (17.49,-5.26) .. controls (11.12,-2.23) and (5.29,-0.48) .. (0,0) .. controls (5.29,0.48) and (11.12,2.23) .. (17.49,5.26)   ;
%Shape: Circle [id:dp7184347547321996] 
\draw  [line width=3]  (110,120) .. controls (110,106.19) and (121.19,95) .. (135,95) .. controls (148.81,95) and (160,106.19) .. (160,120) .. controls (160,133.81) and (148.81,145) .. (135,145) .. controls (121.19,145) and (110,133.81) .. (110,120) -- cycle ;
%Straight Lines [id:da9043909999057977] 
\draw [color={rgb, 255:red, 74; green, 144; blue, 226 }  ,draw opacity=1 ][line width=2.25]    (45,255) -- (111.53,216.98) ;
\draw [shift={(115,215)}, rotate = 150.26] [color={rgb, 255:red, 74; green, 144; blue, 226 }  ,draw opacity=1 ][line width=2.25]    (17.49,-5.26) .. controls (11.12,-2.23) and (5.29,-0.48) .. (0,0) .. controls (5.29,0.48) and (11.12,2.23) .. (17.49,5.26)   ;
%Straight Lines [id:da22773453957248835] 
\draw [color={rgb, 255:red, 74; green, 144; blue, 226 }  ,draw opacity=1 ][line width=2.25]    (45,255) -- (111.42,288.21) ;
\draw [shift={(115,290)}, rotate = 206.57] [color={rgb, 255:red, 74; green, 144; blue, 226 }  ,draw opacity=1 ][line width=2.25]    (17.49,-5.26) .. controls (11.12,-2.23) and (5.29,-0.48) .. (0,0) .. controls (5.29,0.48) and (11.12,2.23) .. (17.49,5.26)   ;
%Shape: Rectangle [id:dp15524996612763187] 
\draw  [line width=3]  (115,190) -- (160,190) -- (160,240) -- (115,240) -- cycle ;
%Shape: Rectangle [id:dp6769556405625714] 
\draw  [line width=3]  (115,265) -- (160,265) -- (160,315) -- (115,315) -- cycle ;
%Rounded Rect [id:dp2084645156577245] 
\draw  [fill={rgb, 255:red, 184; green, 233; blue, 134 }  ,fill opacity=0.5 ][line width=3]  (300,32) .. controls (300,28.13) and (303.13,25) .. (307,25) -- (328,25) .. controls (331.87,25) and (335,28.13) .. (335,32) -- (335,298) .. controls (335,301.87) and (331.87,305) .. (328,305) -- (307,305) .. controls (303.13,305) and (300,301.87) .. (300,298) -- cycle ;
% %Straight Lines [id:da4982904179803531] 
% \draw [color={rgb, 255:red, 74; green, 144; blue, 226 }  ,draw opacity=1 ][line width=2.25]    (250,215) -- (296,215) ;
% \draw [shift={(300,215)}, rotate = 180] [color={rgb, 255:red, 74; green, 144; blue, 226 }  ,draw opacity=1 ][line width=2.25]    (17.49,-5.26) .. controls (11.12,-2.23) and (5.29,-0.48) .. (0,0) .. controls (5.29,0.48) and (11.12,2.23) .. (17.49,5.26)   ;
% %Straight Lines [id:da2690506712494549] 
% \draw [color={rgb, 255:red, 74; green, 144; blue, 226 }  ,draw opacity=1 ][line width=2.25]    (250,290) -- (296,290) ;
% \draw [shift={(300,290)}, rotate = 180] [color={rgb, 255:red, 74; green, 144; blue, 226 }  ,draw opacity=1 ][line width=2.25]    (17.49,-5.26) .. controls (11.12,-2.23) and (5.29,-0.48) .. (0,0) .. controls (5.29,0.48) and (11.12,2.23) .. (17.49,5.26)   ;
% %Rounded Rect [id:dp5933034256263936] 
% \draw  [fill={rgb, 255:red, 184; green, 233; blue, 134 }  ,fill opacity=0.5 ][line width=3]  (190,202) .. controls (190,195.37) and (195.37,190) .. (202,190) -- (238,190) .. controls (244.63,190) and (250,195.37) .. (250,202) -- (250,303) .. controls (250,309.63) and (244.63,315) .. (238,315) -- (202,315) .. controls (195.37,315) and (190,309.63) .. (190,303) -- cycle ;
%Straight Lines [id:da37423358990584354] 
\draw [color={rgb, 255:red, 74; green, 144; blue, 226 }  ,draw opacity=1 ][line width=2.25]    (335,215) -- (391,215) ;
\draw [shift={(395,215)}, rotate = 180] [color={rgb, 255:red, 74; green, 144; blue, 226 }  ,draw opacity=1 ][line width=2.25]    (17.49,-5.26) .. controls (11.12,-2.23) and (5.29,-0.48) .. (0,0) .. controls (5.29,0.48) and (11.12,2.23) .. (17.49,5.26)   ;
%Straight Lines [id:da5232187308501377] 
\draw [color={rgb, 255:red, 74; green, 144; blue, 226 }  ,draw opacity=1 ][line width=2.25]    (335,290) -- (391,290) ;
\draw [shift={(395,290)}, rotate = 180] [color={rgb, 255:red, 74; green, 144; blue, 226 }  ,draw opacity=1 ][line width=2.25]    (17.49,-5.26) .. controls (11.12,-2.23) and (5.29,-0.48) .. (0,0) .. controls (5.29,0.48) and (11.12,2.23) .. (17.49,5.26)   ;
% %Rounded Rect [id:dp7523177544878321] 
% \draw  [fill={rgb, 255:red, 184; green, 233; blue, 134 }  ,fill opacity=0.5 ][line width=3]  (300,197) .. controls (300,193.13) and (303.13,190) .. (307,190) -- (328,190) .. controls (331.87,190) and (335,193.13) .. (335,197) -- (335,303) .. controls (335,306.87) and (331.87,310) .. (328,310) -- (307,310) .. controls (303.13,310) and (300,306.87) .. (300,303) -- cycle ;
%Rounded Rect [id:dp26113956883084555] 
\draw  [color={rgb, 255:red, 208; green, 2; blue, 27 }  ,draw opacity=1 ][dash pattern={on 6.75pt off 4.5pt}][line width=2.25]  (400,71) .. controls (400,51.67) and (417.67,36) .. (437,36) -- (600,36) .. controls (620,36) and (630,51.67) .. (630,71) -- (630,286.5) .. controls (630,305.83) and (620,321.5) .. (600,321.5) -- (437,321.5) .. controls (417.67,321.5) and (400,305.83) .. (400,286.5) -- cycle ;

% Text Node
\draw (415,200) node [anchor=north west][inner sep=0.75pt]   [align=left] {\begin{minipage}[lt]{150pt}\setlength\topsep{0pt}

{\huge \begin{equation*}
\mathbf{h} =F'(\mathbf{h}_{c} ,\mathbf{h}_{s})
\end{equation*}}
\begin{center}

{\huge $\mathbf{z}'$ \textbf{is robust to} $\mathbf{h}_{s}^*$}
\end{center}

\end{minipage}};
% Text Node
\draw (415,40) node [anchor=north west][inner sep=0.75pt]   [align=left] {\begin{minipage}[lt]{145pt}\setlength\topsep{0pt}

{\huge \begin{equation*}
\mathbf{x} =F(c,s)
\end{equation*}}
\begin{center}

{\huge $\mathbf{z}$ \textbf{is robust to} $s^*$}
\end{center}

\end{minipage}};
% Text Node

\draw (20,75) node [anchor=north west][inner sep=0.75pt]   [align=left] {{\Huge $\mathbf{x}$}};
% Text Node
\draw (55,45) node [anchor=north west][inner sep=0.75pt]   [align=left] {{\huge $s^{1}$}};
% Text Node
\draw (55,105) node [anchor=north west][inner sep=0.75pt]   [align=left] {{\huge $s^{2}$}};
% Text Node
\draw (120,35) node [anchor=north west][inner sep=0.75pt]   [align=left] {{\Huge $\mathbf{x}^{1}$}};
% Text Node
\draw (120,105) node [anchor=north west][inner sep=0.75pt]   [align=left] {{\Huge $\mathbf{x}^{2}$}};
% Text Node
\draw (230,35) node [anchor=north west][inner sep=0.75pt]  [rotate=-89.64] [align=left] {\textbf{{\huge Extractor}}};
% % Text Node
% \draw (328,45) node [anchor=north west][inner sep=0.75pt]  [rotate=-89.64] [align=left] {\textbf{{\LARGE Projector}}};
% Text Node
\draw (255,75) node [anchor=north west][inner sep=0.75pt]   [align=left] {{\Huge $\mathbf{h}$}};
% Text Node
\draw (350,75) node [anchor=north west][inner sep=0.75pt]   [align=left] {{\Huge $\mathbf{z}$}};
% Text Node
\draw (115,200) node [anchor=north west][inner sep=0.75pt]   [align=left] {{\Huge $\mathbf{h}^{'1}$}};
% Text Node
\draw (115,275) node [anchor=north west][inner sep=0.75pt]   [align=left] {{\Huge $\mathbf{h}^{'2}$}};
% Text Node
\draw (20,240) node [anchor=north west][inner sep=0.75pt]   [align=left] {{\Huge $\mathbf{h}$}};
% Text Node
\draw (55,205) node [anchor=north west][inner sep=0.75pt]   [align=left] {{\huge $\mathbf{h}_{s}^{1}$}};
% Text Node
\draw (55,275) node [anchor=north west][inner sep=0.75pt]   [align=left] {{\huge $\mathbf{h}_{s}^{2}$}};
% % Text Node
% \draw (231.01,199.02) node [anchor=north west][inner sep=0.75pt]  [rotate=-89.64] [align=left] {\textbf{{\huge Generator}}};
% Text Node
\draw (328,120) node [anchor=north west][inner sep=0.75pt]  [rotate=-89.64] [align=left] {\textbf{{\huge Projector}}};
% Text Node
\draw (210,240) node [anchor=north west][inner sep=0.75pt]   [align=left] {{\Huge $\mathbf{h}'$}};
% Text Node
\draw (350,240) node [anchor=north west][inner sep=0.75pt]   [align=left] {{\Huge $\mathbf{z}'$}};
% Text Node
\draw (390,10) node [anchor=north west][inner sep=0.75pt]   [align=left] {\textbf{\textcolor[rgb]{0.82,0.01,0.11}{{\huge Contrastive Learning}}}};
% Text Node
\draw (-20,325) node [anchor=north west][inner sep=0.75pt]   [align=left] {{\Huge $\mathbf{x}$:image} {\color{white}lllllll}{\Huge $\mathbf{h}$:hidden feature}{\color{white}lllllll}{\Huge $\mathbf{z}$:projected representation}};
\end{tikzpicture}

%% file: sec/2_relatedwork.tex
\section{Related work}

\textbf{Unsupervised Representation Learning}, URL, aims to learn generalized data representations from unlabeled data. Self-supervised Learning (SSL) designs pretext tasks based on either easily constructed labels (discriminative) or reconstructions of original data (generative). Before the surge of contrastive learning, discriminative SSL methods train on manually constructed labels such as colors \cite{zhang2016colorful}, rotations \cite{gidaris2018unsupervised}, positions of split patches \cite{noroozi2016unsupervised}, etc. \textbf{Contrastive Learning}, CL, relies on a simple idea that different views of the same data example should be as similar as possible. In other words, the representation of an image should be robust to any distortions applied to the image's appearance. Many CL approaches compute the InfoNCE loss \cite{oord2018representation}. Minimizing the loss pulls representations of similar/positive examples closer and pushes representations of dissimilar/negative examples further in the latent space. \citet{chen2020simple} formulates augmented views of the same image as positive examples and all other representations in a mini training batch as negative examples. \citet{chen2020improved} recognizes the same positive examples but adopts a memory bank to store many momentum-updated representations as negative examples. Other CL methods exclude the notion of negative examples and only predict positive representations from a projected view (BYOL \cite{grill2020bootstrap} and SimSiam \cite{chen2021exploring}) or reduce redundancy in the feature dimension space (Barlow Twin \cite{zbontar2021barlow}).

\textbf{Federated Unsupervised Representation Learning}, FURL, adapts URL in a distributed system that learns a common representation model without supervision while preserving data privacy. \textbf{Federated Learning} allows selected local models to be trained on local data and aggregates these local models to form a global model in a iterative communication process. In the absence of labeled data, unsupervised representation learning such as contrastive learning is applied to the distributed system. However, simply aggregating local models causes deterioration in performances due to local data heterogeneity and small local training size. Methods have been developed to mitigate the deterioration. FedCA \cite{zhang2020federated} utilizes a dictionary storing representations from clients to address the heterogeneity and an alignment model trained on a public dataset on the server to address the misalignment. FedU \cite{zhuang2021collaborative} modifies BYOL that only updates the predictor of the local model if the divergence from the previous communication round is below a certain threshold. \citet{he2021ssfl} explores the flexibility of centralized CL methods in the distributed system and emphasizes the personalization of local models by updating global parameters locally and updating local parameters separately. \citet{wu2021federated} proposes the fusion of features from other clients to reduce the false negative ratio and feature space matching of neighbouring clients to alleviate misalignment. Nonetheless, these methods may violate the user privacy protocol as they share information between local clients, or they require extraneous architecture for each client, which increases the memory burden at clients' ends. FedX \cite{han2022fedx} resolves privacy concerns via distilling knowledge locally and globally with additional loss terms to maximize the embedding similarity between different views. The aforementioned methods are composed for a specific contrastive learning algorithm. Nonetheless, our proposed method can be a supplement for different contrastive learning methods to improve both generalization and local personalization.

\textbf{Personalized Federated Learning}, PFL, focuses on better local performances with auxiliary local architecture or information. Aggregating diverse local models trained on heterogeneous data causes a slow convergence of the global model and results in poor local performances. Multitask-learning-based PFL \cite{dinh2021fedu,marfoq2021federated} trains a personalized model for each client by considering each client as a different task and computing similarity between pairwise client relationships. Clustering-based PFL \cite{sattler2020clustered,briggs2020federated} clusters local models based on group-level client relationships and adapts new client models based on the clusters. Meta-learning-based PFL \cite{fallah2020personalized, khodak2019adaptive} adapts the global model to a personalized model by finetuning locally. However, the aforementioned types of PFL require extra computational power to compute similarity or finetuning, and this requirement slows down the learning process. Without labeled data, unsupervised learning expects an even longer training time to yield good performance. Model-based PFL learns a global model for the common knowledge while learning local models for local knowledge. To learn the local model only requires regularization to the global model or partial local model update on the fly. FedPer \cite{arivazhagan2019federated} only updates the base of the global model at local ends while finetunes the MLP classifier locally. FedRep \cite{collins2021exploiting} trains the encoder and the classifier separately for different local epochs and only aggregates the local base encoders on the server. APFL \cite{deng2020adaptive} learns a  convex mixture of a local and a global model as the personalized model. FedProx \cite{li2020federated} regularizes the local model to the current global model to avoid fast divergence while allowing the local model to converge better on local data. Ditto \cite{li2021ditto} extends FedProx by optimizing the global model locally and regularizing the local model to the optimized global model. We compare with these model-based PFL methods as the proposed method includes a style model as the local model and the training involves two models.

%% file: sec/3_fedstyle.tex
\section{Federated with Style (FedStyle)}
Under a decentralized system with unlabeled local data, we adapt unsupervised representation learning to train a generalized global model. Figure \ref{fig:fedstyle} demonstrates three components of the proposed method. We propose to extract local style information by contrasting local original images with sobel filtered images (\ref{fig:extraction}). We propose to improve the generalization of the global model by importing the style information for a style infusion (\ref{fig:infusion}) that is beneficial to unsupervised representation learning. And we propose improve personalization of the local models via stylized content features (\ref{fig:personalize}) that are tailored to local heterogeneous data. Additionally, we provide an analysis of the effectiveness of proposed style-infused loss on learning better representations.

\begin{figure*}[t]
\centering
    \begin{subfigure}[b]{0.3\textwidth}
          \centering
          \resizebox{\linewidth}{!}{\input{images/style_extraction.tikz}}  
          \caption{Local Style Extraction}
          \label{fig:extraction}
     \end{subfigure}
     \rulesep
     \begin{subfigure}[b]{0.42\textwidth}
          \centering
          \resizebox{\linewidth}{!}{\input{images/style_infusion.tikz}}  
          \caption{Local Style Infusion}
          \label{fig:infusion}
     \end{subfigure}
     \rulesep
     \begin{subfigure}[b]{0.2\textwidth}
          \centering
          \resizebox{\linewidth}{!}{\input{images/personalize.tikz}}  
          \caption{Personalization}
          \label{fig:personalize}
     \end{subfigure}
     \caption{Proposed FedStyle process: (a) Extract local style by contrasting original with the Sobel filtered images. (b) Infuse local style feature with the content feature for contrastive learning. (c) Impart both content feature and stylized content feature for personalized tasks.}
     \label{fig:fedstyle}
     \vspace{-5mm}
\end{figure*}

\textbf{Problem Setting} In a distributed system, there are $N$ clients. The current global model weight is updated to a set $\{C\}$ containing $\alpha$ portion of total clients at each communication round. Each selected client, $k\in \{C\}$, trains a model with weights $w_k$ on local data $\mathcal{D}_k$ for $E_l$ local epochs. After local training, the selected clients send $w_k$ to the server, and the server aggregates it as a global model with weights $w$. This process is repeated for $E_g$ communication rounds. The overall goal is to minimize:
\begin{equation}
    \min_{w} \mathcal{L}(w) = \frac{1}{N}\sum_{k=1}^{N}\mathcal{L}_k(w,\mathcal{D}_k),
\end{equation}
where $\mathcal{L}_k(w,\mathcal{D}_k)$ is the local loss of client $k$ on data distribution $\mathcal{D}_k$. In this problem, we assume each $\mathcal{D}_k$ has an user style $s_k$. The federated unsupervised representation learning approach trains a local content model to compute a local loss and averages local content models on the server. Algorithm \ref{algo:furl} outlines the overall process of federated unsupervised representation learning.

\subsection{Local style extraction} \label{sec:styleextract}
We propose to extract local style information for each client by contrasting original local data with a Sobel filtered \cite{kanopoulos1988design} data which is a strong augmentation with a consistent and distinctive style. (Figure \ref{fig:extraction} as a visual description). This consistent style serves as a reference across clients to isolate user styles without sharing local information. Each client is equipped with a style model, $f_{w_s}$, and a style projector, $g_{w_s}$, to extract a style representation $\mathbf{z}_s$ from each data $\mathbf{x}$. $\mathbf{z}_s = g_{w_s}(f_{w_s}(\mathbf{x}))$. To contrast local style $s_k$ with the Sobel filtered style, we compute a supervised InfoNCE loss to separate local style representations from the Sobel style representations. Furthermore, InfoNCE maximizes the mutual information between data with the same local style \cite{tschannen2019mutual} hence the extracted style representations contain as much local information as possible. The style loss of a mini-batch $\mathbf{x}_{\mathcal{B}}\sim{\mathcal{D}_k}$ is:
\begin{equation} \label{eq:style}
    \begin{aligned}
    \mathcal{L}_{style}(\mathbf{x}_{\mathcal{B}}) &=  \sum_{\mathbf{x}_i\in{\mathbf{x}_{\mathcal{B}}},i=1}^{2B}\frac{1}{2B-1}\sum_{\mathbf{x}_j\in{\mathbf{x}_{\mathcal{B}}},j=1}^{2B}\mathbf{1}_{i\neq{j}}\cdot \mathbf{1}_{s_i=s_j} \\
    &\cdot-log\frac{exp(\mathbf{z}_i\cdot{\mathbf{z}_j}/\tau)}{\sum_{n=1}^{2B}\mathbf{1}_{i\neq{n}}\cdot{exp(\mathbf{z}_i\cdot{\mathbf{z}_n/\tau})}},
    \end{aligned}
\end{equation}

where $B=|\mathbf{x}_{\mathcal{B}}|$, $\mathbf{z}_i = g_{w_s}(f_{w_s}(\mathbf{x}_i))$ and $\tau$ is a temperature term. Each image $x_i$ in the mini-batch is processed via a Sobel filter. This doubles the total number of images in the mini-batch. In the end, $2B$ images have two styles with an equal number. Minimizing (\ref{eq:style}) will maximize the similarity between representations of the same style. A detailed style extraction procedure is listed in Algorithm \ref{algo:style}. 

\begin{algorithm}[ht]
\caption{Local style extraction algorithm.}
\label{algo:style}
\textbf{Input}: {\small Local style model, $f_{w_{s},k}$. Local style projector, $g_{w_{s},k}$. Local data distribution, $\mathcal{D}_k$. Local style extraction epochs, $E_s$.}
\begin{algorithmic}[1] %[1] enables line numbers
\FOR {epoch=1:$E_s$}
\FOR {mini batch $\mathbf{x}_{\mathcal{B}}\sim{\mathcal{D}_k}$}
\FOR {$\mathbf{x}_i \sim \mathbf{x}_{\mathcal{B}}$}
\STATE $\mathbf{h}_{s,i} \leftarrow{f_{w_s}(\mathbf{x}_i)}$, $\mathbf{h}_{sobel,i} \leftarrow{f_{w_s}(Sobel(\mathbf{x}_i))}$
\STATE $\mathbf{z}_{s,i} \leftarrow{g_{w_s}(\mathbf{h}_{s,i})}$, $\mathbf{z}_{sobel,i} \leftarrow{g_{w_s}(\mathbf{h}_{sobel,i}))}$
\ENDFOR
\STATE $g_{w_s},f_{w_s} \leftarrow{backward(\mathcal{L}_{style}(\mathbf{x}_{\mathcal{B}}))}$
\ENDFOR
\ENDFOR
\end{algorithmic}
\end{algorithm}

\subsection{Local style infusion for contrastive learning} \label{sec:styleinfusion}
With a style model, each client can extract local style for each local data point. By infusing the extracted local style feature with the generalized content feature, we allow the content model to be also robust to local style distortion. Therefore, we can obtain a more generalized global model when aggregating improved local content models. The process is described in Figure \ref{fig:infusion}. To combine the style feature with the content feature and generate a style-infused content feature, we apply a non-linear MLP as a generator $G_{w_g}$ that takes the concatenation of the content feature and the style feature as inputs. $\mathbf{h}_{c'}^*=G_{w_g}(\textit{concat}[\mathbf{h}_c^*,\mathbf{h}_{s'}])$. The style-infused feature is passed to the content projector, $g_{w_c}$, and we compute a style-infused unsupervised loss based on the output of $g_{w_c}$. The total loss of local training is:
\begin{equation} \label{eq:style_infuse}
\begin{aligned}
    \mathcal{L} = \mathcal{L}_{unsup}(g_{w_c}(&\mathbf{h}_c^1),g_{w_c}(\mathbf{h}_c^2)) + \\ &\lambda\times\mathcal{L}_{unsup}(g_{w_c}(\mathbf{h}_{c'}^1),g_{w_c}(\mathbf{h}_{c''}^2)),
\end{aligned}
\end{equation}
where $\mathcal{L}_{unsup}$ can be any general contrastive learning algorithm (SimCLR, MoCo, SimSiam, BYOL, Barlow Twins, etc.), and $\lambda$ is a scale factor modulating the effect on the style-infused unsupervised learning loss. A detailed algorithm is outlined in Algorithm \ref{algo:infuse_contrast}. The clients in FedStyle only transmit local content models to the server. 
%The overall FedStyle algorithm is listed in Algorithm \ref{algo:fedstyle}. 
\begin{algorithm}[t]
\caption{Local style infusion algorithm.}
\label{algo:infuse_contrast}
\textbf{Input}: {\small Local content model, $f_{w_{c},k}$. Local content projector, $g_{w_{c},k}$. Local style model, $f_{w_{s},k}$. Local generator, $G_{w_g,k}$. Local data distribution, $\mathcal{D}_k$. Local unsupervised loss, $\mathcal{L}_{unsup}$. Scale on style-infused loss, $\lambda$.}
\begin{algorithmic}[1] %[1] enables line numbers
\STATE freeze $f_{w_{s},k}$
\FOR {mini batch $\mathbf{x}_{\mathcal{B}}\sim{\mathcal{D}_k}$}
\FOR{$\mathbf{x}_i \sim \mathbf{x}_{\mathcal{B}}$}
\STATE $\mathbf{h}_{s,i} \leftarrow{f_{w_s}(\mathbf{x}_i)}$, $\mathbf{h}_{sobel,i} \leftarrow{f_{w_s}(Sobel(\mathbf{x}_i))}$
\STATE $\mathbf{x}_i^1,\mathbf{x}_i^2 \leftarrow{Augmentations(\mathbf{x}_i)}$
\STATE $\mathbf{h}_{c,i}^1 \leftarrow{f_{w_c}(\mathbf{x}_i^1)},\mathbf{h}_{c,i}^2 \leftarrow{f_{w_c}(\mathbf{x}_i^2)}$
\STATE $\mathbf{h}_{c',i}^1 \leftarrow{G_{w_g}(\textit{concat}[\mathbf{h}_{c,i}^1,\mathbf{h}_{s,i}])}$
\STATE $\mathbf{h}_{c'',i}^2 \leftarrow{G_{w_g}(\textit{concat}[\mathbf{h}_{c,i}^2,\mathbf{h}_{sobel,i}])}$
\ENDFOR
\STATE {$\begin{aligned}
    g_{w_c},f_{w_c} \leftarrow{backward(\mathcal{L}_{unsup}(g_{w_c}(\mathbf{h}_{c,\mathcal{B}}^1),g_{w_c}(\mathbf{h}_{c,\mathcal{B}}^2))} \\ +\lambda\times\mathcal{L}_{unsup}(g_{w_c}(\mathbf{h}_{c',\mathcal{B}}^1),g_{w_c}(\mathbf{h}_{c'',\mathcal{B}}^2)))
\end{aligned}$}
\ENDFOR
\end{algorithmic}
\end{algorithm}

\begin{algorithm}[h]
\caption{FedStyle algorithm}
\label{algo:fedstyle}
\begin{algorithmic}[1] %[1] enables line numbers
\STATE Initialize $w$, $w_{c,k}$, $w_{s,k}$,$w_{g,k}$, $k\in{[1:N]}$.
\STATE Algorithm \ref{algo:style}$(w_{s,k},\mathcal{D}_k,E_s),k\in[1:N]$
\FOR {$round=1:E_g$}
\STATE Sample $K=\lfloor{\alpha\times{N}}\rfloor$ clients set $\{C\}$.
\FOR {$i\in \{C\}$}
\STATE $w_{c,i}\leftarrow w$
\FOR {$epoch=1:E_l$}
\STATE Algorithm \ref{algo:infuse_contrast}($w_{c,i}$,$w_{s,i}$,$w_{g,i}$,$\mathcal{D}_i$,$\lambda$)
\ENDFOR
\ENDFOR
\STATE $w=\sum_{k\in{\{C\}}}\frac{1}{|\mathcal{D}_k|}w_{c,k}$
\ENDFOR
\end{algorithmic}
\end{algorithm}

\subsection{Personalizing with stylized content feature} \label{sec:stylized}
For personalized downstream tasks, apart from the learned generalized content feature, we also impart the stylized content feature for the tasks (\ref{eq:personalize}). By providing both generalized content feature and stylized latent feature, the downstream model can leverage both generalization and personalization to better fit local distribution. For the personalization linear evaluation, we optimize the generator along with the linear classifier as  the generator continues to learn stylized latent feature. 
\begin{equation} \label{eq:personalize}
    \mathbf{h} = \mathbf{h}_c+G_{w_g}(\mathbf{\textit{concat}[\mathbf{h}_c,\mathbf{h}_s}])
\end{equation}
\subsection{Analysis of style-infused representation} \label{sec:analysis}
During the training of FedStyle, we introduce style-infused representations to compute an additional unsupervised loss in (\ref{eq:style_infuse}). Like the first part of the total loss, the second unsupervised loss trains the content model to learn invariant features between two representations infused with different local styles. In \cite{von2021self}, authors prove that self-supervised learning with data augmentations trains the model to approximate an inverse of a function that simulates the generation process of an image from a content factor and a style factor. The image generation process is assumed to be a smooth and invertible function of the content, $c$, and the style, $s$. $\mathbf{x}=F(c,s)$. The global minimum of the unsupervised loss $\mathcal{L}_{unsup}(g_{w_c}\circ{f_{w_c}}(\mathbf{x}^1),g_{w_c}\circ{f_{w_c}(\mathbf{x}^2)})$ is 
reached when $g\circ{f}$ is approaching the smooth inverse function of $F$. And the content factor is identifiable from the extracted representation through $g\circ{f}$. In other words, the $g\circ{f}$ is learning invariant features that are robust at identifying the content factor. However, representation collapse can happen when $g\circ{f}$ is not invertible. With different constraints like architecture regularizations (batch norm layer or deeper network \cite{fetterman2020understanding}) to maximize the output entropy of the $g\circ{f}$, the collapse can be prevented. Similar to the argument, we assume that the image representation in the latent space, $\mathbf{z}$, is a product of the content feature, $\mathbf{h}_c$, and the style feature, $\mathbf{h}_s$. $\mathbf{z}=F'(\mathbf{h}_c,\mathbf{h}_s)$. We deploy the generator, $G$, so that $g\circ{G}$ is approximating the $F'$ and the content feature is identifiable from the style-infused representation by minimizing the second part of (\ref{eq:style_infuse}). Since this generation process takes place in a low-dimensional latent space, $F'$ can be approximated by two simple networks ($g$ and $G$ are both two-layer MLPs). However, since $g\circ{G}$ is not invertible, a strong emphasis (large $\lambda$) on the second unsupervised loss can cause $G$ to collapse due to the weak architecture regularization. With both unsupervised losses, the local content model learns robust representations that are invariant to local distortions in the latent space as well as various augmentation distortions on the original data.

%% file: images/style_extraction.tikz
\tikzset{every picture/.style={line width=0.75pt}} %set default line width to 0.75pt        

\begin{tikzpicture}[x=0.75pt,y=0.75pt,yscale=-1,xscale=1]
%uncomment if require: \path (0,503); %set diagram left start at 0, and has height of 503

% %Shape: Square [id:dp43723328246731863] 
% \draw  [fill={rgb, 255:red, 74; green, 144; blue, 226 }  ,fill opacity=0.5 ] (270,30) -- (390,30) -- (390,150) -- (270,150) -- cycle ;
% %Shape: Square [id:dp939095924185708] 
% \draw  [fill={rgb, 255:red, 74; green, 144; blue, 226 }  ,fill opacity=0.5 ] (260,40) -- (380,40) -- (380,160) -- (260,160) -- cycle ;
% %Shape: Square [id:dp24798867689640103] 
% \draw  [fill={rgb, 255:red, 74; green, 144; blue, 226 }  ,fill opacity=0.5 ] (40,30) -- (160,30) -- (160,150) -- (40,150) -- cycle ;
% %Shape: Square [id:dp37354294284459844] 
% \draw  [fill={rgb, 255:red, 74; green, 144; blue, 226 }  ,fill opacity=0.5 ] (30,40) -- (150,40) -- (150,160) -- (30,160) -- cycle ;
%Image [id:dp6328568917148987] 
\draw (80,110) node  {\includegraphics[width=90pt,height=90pt]{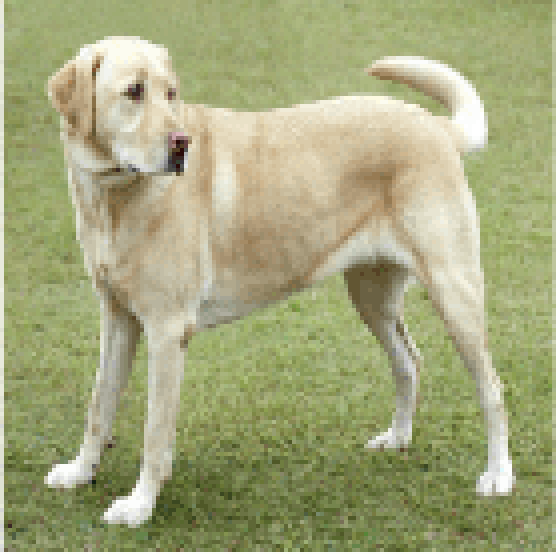}};
%Rounded Rect [id:dp19325491461793876] 
\draw  [fill={rgb, 255:red, 184; green, 233; blue, 134 }  ,fill opacity=0.4 ] (100,260) .. controls (100,250) and (107.16,245) .. (116,245) -- (284,245) .. controls (292.84,245) and (300,250) .. (300,260) -- (300,265) .. controls (300,275) and (292.84,280) .. (284,280) -- (116,280) .. controls (107.16,280) and (100,275) .. (100,265) -- cycle ;
%Image [id:dp12373984815409012] 
\draw (310,110) node  {\includegraphics[width=90pt,height=90pt]{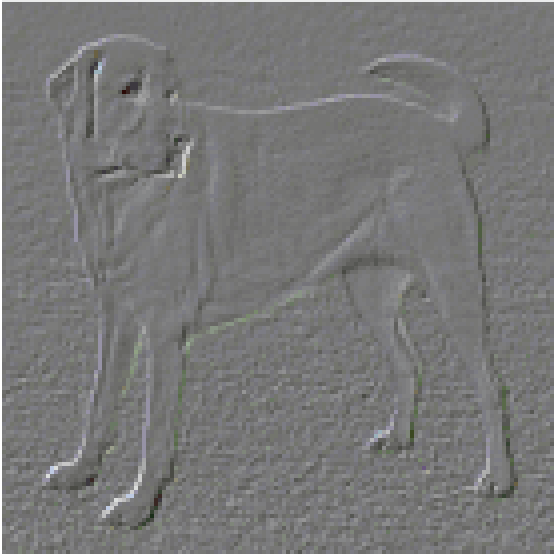}};
%Rounded Rect [id:dp2201970595898295] 
\draw  [fill={rgb, 255:red, 184; green, 233; blue, 134 }  ,fill opacity=0.4 ] (90,392) .. controls (90,396.42) and (95,400) .. (100,400) -- (305,400) .. controls (310,400) and (315,396.42) .. (315,392) -- (315,368) .. controls (315,363.58) and (310,360) .. (305,360) -- (100,360) .. controls (95,360) and (90,363.58) .. (90,368) -- cycle ;
%Straight Lines [id:da7623792349093235] 
\draw [line width=2.25]    (140,100) -- (246,100) ;
\draw [shift={(250,100)}, rotate = 180] [color={rgb, 255:red, 0; green, 0; blue, 0 }  ][line width=2.25]    (17.49,-5.26) .. controls (11.12,-2.23) and (5.29,-0.48) .. (0,0) .. controls (5.29,0.48) and (11.12,2.23) .. (17.49,5.26)   ;
%Straight Lines [id:da8634474582250466] 
\draw [color={rgb, 255:red, 208; green, 2; blue, 27}  ,draw opacity=1 ][line width=2.25]    (80,170) -- (80,190) -- (150,190) -- (150,240) ;
\draw [shift={(150,245)}, rotate = 270] [color={rgb, 255:red, 208; green, 2; blue, 27}  ,draw opacity=1 ][line width=2.25]    (17.49,-5.26) .. controls (11.12,-2.23) and (5.29,-0.48) .. (0,0) .. controls (5.29,0.48) and (11.12,2.23) .. (17.49,5.26)   ;
%Straight Lines [id:da44237544274725193] 
\draw [color={rgb, 255:red, 208; green, 2; blue, 27}  ,draw opacity=1 ][line width=2.25]    (310,170) -- (310,190) -- (240,190) -- (240,240) ;
\draw [shift={(240,245)}, rotate = 270] [color={rgb, 255:red, 208; green, 2; blue, 27}  ,draw opacity=1 ][line width=2.25]    (17.49,-5.26) .. controls (11.12,-2.23) and (5.29,-0.48) .. (0,0) .. controls (5.29,0.48) and (11.12,2.23) .. (17.49,5.26)   ;
%Straight Lines [id:da19308857447422256] 
\draw [color={rgb, 255:red, 208; green, 2; blue, 27}  ,draw opacity=1 ][line width=2.25]    (150,280) -- (150,356) ;
\draw [shift={(150,360)}, rotate = 270] [color={rgb, 255:red, 208; green, 2; blue, 27}  ,draw opacity=1 ][line width=2.25]    (17.49,-5.26) .. controls (11.12,-2.23) and (5.29,-0.48) .. (0,0) .. controls (5.29,0.48) and (11.12,2.23) .. (17.49,5.26)   ;
%Straight Lines [id:da5857435116116934] 
\draw [color={rgb, 255:red, 208; green, 2; blue, 27}  ,draw opacity=1 ][line width=2.25]    (240,280) -- (240,356) ;
\draw [shift={(240,360)}, rotate = 270] [color={rgb, 255:red, 208; green, 2; blue, 27}  ,draw opacity=1 ][line width=2.25]    (17.49,-5.26) .. controls (11.12,-2.23) and (5.29,-0.48) .. (0,0) .. controls (5.29,0.48) and (11.12,2.23) .. (17.49,5.26)   ;
%Straight Lines [id:da02572878396092837] 
\draw [color={rgb, 255:red, 208; green, 2; blue, 27}  ,draw opacity=1 ][line width=2.25]    (150,400) -- (150,456) ;
\draw [shift={(150,460)}, rotate = 270] [color={rgb, 255:red, 208; green, 2; blue, 27}  ,draw opacity=1 ][line width=2.25]    (17.49,-5.26) .. controls (11.12,-2.23) and (5.29,-0.48) .. (0,0) .. controls (5.29,0.48) and (11.12,2.23) .. (17.49,5.26)   ;
%Straight Lines [id:da765965803252326] 
\draw [color={rgb, 255:red, 208; green, 2; blue, 27}  ,draw opacity=1 ][line width=2.25]    (240,400) -- (240,456) ;
\draw [shift={(240,460)}, rotate = 270] [color={rgb, 255:red, 208; green, 2; blue, 27}  ,draw opacity=1 ][line width=2.25]    (17.49,-5.26) .. controls (11.12,-2.23) and (5.29,-0.48) .. (0,0) .. controls (5.29,0.48) and (11.12,2.23) .. (17.49,5.26)   ;

% Text Node
\draw (110,250) node [anchor=north west][inner sep=0.75pt]   [align=left] 
{\huge \textbf{Style model }$f_{w_{s}}$};
% Text Node
\draw (30,20) node [anchor=north west][inner sep=0.75pt]   [align=left] {{\huge \textbf{Original}}};
% Text Node
\draw (235,20) node [anchor=north west][inner sep=0.75pt]   [align=left] {{\huge \textbf{Sobel filtering}}};
% Text Node
\draw (95,365) node [anchor=north west][inner sep=0.75pt]   [align=left] {{\huge \textbf{Style projector }$\displaystyle g_{w_{s}}$}};
% Text Node
\draw (115,420) node [anchor=north west][inner sep=0.75pt]   [align=left] {{\Huge $\displaystyle \mathbf{z}_{s}$}};
% Text Node
\draw (0,470) node [anchor=north west][inner sep=0.75pt]  [font=\Huge] [align=left] {$\displaystyle \mathcal{L}_{style} =InfoNCE(\mathbf{z}_{s} ,\mathbf{z}_{sobel})$};
% Text Node
\draw (110,300) node [anchor=north west][inner sep=0.75pt]   [align=left] {{\Huge $\displaystyle \mathbf{h}_{s}$}};
% Text Node
\draw (250,300) node [anchor=north west][inner sep=0.75pt]   [align=left] {{\Huge $\displaystyle \mathbf{h}_{sobel}$}};
% Text Node
\draw (160,105) node [anchor=north west][inner sep=0.75pt]   [align=left] {{\LARGE $\displaystyle Sobel( .)$}};
% Text Node
\draw (250,420) node [anchor=north west][inner sep=0.75pt]   [align=left] {{\Huge $\displaystyle \mathbf{z}_{sobel}$}};

\end{tikzpicture}

%% file: images/style_infusion.tikz
\tikzset{every picture/.style={line width=0.75pt}} %set default line width to 0.75pt        

\begin{tikzpicture}[x=0.75pt,y=0.75pt,yscale=-1,xscale=1]
%uncomment if require: \path (0,650); %set diagram left start at 0, and has height of 650

%Image [id:dp6641527945394405] 
\draw (70,120) node  {\includegraphics[width=90pt,height=90pt]{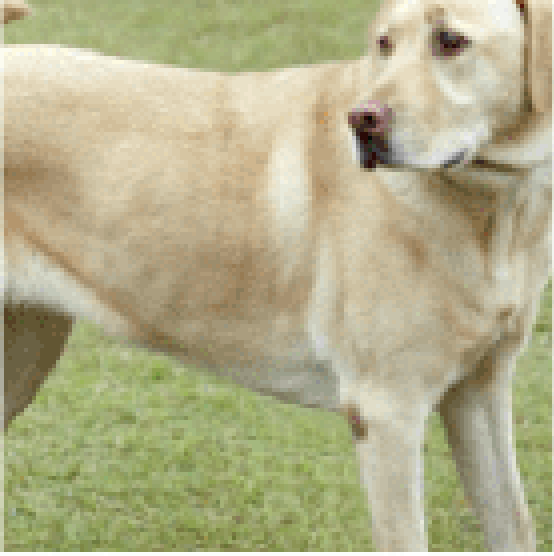}};
%Rounded Rect [id:dp2949236746029156] 
\draw  [fill={rgb, 255:red, 184; green, 233; blue, 134 }  ,fill opacity=0.4 ] (40,235) .. controls (40,230) and (47.16,225) .. (56,225) -- (224,225) .. controls (232.84,225) and (240,230) .. (240,235) -- (240,280) .. controls (240,285) and (232.84,290) .. (224,290) -- (56,290) .. controls (47.16,290) and (40,285) .. (40,280) -- cycle ;
%Image [id:dp09806310786949224] 
\draw (210,120) node  {\includegraphics[width=90pt,height=90pt]{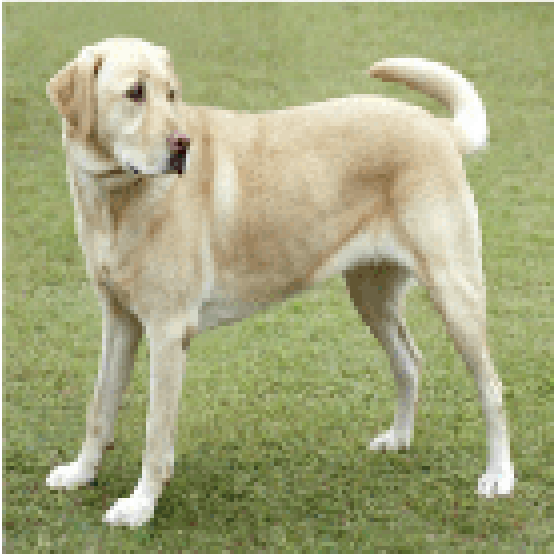}};
%Rounded Rect [id:dp13790791567487592] 
\draw  [fill={rgb, 255:red, 184; green, 233; blue, 134 }  ,fill opacity=0.4 ] (25,442) .. controls (25,446.42) and (28.58,450) .. (33,450) -- (517,450) .. controls (521.42,450) and (525,446.42) .. (525,442) -- (525,418) .. controls (525,413.58) and (521.42,410) .. (517,410) -- (33,410) .. controls (28.58,410) and (25,413.58) .. (25,418) -- cycle ;
%Straight Lines [id:da893566883655353] 
\draw [color={rgb, 255:red, 208; green, 2; blue, 27 }  ,draw opacity=1 ][line width=2.25]    (100,290) -- (100,406) ;
\draw [shift={(100,410)}, rotate = 270] [color={rgb, 255:red, 208; green, 2; blue, 27 }  ,draw opacity=1 ][line width=2.25]    (17.49,-5.26) .. controls (11.12,-2.23) and (5.29,-0.48) .. (0,0) .. controls (5.29,0.48) and (11.12,2.23) .. (17.49,5.26)   ;
%Straight Lines [id:da6884651459614193] 
\draw [color={rgb, 255:red, 208; green, 2; blue, 27 }  ,draw opacity=1 ][line width=2.25]    (180,290) -- (179.03,406) ;
\draw [shift={(179,410)}, rotate = 270.44] [color={rgb, 255:red, 208; green, 2; blue, 27 }  ,draw opacity=1 ][line width=2.25]    (17.49,-5.26) .. controls (11.12,-2.23) and (5.29,-0.48) .. (0,0) .. controls (5.29,0.48) and (11.12,2.23) .. (17.49,5.26)   ;
%Straight Lines [id:da7195266395815691] 
\draw [color={rgb, 255:red, 208; green, 2; blue, 27 }  ,draw opacity=1 ][line width=2.25]    (100,450) -- (100,506) ;
\draw [shift={(100,510)}, rotate = 270] [color={rgb, 255:red, 208; green, 2; blue, 27 }  ,draw opacity=1 ][line width=2.25]    (17.49,-5.26) .. controls (11.12,-2.23) and (5.29,-0.48) .. (0,0) .. controls (5.29,0.48) and (11.12,2.23) .. (17.49,5.26)   ;
%Straight Lines [id:da9860318845043452] 
\draw [color={rgb, 255:red, 208; green, 2; blue, 27 }  ,draw opacity=1 ][line width=2.25]    (180,450) -- (180,506) ;
\draw [shift={(180,510)}, rotate = 270] [color={rgb, 255:red, 208; green, 2; blue, 27 }  ,draw opacity=1 ][line width=2.25]    (17.49,-5.26) .. controls (11.12,-2.23) and (5.29,-0.48) .. (0,0) .. controls (5.29,0.48) and (11.12,2.23) .. (17.49,5.26)   ;
%Image [id:dp9553803211101468] 
\draw (355,100) node  {\includegraphics[width=60pt,height=60pt]{images/original.png}};
%Rounded Rect [id:dp18786318438038418] 
\draw  [fill={rgb, 255:red, 74; green, 144; blue, 226 }  ,fill opacity=0.4 ] (360,181.79) .. controls (360,175.28) and (365.28,170) .. (371.79,170) -- (490,170) .. controls (495,170) and (500,175.28) .. (500,181.79) -- (500,217.15) .. controls (500,223.66) and (495,228.94) .. (490,228.94) -- (371.79,228.94) .. controls (365.28,228.94) and (360,223.66) .. (360,217.15) -- cycle ;
%Image [id:dp9702391908030636] 
\draw (515,100) node  {\includegraphics[width=60pt,height=60pt]{images/sobel.png}};
%Straight Lines [id:da2733656827139892] 
\draw [line width=2.25]    (395,100) -- (475,100) ;
\draw [shift={(475,100)}, rotate = 180] [color={rgb, 255:red, 0; green, 0; blue, 0 }  ][line width=2.25]    (17.49,-5.26) .. controls (11.12,-2.23) and (5.29,-0.48) .. (0,0) .. controls (5.29,0.48) and (11.12,2.23) .. (17.49,5.26)   ;
%Straight Lines [id:da6537633066900919] 
\draw [color={rgb, 255:red, 74; green, 144; blue, 226 }  ,draw opacity=1 ][line width=2.25]    (360,140) -- (360,150) -- (400,150) -- (400,166) ;
\draw [shift={(400,170)}, rotate = 270] [color={rgb, 255:red, 74; green, 144; blue, 226 }  ,draw opacity=1 ][line width=2.25]    (17.49,-5.26) .. controls (11.12,-2.23) and (5.29,-0.48) .. (0,0) .. controls (5.29,0.48) and (11.12,2.23) .. (17.49,5.26)   ;
%Straight Lines [id:da16599616820266583] 
\draw [color={rgb, 255:red, 74; green, 144; blue, 226 }  ,draw opacity=1 ][line width=2.25]    (510,140) -- (510,150) -- (460,150) -- (460,166) ;
\draw [shift={(460,170)}, rotate = 270] [color={rgb, 255:red, 74; green, 144; blue, 226 }  ,draw opacity=1 ][line width=2.25]    (17.49,-5.26) .. controls (11.12,-2.23) and (5.29,-0.48) .. (0,0) .. controls (5.29,0.48) and (11.12,2.23) .. (17.49,5.26)   ;
%Straight Lines [id:da27963457010235016] 
\draw [color={rgb, 255:red, 74; green, 144; blue, 226 }  ,draw opacity=1 ][line width=2.25]    (400,230) -- (400,266) ;
\draw [shift={(400,270)}, rotate = 270] [color={rgb, 255:red, 74; green, 144; blue, 226 }  ,draw opacity=1 ][line width=2.25]    (17.49,-5.26) .. controls (11.12,-2.23) and (5.29,-0.48) .. (0,0) .. controls (5.29,0.48) and (11.12,2.23) .. (17.49,5.26)   ;
%Straight Lines [id:da6208467423118762] 
\draw [color={rgb, 255:red, 74; green, 144; blue, 226 }  ,draw opacity=1 ][line width=2.25]    (460,230) -- (460,266) ;
\draw [shift={(460,270)}, rotate = 270] [color={rgb, 255:red, 74; green, 144; blue, 226 }  ,draw opacity=1 ][line width=2.25]    (17.49,-5.26) .. controls (11.12,-2.23) and (5.29,-0.48) .. (0,0) .. controls (5.29,0.48) and (11.12,2.23) .. (17.49,5.26)   ;
%Rounded Rect [id:dp6746644718414716] 
\draw  [fill={rgb, 255:red, 184; green, 233; blue, 134 }  ,fill opacity=0.4 ] (335,362) .. controls (335,366.42) and (338.58,370) .. (343,370) -- (527,370) .. controls (531.42,370) and (535,366.42) .. (535,362) -- (535,338) .. controls (535,333.58) and (531.42,330) .. (527,330) -- (343,330) .. controls (338.58,330) and (335,333.58) .. (335,338) -- cycle ;
%Straight Lines [id:da12186907348332698] 
\draw [color={rgb, 255:red, 208; green, 2; blue, 27 }  ,draw opacity=1 ][line width=2.25]    (400,290) -- (400,326) ;
\draw [shift={(400,330)}, rotate = 270] [color={rgb, 255:red, 208; green, 2; blue, 27 }  ,draw opacity=1 ][line width=2.25]    (17.49,-5.26) .. controls (11.12,-2.23) and (5.29,-0.48) .. (0,0) .. controls (5.29,0.48) and (11.12,2.23) .. (17.49,5.26)   ;
%Flowchart: Summing Junction [id:dp1983701222388743] 
\draw  [line width=2.25]  (30,590) .. controls (30,585) and (34.48,580) .. (40,580) .. controls (40.52,580) and (50,585) .. (50,590) .. controls (50,595) and (45.52,600) .. (40,600) .. controls (34.48,600) and (30,595) .. (30,590) -- cycle ; \draw  [line width=2.25]  (32.93,582.93) -- (47.07,597.07) ; \draw  [line width=2.25]  (47.07,582.93) -- (32.93,597.07) ;

%Shape: Brace [id:dp4593050008983628] 
\draw  [line width=2.25]  (180.33,59) .. controls (180.33,54.33) and (178,52) .. (173.33,52) -- (150.1,52) .. controls (143.43,52) and (140.1,49.67) .. (140.1,45) .. controls (140.1,49.67) and (136.77,52) .. (130.1,52)(133.1,52) -- (106.86,52) .. controls (102.19,52) and (99.86,54.33) .. (99.86,59) ;
%Curve Lines [id:da6242768177670135] 
\draw [line width=2.25]    (315,100) .. controls (303.17,7.93) and (232.43,18.77) .. (142.72,44.22) ;
\draw [shift={(140,45)}, rotate = 344.05] [color={rgb, 255:red, 0; green, 0; blue, 0 }  ][line width=2.25]    (17.49,-5.26) .. controls (11.12,-2.23) and (5.29,-0.48) .. (0,0) .. controls (5.29,0.48) and (11.12,2.23) .. (17.49,5.26)   ;
%Curve Lines [id:da61550754963921] 
\draw [color={rgb, 255:red, 208; green, 2; blue, 27 }  ,draw opacity=1 ][line width=2.25]    (100,320) .. controls (128.19,339.57) and (270.24,360.79) .. (447.33,281.21) ;
\draw [shift={(450,280)}, rotate = 155.57] [color={rgb, 255:red, 208; green, 2; blue, 27 }  ,draw opacity=1 ][line width=2.25]    (17.49,-5.26) .. controls (11.12,-2.23) and (5.29,-0.48) .. (0,0) .. controls (5.29,0.48) and (11.12,2.23) .. (17.49,5.26)   ;
%Curve Lines [id:da8092587790035053] 
\draw [color={rgb, 255:red, 208; green, 2; blue, 27 }  ,draw opacity=1 ][line width=2.25]    (180,320) .. controls (288.13,320.33) and (335.11,306.25) .. (386.83,281.53) ;
\draw [shift={(390,280)}, rotate = 154.16] [color={rgb, 255:red, 208; green, 2; blue, 27 }  ,draw opacity=1 ][line width=2.25]    (17.49,-5.26) .. controls (11.12,-2.23) and (5.29,-0.48) .. (0,0) .. controls (5.29,0.48) and (11.12,2.23) .. (17.49,5.26)   ;
%Straight Lines [id:da42196213573021546] 
\draw [color={rgb, 255:red, 208; green, 2; blue, 27 }  ,draw opacity=1 ][line width=2.25]    (460,290) -- (460,326) ;
\draw [shift={(460,330)}, rotate = 270] [color={rgb, 255:red, 208; green, 2; blue, 27 }  ,draw opacity=1 ][line width=2.25]    (17.49,-5.26) .. controls (11.12,-2.23) and (5.29,-0.48) .. (0,0) .. controls (5.29,0.48) and (11.12,2.23) .. (17.49,5.26)   ;
%Straight Lines [id:da35279701700200516] 
\draw [color={rgb, 255:red, 208; green, 2; blue, 27 }  ,draw opacity=1 ][line width=2.25]    (400,370) -- (400,406) ;
\draw [shift={(400,410)}, rotate = 270] [color={rgb, 255:red, 208; green, 2; blue, 27 }  ,draw opacity=1 ][line width=2.25]    (17.49,-5.26) .. controls (11.12,-2.23) and (5.29,-0.48) .. (0,0) .. controls (5.29,0.48) and (11.12,2.23) .. (17.49,5.26)   ;
%Straight Lines [id:da2384588264532288] 
\draw [color={rgb, 255:red, 208; green, 2; blue, 27 }  ,draw opacity=1 ][line width=2.25]    (460,370) -- (460,406) ;
\draw [shift={(460,410)}, rotate = 270] [color={rgb, 255:red, 208; green, 2; blue, 27 }  ,draw opacity=1 ][line width=2.25]    (17.49,-5.26) .. controls (11.12,-2.23) and (5.29,-0.48) .. (0,0) .. controls (5.29,0.48) and (11.12,2.23) .. (17.49,5.26)   ;
%Straight Lines [id:da31971732975740275] 
\draw [color={rgb, 255:red, 208; green, 2; blue, 27 }  ,draw opacity=1 ][line width=2.25]    (400,450) -- (400,506) ;
\draw [shift={(400,510)}, rotate = 270] [color={rgb, 255:red, 208; green, 2; blue, 27 }  ,draw opacity=1 ][line width=2.25]    (17.49,-5.26) .. controls (11.12,-2.23) and (5.29,-0.48) .. (0,0) .. controls (5.29,0.48) and (11.12,2.23) .. (17.49,5.26)   ;
%Straight Lines [id:da9757820522757732] 
\draw [color={rgb, 255:red, 208; green, 2; blue, 27 }  ,draw opacity=1 ][line width=2.25]    (460,450) -- (460,506) ;
\draw [shift={(460,510)}, rotate = 270] [color={rgb, 255:red, 208; green, 2; blue, 27 }  ,draw opacity=1 ][line width=2.25]    (17.49,-5.26) .. controls (11.12,-2.23) and (5.29,-0.48) .. (0,0) .. controls (5.29,0.48) and (11.12,2.23) .. (17.49,5.26)   ;
%Straight Lines [id:da8554913432193709] 
\draw [color={rgb, 255:red, 208; green, 2; blue, 27 }  ,draw opacity=1 ][line width=2.25]    (100,180) -- (100,220) ;
\draw [shift={(100,225)}, rotate = 270] [color={rgb, 255:red, 208; green, 2; blue, 27 }  ,draw opacity=1 ][line width=2.25]    (17.49,-5.26) .. controls (11.12,-2.23) and (5.29,-0.48) .. (0,0) .. controls (5.29,0.48) and (11.12,2.23) .. (17.49,5.26)   ;
%Straight Lines [id:da9439071842786626] 
\draw [color={rgb, 255:red, 208; green, 2; blue, 27 }  ,draw opacity=1 ][line width=2.25]    (180,180) -- (180,220) ;
\draw [shift={(180,225)}, rotate = 270] [color={rgb, 255:red, 208; green, 2; blue, 27 }  ,draw opacity=1 ][line width=2.25]    (17.49,-5.26) .. controls (11.12,-2.23) and (5.29,-0.48) .. (0,0) .. controls (5.29,0.48) and (11.12,2.23) .. (17.49,5.26)   ;
%Flowchart: Summing Junction [id:dp7459515413589226] 
\draw  [line width=2.25]  (390,280) .. controls (390,274.48) and (394.48,270) .. (400,270) .. controls (405.52,270) and (410,274.48) .. (410,280) .. controls (410,285.52) and (405.52,290) .. (400,290) .. controls (394.48,290) and (390,285.52) .. (390,280) -- cycle ; \draw  [line width=2.25]  (392.93,272.93) -- (407.07,287.07) ; \draw  [line width=2.25]  (407.07,272.93) -- (392.93,287.07) ;
%Flowchart: Summing Junction [id:dp7346750755928633] 
\draw  [line width=2.25]  (450,280) .. controls (450,274.48) and (454.48,270) .. (460,270) .. controls (465.52,270) and (470,274.48) .. (470,280) .. controls (470,285.52) and (465.52,290) .. (460,290) .. controls (454.48,290) and (450,285.52) .. (450,280) -- cycle ; \draw  [line width=2.25]  (452.93,272.93) -- (467.07,287.07) ; \draw  [line width=2.25]  (467.07,272.93) -- (452.93,287.07) ;
%Straight Lines [id:da7088965385084298] 
\draw [color={rgb, 255:red, 208; green, 2; blue, 27 }  ,draw opacity=1 ][line width=2.25]    (230,590) -- (270,590) ;
\draw [shift={(270,590)}, rotate = 180] [color={rgb, 255:red, 208; green, 2; blue, 27 }  ,draw opacity=1 ][line width=2.25]    (17.49,-5.26) .. controls (11.12,-2.23) and (5.29,-0.48) .. (0,0) .. controls (5.29,0.48) and (11.12,2.23) .. (17.49,5.26)   ;
%Straight Lines [id:da4036436973327828] 
\draw [color={rgb, 255:red, 74; green, 144; blue, 226 }  ,draw opacity=1 ][line width=2.25]    (400,590) -- (440,590) ;
\draw [shift={(440,590)}, rotate = 180] [color={rgb, 255:red, 74; green, 144; blue, 226 }  ,draw opacity=1 ][line width=2.25]    (17.49,-5.26) .. controls (11.12,-2.23) and (5.29,-0.48) .. (0,0) .. controls (5.29,0.48) and (11.12,2.23) .. (17.49,5.26)   ;

% Text Node
\draw (48,240) node [anchor=north west][inner sep=0.75pt]   [align=center] {\huge \textbf{Content model }\\ \huge$\displaystyle f_{w_{c}}$};
% Text Node
\draw (160,418) node [anchor=north west][inner sep=0.75pt]   [align=center] {{\huge \textbf{Content projector }$\displaystyle g_{w_{c}}$}};
% Text Node
\draw (65,455) node [anchor=north west][inner sep=0.75pt]   [align=left] {{\huge $\displaystyle \mathbf{z}_{c}^{1}$}};
% Text Node
\draw (65,370) node [anchor=north west][inner sep=0.75pt]   [align=left] {{\huge $\displaystyle \mathbf{h}_{c}^{1}$}};
% Text Node
\draw (185,370) node [anchor=north west][inner sep=0.75pt]   [align=left] {{\huge $\displaystyle \mathbf{h}_{c}^{2}$}};
% Text Node
\draw (185,455) node [anchor=north west][inner sep=0.75pt]   [align=left] {{\huge $\displaystyle \mathbf{z}_{c}^{2}$}};
% Text Node
\draw (360,175) node [anchor=north west][inner sep=0.75pt]   [align=center]
{\huge \textbf{Style model } \\ \huge $\displaystyle f_{w_{s}}$};
% Text Node
\draw (310,35) node [anchor=north west][inner sep=0.75pt]   [align=left] {{\huge \textbf{Original}}};
% Text Node
\draw (485,35) node [anchor=north west][inner sep=0.75pt]   [align=left] {{\huge \textbf{Sobel}}};
% Text Node
\draw (360,235) node [anchor=north west][inner sep=0.75pt]   [align=left] {{\huge $\displaystyle \mathbf{h}_{s}$}};
% Text Node
\draw (470,235) node [anchor=north west][inner sep=0.75pt]   [align=left] {{\huge $\displaystyle \mathbf{h}_{sobel}$}};
% Text Node
\draw (400,105) node [anchor=north west][inner sep=0.75pt]   [align=left] {{\Large $\displaystyle Sobel( .)$}};
% Text Node
\draw (347,335) node [anchor=north west][inner sep=0.75pt]   [align=left] {{\huge \textbf{Generator }$\displaystyle G_{w_{g}}$}};
% Text Node
\draw (55,580) node [anchor=north west][inner sep=0.75pt]   [align=left] {{\huge \textbf{Concatenation}}};
% Text Node
\draw (140,0) node [anchor=north west][inner sep=0.75pt]   [align=left] {{\huge $\displaystyle Augmentations( .)$}};
% Text Node
\draw (360,370) node [anchor=north west][inner sep=0.75pt]   [align=left] {{\huge $\displaystyle \mathbf{h}_{c'}^{2}$}};
% Text Node
\draw (470,370) node [anchor=north west][inner sep=0.75pt]   [align=left] {{\huge $\displaystyle \mathbf{h}_{c''}^{1}$}};
% Text Node
\draw (360,455) node [anchor=north west][inner sep=0.75pt]   [align=left] {{\huge $\displaystyle \mathbf{z}_{c'}^{2}$}};
% Text Node
\draw (470,455) node [anchor=north west][inner sep=0.75pt]   [align=left] {{\huge $\displaystyle \mathbf{z}_{c''}^{1}$}};
% Text Node
\draw (-10,515) node [anchor=north west][inner sep=0.75pt]   [align=left] {{\Huge $\displaystyle \mathcal{L} =\mathcal{L}_{unsup}\left(\mathbf{z}_{c}^{1} ,\mathbf{z}_{c}^{2}\right) +\lambda \times \mathcal{L}_{unsup}\left(\mathbf{z}_{c'}^{1} ,\mathbf{z}_{c''}^{2}\right)$}};
% Text Node
\draw (275,580) node [anchor=north west][inner sep=0.75pt]   [align=left] {\textbf{{\huge \textcolor[rgb]{0.82,0.01,0.11}{Trainable}}}};
% Text Node
\draw (440,580) node [anchor=north west][inner sep=0.75pt]   [align=left] {\textbf{\textcolor[rgb]{0.29,0.56,0.89}{{\huge Detached}}}};

\end{tikzpicture}

%% file: images/personalize.tikz
\tikzset{every picture/.style={line width=0.75pt}} %set default line width to 0.75pt        

\begin{tikzpicture}[x=0.75pt,y=0.75pt,yscale=-1,xscale=1]
%uncomment if require: \path (0,905); %set diagram left start at 0, and has height of 905

%Image [id:dp5637728194226785] 
\draw (180,90) node  {\includegraphics[width=90pt,height=90pt]{images/original.png}};
%Rounded Rect [id:dp19775856988786877] 
\draw  [fill={rgb, 255:red, 74; green, 144; blue, 226 }  ,fill opacity=0.4 ] (10,214) .. controls (10,200.75) and (20.75,190) .. (34,190) -- (166,190) .. controls (179.25,190) and (190,200.75) .. (190,214) -- (190,286) .. controls (190,299.25) and (179.25,310) .. (166,310) -- (34,310) .. controls (20.75,310) and (10,299.25) .. (10,286) -- cycle ;
%Rounded Rect [id:dp33564433900833524] 
\draw  [fill={rgb, 255:red, 74; green, 144; blue, 226 }  ,fill opacity=0.4 ] (210,214) .. controls (210,200.75) and (220.75,190) .. (234,190) -- (366,190) .. controls (379.25,190) and (390,200.75) .. (390,214) -- (390,286) .. controls (390,299.25) and (379.25,310) .. (366,310) -- (234,310) .. controls (220.75,310) and (210,299.25) .. (210,286) -- cycle ;
%Straight Lines [id:da42478251448671567] 
\draw [color={rgb, 255:red, 74; green, 144; blue, 226 }  ,draw opacity=1 ][line width=2.25]    (120,85) -- (80,85) -- (80,186) ;
\draw [shift={(80,190)}, rotate = 270] [color={rgb, 255:red, 74; green, 144; blue, 226 }  ,draw opacity=1 ][line width=2.25]    (17.49,-5.26) .. controls (11.12,-2.23) and (5.29,-0.48) .. (0,0) .. controls (5.29,0.48) and (11.12,2.23) .. (17.49,5.26)   ;
%Straight Lines [id:da8704457569157174] 
\draw [color={rgb, 255:red, 74; green, 144; blue, 226 }  ,draw opacity=1 ][line width=2.25]    (240,85) -- (280,85) -- (280,186) ;
\draw [shift={(280,190)}, rotate = 270] [color={rgb, 255:red, 74; green, 144; blue, 226 }  ,draw opacity=1 ][line width=2.25]    (17.49,-5.26) .. controls (11.12,-2.23) and (5.29,-0.48) .. (0,0) .. controls (5.29,0.48) and (11.12,2.23) .. (17.49,5.26)   ;
%Rounded Rect [id:dp9422126338824286] 
\draw  [fill={rgb, 255:red, 184; green, 233; blue, 134 }  ,fill opacity=0.4 ] (210,533) .. controls (210,539.63) and (215.37,545) .. (222,545) -- (378,545) .. controls (384.63,545) and (390,539.63) .. (390,533) -- (390,497) .. controls (390,490.37) and (384.63,485) .. (378,485) -- (222,485) .. controls (215.37,485) and (210,490.37) .. (210,497) -- cycle ;
%Straight Lines [id:da7442445920539844] 
\draw [color={rgb, 255:red, 74; green, 144; blue, 226 }  ,draw opacity=1 ][line width=2.25]    (280,310) -- (280,381) ;
\draw [shift={(280,385)}, rotate = 270] [color={rgb, 255:red, 74; green, 144; blue, 226 }  ,draw opacity=1 ][line width=2.25]    (17.49,-5.26) .. controls (11.12,-2.23) and (5.29,-0.48) .. (0,0) .. controls (5.29,0.48) and (11.12,2.23) .. (17.49,5.26)   ;
%Straight Lines [id:da8996562815843063] 
\draw [color={rgb, 255:red, 74; green, 144; blue, 226 }  ,draw opacity=1 ][line width=2.25]    (280,425) -- (280,481) ;
\draw [shift={(280,485)}, rotate = 270] [color={rgb, 255:red, 74; green, 144; blue, 226 }  ,draw opacity=1 ][line width=2.25]    (17.49,-5.26) .. controls (11.12,-2.23) and (5.29,-0.48) .. (0,0) .. controls (5.29,0.48) and (11.12,2.23) .. (17.49,5.26)   ;
%Straight Lines [id:da9875219778915729] 
\draw [color={rgb, 255:red, 208; green, 2; blue, 27 }  ,draw opacity=1 ][line width=2.25]    (280,545) -- (280,611) -- (204,611) ;
\draw [shift={(200,611)}, rotate = 360] [color={rgb, 255:red, 208; green, 2; blue, 27 }  ,draw opacity=1 ][line width=2.25]    (17.49,-5.26) .. controls (11.12,-2.23) and (5.29,-0.48) .. (0,0) .. controls (5.29,0.48) and (11.12,2.23) .. (17.49,5.26)   ;
%Straight Lines [id:da36384160198784876] 
\draw [color={rgb, 255:red, 74; green, 144; blue, 226 }  ,draw opacity=1 ][line width=2.25]    (80,310) -- (80,612) -- (156,612) ;
\draw [shift={(160,612)}, rotate = 180] [color={rgb, 255:red, 74; green, 144; blue, 226 }  ,draw opacity=1 ][line width=2.25]    (17.49,-5.26) .. controls (11.12,-2.23) and (5.29,-0.48) .. (0,0) .. controls (5.29,0.48) and (11.12,2.23) .. (17.49,5.26)   ;
%Straight Lines [id:da0847410278751457] 
\draw [color={rgb, 255:red, 74; green, 144; blue, 226 }  ,draw opacity=1 ][line width=2.25]    (80,405) -- (256,405) ;
\draw [shift={(260,405)}, rotate = 180] [color={rgb, 255:red, 74; green, 144; blue, 226 }  ,draw opacity=1 ][line width=2.25]    (17.49,-5.26) .. controls (11.12,-2.23) and (5.29,-0.48) .. (0,0) .. controls (5.29,0.48) and (11.12,2.23) .. (17.49,5.26)   ;
%Straight Lines [id:da3066157896524495] 
\draw [color={rgb, 255:red, 208; green, 2; blue, 27 }  ,draw opacity=1 ][line width=2.25]    (180,621) -- (180,652) ;
\draw [shift={(180,656)}, rotate = 270] [color={rgb, 255:red, 208; green, 2; blue, 27 }  ,draw opacity=1 ][line width=2.25]    (17.49,-5.26) .. controls (11.12,-2.23) and (5.29,-0.48) .. (0,0) .. controls (5.29,0.48) and (11.12,2.23) .. (17.49,5.26)   ;
%Flowchart: Summing Junction [id:dp016005242949060916] 
\draw  [line width=2.25]  (260,405) .. controls (260,393.95) and (268.95,385) .. (280,385) .. controls (291.05,385) and (300,393.95) .. (300,405) .. controls (300,416.05) and (291.05,425) .. (280,425) .. controls (268.95,425) and (260,416.05) .. (260,405) -- cycle ; \draw  [line width=2.25]  (265.86,390.86) -- (294.14,419.14) ; \draw  [line width=2.25]  (294.14,390.86) -- (265.86,419.14) ;
%Flowchart: Or [id:dp4134420675478392] 
\draw  [line width=2.25]  (160,612) .. controls (160,600.95) and (168.95,592) .. (180,592) .. controls (191.05,592) and (200,600.95) .. (200,612) .. controls (200,623.05) and (191.05,632) .. (180,632) .. controls (168.95,632) and (160,623.05) .. (160,612) -- cycle ; \draw  [line width=2.25]  (160,612) -- (200,612) ; \draw  [line width=2.25]  (180,592) -- (180,632) ;
%Flowchart: Summing Junction [id:dp7865418665954582] 
\draw  [line width=2.25]  (10,727) .. controls (10,715.95) and (18.95,707) .. (30,707) .. controls (41.05,707) and (50,715.95) .. (50,727) .. controls (50,738.05) and (41.05,747) .. (30,747) .. controls (18.95,747) and (10,738.05) .. (10,727) -- cycle ; \draw  [line width=2.25]  (15.86,712.86) -- (44.14,741.14) ; \draw  [line width=2.25]  (44.14,712.86) -- (15.86,741.14) ;
%Flowchart: Or [id:dp5660336950634852] 
\draw  [line width=2.25]  (10,777) .. controls (10,765.95) and (18.95,757) .. (30,757) .. controls (41.05,757) and (50,765.95) .. (50,777) .. controls (50,788.05) and (41.05,797) .. (30,797) .. controls (18.95,797) and (10,788.05) .. (10,777) -- cycle ; \draw  [line width=2.25]  (10,777) -- (50,777) ; \draw  [line width=2.25]  (30,757) -- (30,797) ;
%Straight Lines [id:da9750572956277035] 
\draw [color={rgb, 255:red, 208; green, 2; blue, 27 }  ,draw opacity=1 ][line width=2.25]    (10,822) -- (46,822) ;
\draw [shift={(50,822)}, rotate = 180] [color={rgb, 255:red, 208; green, 2; blue, 27 }  ,draw opacity=1 ][line width=2.25]    (17.49,-5.26) .. controls (11.12,-2.23) and (5.29,-0.48) .. (0,0) .. controls (5.29,0.48) and (11.12,2.23) .. (17.49,5.26)   ;
%Straight Lines [id:da7322360696995944] 
\draw [color={rgb, 255:red, 74; green, 144; blue, 226 }  ,draw opacity=1 ][line width=2.25]    (10,857) -- (46,857) ;
\draw [shift={(50,857)}, rotate = 180] [color={rgb, 255:red, 74; green, 144; blue, 226 }  ,draw opacity=1 ][line width=2.25]    (17.49,-5.26) .. controls (11.12,-2.23) and (5.29,-0.48) .. (0,0) .. controls (5.29,0.48) and (11.12,2.23) .. (17.49,5.26)   ;

% Text Node
\draw (10,210) node [anchor=north west][inner sep=0.75pt]   [align=left] {\begin{minipage}[lt]{131.14pt}\setlength\topsep{0pt}
\begin{center}
{\Huge \textbf{Content model }}\\{\Huge $\displaystyle f_{w_{c}}$}
\end{center}

\end{minipage}};
% Text Node
\draw (225,210) node [anchor=north west][inner sep=0.75pt]   [align=left] {\begin{minipage}[lt]{106.67pt}\setlength\topsep{0pt}
\begin{center}
{\Huge \textbf{Style model }}\\{\Huge $\displaystyle f_{w_{s}}$}
\end{center}

\end{minipage}};
% Text Node
\draw (225,485) node [anchor=north west][inner sep=0.75pt]   [align=center] {\Huge \textbf{Generator }\\ \Huge $\displaystyle G_{w_{g}}$};
% Text Node
\draw (130,3) node [anchor=north west][inner sep=0.75pt]   [align=left] {\textbf{{\Huge Original}}};
% Text Node
\draw (47,660) node [anchor=north west][inner sep=0.75pt]   [align=left] {\textbf{{\Huge Downstream tasks}}};
% Text Node
\draw (47,565) node [anchor=north west][inner sep=0.75pt]   [align=left] {{\Huge $\displaystyle \mathbf{h}_{c}$}};
% Text Node
\draw (292,565) node [anchor=north west][inner sep=0.75pt]   [align=left] {{\Huge $\displaystyle \mathbf{h}_{c'}$}};
% Text Node
\draw (292,328) node [anchor=north west][inner sep=0.75pt]   [align=left] {{\Huge $\displaystyle \mathbf{h}_{s}$}};
% Text Node
\draw (57,710) node [anchor=north west][inner sep=0.75pt]   [align=left] {\textbf{{\Huge Concatenation}}};
% Text Node
\draw (57,760) node [anchor=north west][inner sep=0.75pt]   [align=left] {\textbf{{\Huge Addition}}};
% Text Node
\draw (57,800) node [anchor=north west][inner sep=0.75pt]   [align=left] {\textbf{{\Huge \textcolor[rgb]{0.82,0.01,0.11}{Trainable}}}};
% Text Node
\draw (57,840) node [anchor=north west][inner sep=0.75pt]   [align=left] {\textbf{\textcolor[rgb]{0.29,0.56,0.89}{{\Huge Detached}}}};

\end{tikzpicture}

%% file: sec/4_experiments.tex
\section{Experiments}
\textbf{Datasets} We run experiments on three different data types that each has different number of styles of same objects: Digit (MNIST, USPS, SVHN), Office Home (Art, Clipart, Product, Real World), and Adaptiope (Synthetic, Product, Real Life). Detailed information on each data type is listed in the Appendix \ref{appendix:datasets}. Since Office Home and Adaptiope do not have designed training and testing splits, we run 4 random trials\footnote{Random seeds 2011, 2015, 2021, 2022} with different training and testing splits (0.7/0.3 training/testing split) and report the average performance with a 95\% confidence interval. In IID settings, each client has the same amount of samples across classes, and the total number of images for each client is the same so that local models are optimized with the same training steps. In non-IID settings, the amount of samples in each class on each client is sampled from a Dirichlet distribution parameterized by the concentration $\beta$ (see Appendix \ref{appendix:non-iid}), and all clients have the same amount of total images.

\noindent\textbf{Federated learning settings} Both local style model and local content model are ResNet18 \cite{he2016deep}, and the projector and the generator are two-layer MLPs \cite{haykin1994neural} with the ReLU activation. The hidden dimension of the MLPs is set to 256 by default, and the output dimension is set to 128. We run $E_g=100$ communication rounds, and $E_l=5$ local epochs. $\alpha=1.0$ when there is only one client for each style, and $\alpha=0.5$ when there are \{5,10\} clients for each style. For the FedStyle setting, $E_s=10$ and $\lambda$ is listed in the result figures. The input size of Digit datasets is 32 by 32 and other data types is 224 by 224. For other personalized federated learning methods, the adaptation to federated unsupervised representation learning and the setting are described in \ref{appendix:adaptation}. We use the SGD optimizer with a learning rate of 0.1 for all approaches. The batch size for all experiments is 64. For unsupervised learning methods, we apply SimCLR \cite{chen2020simple} and SimSiam \cite{chen2021exploring} since these two methods only require one content model. Hence the memory requirement for each client is minimized. Note that SimSiam has an additional two-layer MLP predictor. The temperature term $\tau=0.07$ for all InfoNCE loss functions.

%% file: sec/5_evaluations.tex
\section{Evaluations and Results}
\begin{figure}[htp]
    \centering
    \includegraphics[scale=0.08]{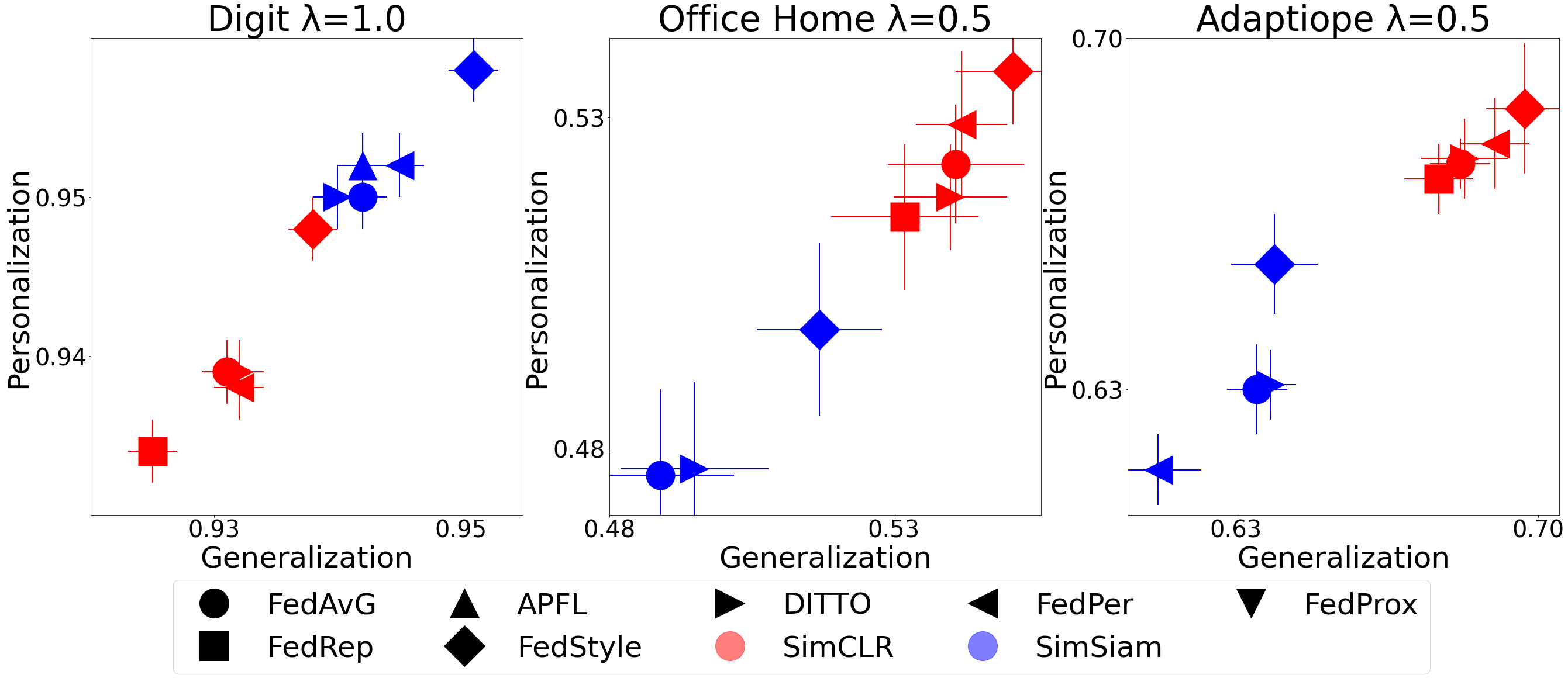}
    \caption{Top right section of Generlization v.s. Personalization accuracy. Points with low generalization and/or personalization accuracy are not shown. See Appendix \ref{appendix:whole results} for detailed results.}
    \label{fig:zoom_result}
\end{figure}

\begin{figure}[htbp]
    \centering
    \includegraphics[scale=0.08]{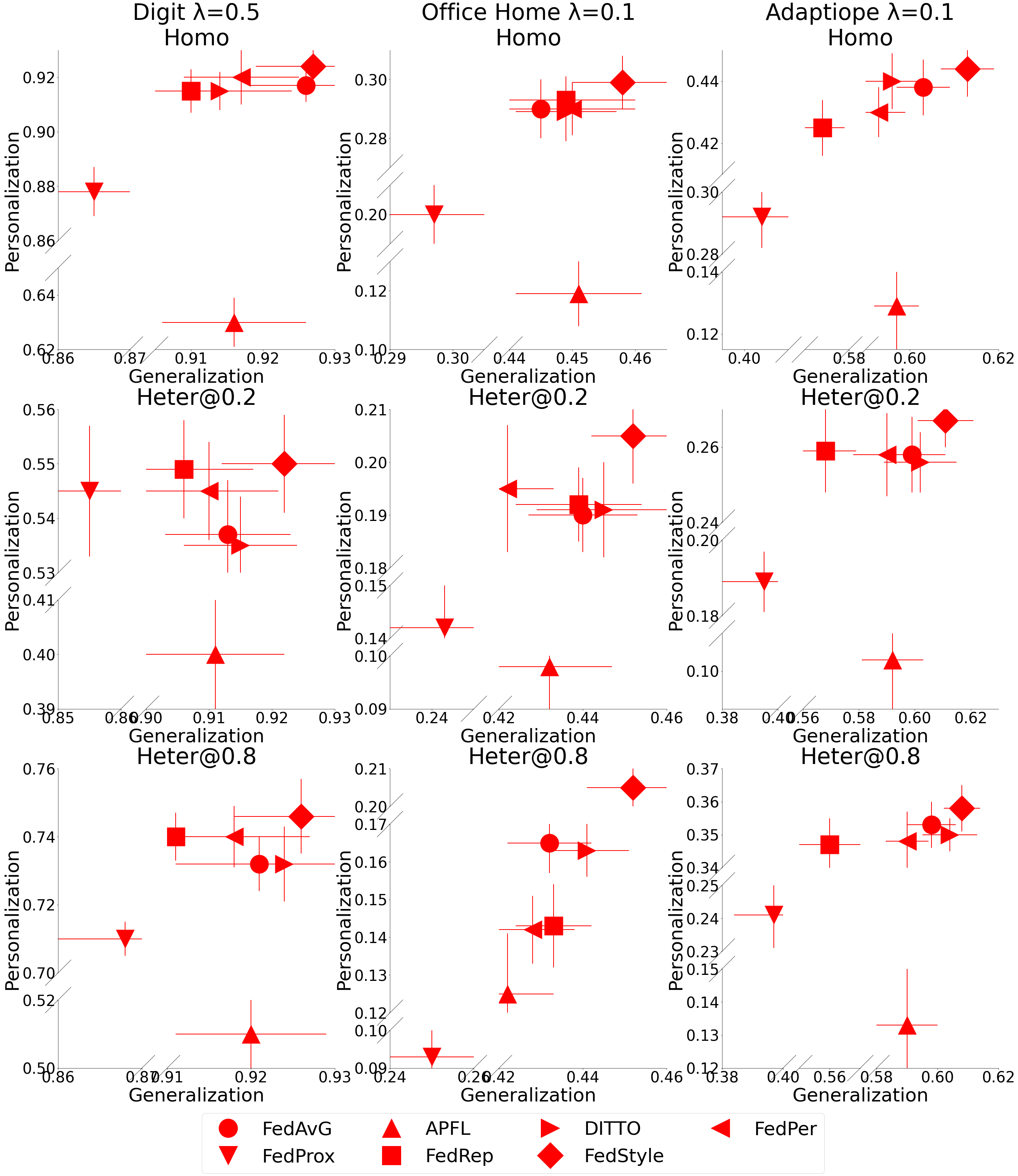}
    \caption{\textbf{SimCLR} \textit{Personalization} v.s. \textit{Generalization} accuracies for three distribution settings. For \textbf{SimSiam} see Appendix \ref{appendix:whole results}.}
    \label{fig:whole_result}
\end{figure}

We evaluate both generalization and personalization by linear evaluation. We follow \cite{zhuang2021collaborative} for the linear evaluation in a decentralized setting. To evaluate the generalization ability of the aggregated global model, the global model is fixed and a linear classifier is trained on all data in each style dataset and evaluates the Top-1 accuracy on the corresponding testing data. To evaluate the personalization ability of a local model, FedStyle fixes both the local content model and local style model and follow (\ref{eq:personalize}) to extract local features. Other baseline methods fix local model(s) to extract local features. The linear classifier is trained on local data only and evaluates the Top-1 accuracy on the corresponding testing data in each style dataset. In Figure \ref{fig:zoom_result} and \ref{fig:whole_result}, the 95\% confidence intervals are indicated as error bars.

% \begin{wrapfigure}{r}{0.45\textwidth}
%     \centering
%     \includegraphics[scale=0.07]{images/single_zoom.png}
%     \caption{Top right section of Generlization v.s. Personalization accuracy.}
%     \label{fig:zoom_result}
%     \vspace{-5mm}
% \end{wrapfigure}

\textbf{Single client in each style dataset} In this setting, each style dataset has only one client and all clients are sampled with an equal amount of unlabeled data. We only evaluate the IID setting where there is no heterogeneity in user preferences (class distributions). In Figure \ref{fig:zoom_result}, we present the \textit{personalization} v.s. \textit{generalization} top-1 accuracy of the proposed method and baseline methods. %Although in digit data type the improvement is limited,
FedStyle is consistently performing better than baseline methods in terms of generalization and personalization. In terms of personalization ability, FedStyle works better than all baseline methods in all datasets, and on average it achieves the highest accuracy in both SimCLR and SimSiam contrastive learning variations. %In digit data type, the improvement is significant in the SVHN dataset. We deduce the reason is that SVHN is a more realistic dataset and hence the style is more dynamic compared to MNIST and USPS datasets. 
In terms of generalization, introducing style-infused representations as additional distortions improves the content model more. The improvement of FedStyle shown in Figure \ref{fig:zoom_result} suggests the effectiveness of the proposed method when there is heterogeneity in user styles. (Points with low generalization and/or personalization accuracy are not shown. See Appendix \ref{appendix:whole results} for detailed results.)

% \begin{wrapfigure}{r}{0.45\textwidth}
%     \centering
%     \includegraphics[scale=0.065]{images/multi_simclr_bar.png}
%     \caption{\textbf{SimCLR} \textit{Personalization} v.s. \textit{Generalization} accuracies for three distribution settings. For \textbf{SimSiam} see Appendix \ref{appendix:whole results}.}
%     \label{fig:whole_result}
% \end{wrapfigure}

\textbf{Multiple clients in each style dataset} Next we evaluate all methods on a slightly larger scale of federated learning. We perform the linear evaluation when there are 5 clients in each style dataset for Office Home and Adaptiope data types and 10 clients in Digit data type with considerations of the size of each data type. We also evaluate all methods in both IID (Homogeneous) and two non-IID (Heterogeneous @$\beta=0.2$ and @$\beta=0.8$ ) settings. %The variance in personalization accuracy is generally higher than that in generalization accuracy because there are many more client models than just one global model and client models are visible to limited local data whereas the global model are visible to the whole dataset. 
As shown in Figure \ref{fig:whole_result} and Figure \ref{fig:whole_simsiam}, FedStyle outperforms all the baselines in both generalization and personalization in all setting of all the datasets. This consistent improvement suggests that FedStyle works at a larger scale where the data is heterogeneous in both style and preference. The overall improvement seems less significant w.r.t the single client setting because only half of the clients are selected at each communication round. On average, each client is only trained for 250 epochs on fewer data. Under an SSL framework, a larger data size and longer training time are essential to achieve better performance. However, there are cases where FedStyle outperforms baselines to a larger extent (Office Home Heter@0.2 and Heter@0.8).
As shown in Figure \ref{fig:whole_result}, the overall improvement seems less significant w.r.t the single client setting for all three data types. This is because that only half of the clients are selected at each communication round. On average, each client is only trained for 250 epochs on fewer data. Under an SSL framework, a larger data size and longer training time are essential to achieve better performance. However, there are cases where FedStyle outperforms baselines to a larger extent (Office Home Heter@0.2 and Heter@0.8). The overall performance of FedStyle in multi-client IID and non-IID settings is the best in terms of generalization and personalization. This consistent improvement suggests that FedStyle works at a larger scale where the data is heterogeneous in both style and preference.

\begin{figure}[h]
    \centering
    \begin{subfigure}[b]{0.45\textwidth}
          \centering
          \resizebox{\linewidth}{!}{\includegraphics{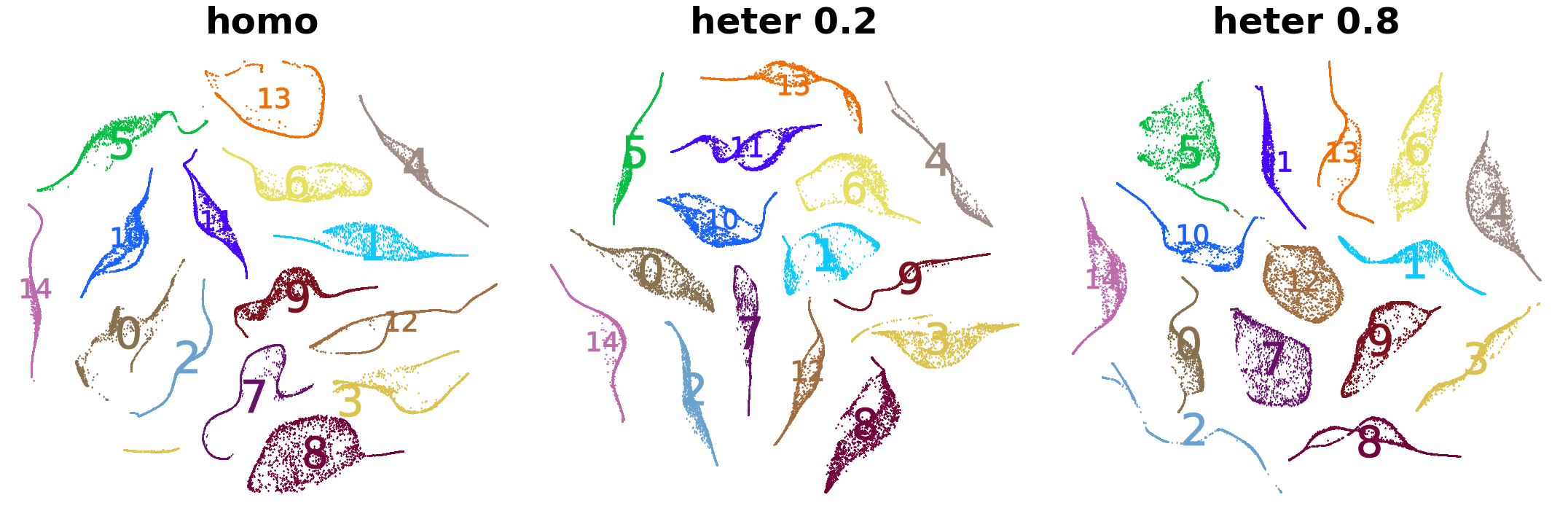}} 
          \caption{Adaptiope @ 15 total clients}
          \label{fig:style_dist_adaptiope}
     \end{subfigure}
     \begin{subfigure}[b]{0.45\textwidth}
          \centering
          \resizebox{\linewidth}{!}{\includegraphics{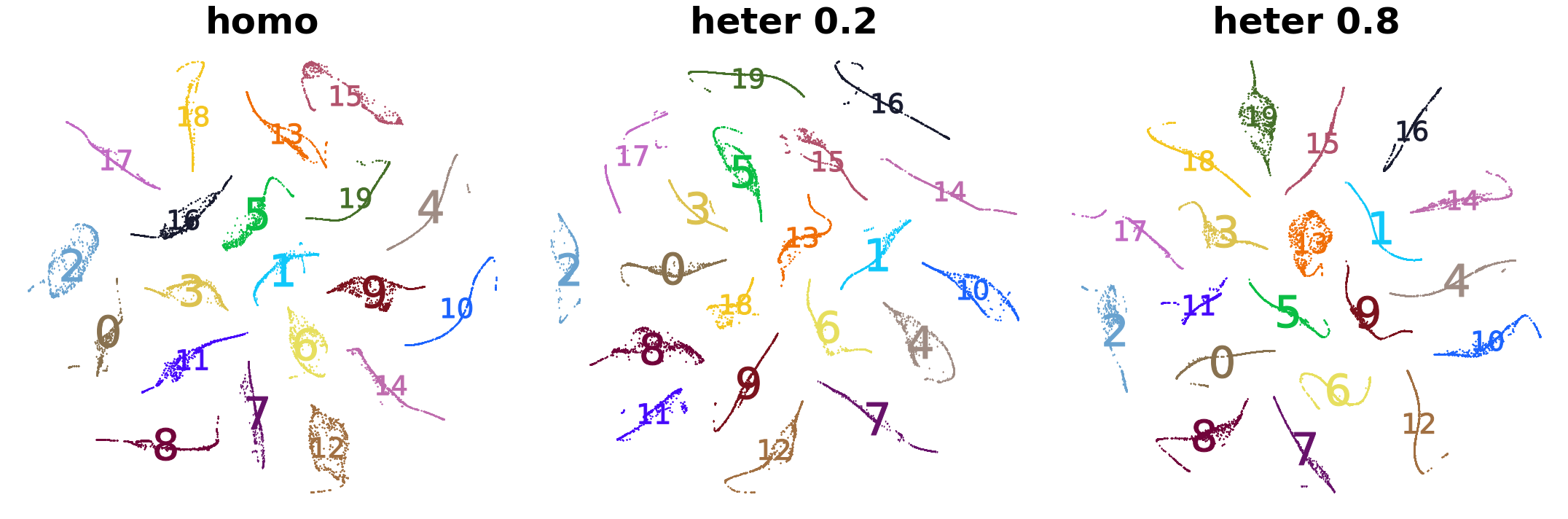}}  
          % \vspace{-5mm}
          \caption{Office Home @ 20 total clients}
          \label{fig:style_dist_office_home}
     \end{subfigure}
    % \vspace{-3mm}
    \caption{t-SNE of extracted style feature distributions.}
    \label{fig:style_dist}
\end{figure}

\textbf{Style feature under non-IID data distribution}
We visualize the distributions of extracted style features, $\mathbf{h}_s$, of all clients using t-SNE \cite{van2008visualizing}. As illustrated in Figure \ref{fig:style_dist}, the relative positions of style features extracted from all clients are consistent across different settings. This suggests that the features extracted via the local style model are consistent with the local style. The shape/distribution of the same client is different across different settings due to the difference in class distributions within each client. The distinctive cluster of each client style feature suggests that the local style model extracts style information that is tailored to the local client and therefore is beneficial for personalized tasks.

\subsection{Ablation studies}

\begin{figure}[htp]
    \centering
    \includegraphics[scale=0.23]{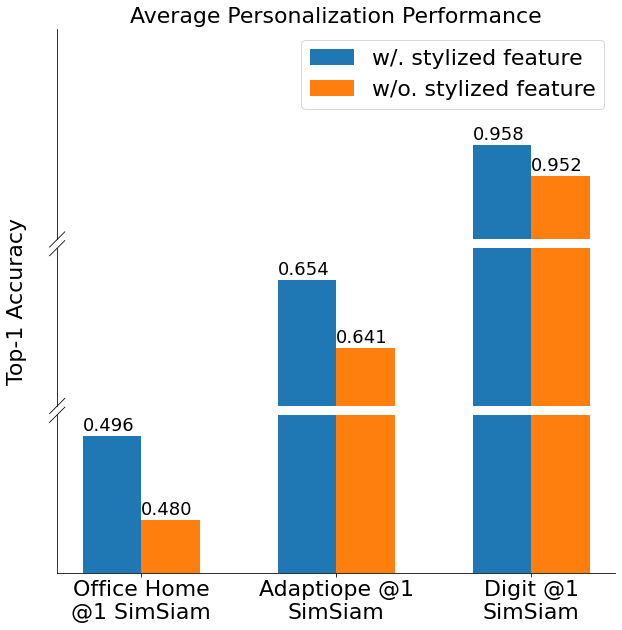}
    \caption{Improvement of introducing stylized features for personal downstream tasks.}
    \label{fig:ablation_stylized}
\end{figure}

\textbf{Stylized content feature} The benefit of the stylized content feature while deploying the downstream tasks is to improve personalization. By imparting both generalized and stylized content features, the combined feature results in better performance. We quantify this benefit by comparing the performances with only the generalized feature. The comparison results of personalization performance in the single-client setting for each style dataset are shown in Figure~\ref{fig:ablation_stylized}. It is shown that by infusing the stylized features, personalized accuracy improves, which illustrates the effectiveness of the stylized content features. It is notable that the improvement of the stylized features on Digit dataset is less significant compared with Adaptiope and Office Home. The reason is that the domain gap between the images of different styles in Digit is much smaller than the other two datasets (e.g., MNIST and USPS in Digit are both binary image datasets of handwritten digits).

With multiple client for each style dataset, the personalization performance is averaged over \{Ho,He@0.2,He@0.8\}. With only the generalized content representation, the performance drops slightly. With the addition of a stylized content feature, the combined feature is more customized to local data and hence delivers better performance. However, the improvement is minimal. Since the projector and the generator approximate $F'\approx{g\circ G}$ by infusing extracted content features and style features, $\mathbf{h}_c,\mathbf{h}_s$, the roles of the structures of $G$ and the quality of extracted style features in learning better stylized content features need to be studied. 

\begin{figure}[htp]
    \centering
    \includegraphics[scale=0.25]{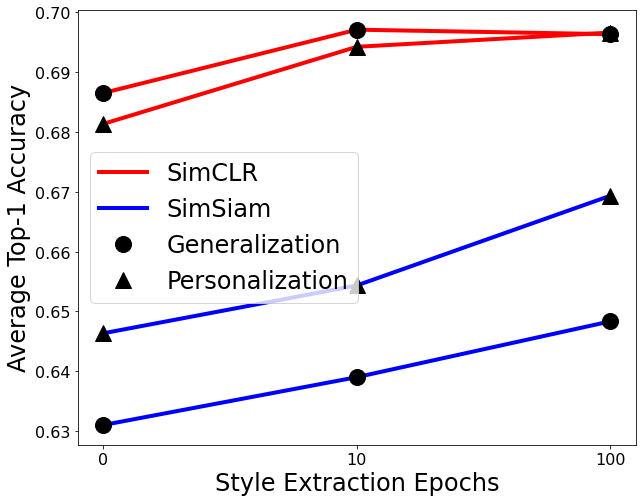}
    \caption{The effect of style extraction epochs $E_s$ on the performance.}
    \label{fig:sepochs}
\end{figure}

\textbf{Style extraction epochs} ($E_s$) To investigate the effect of $E_s$, we run experiments on Adaptiope @ 1 client with $E_s=\{0,10,100\}$. When $E_s=0$, there is only biased noise infused with the content feature. Though this allows the content model to be robust to biased noise in the latent space, there is an improvement when the trained style feature is infused instead of biased noise. Figure \ref{fig:sepochs} shows such improvement when style extraction time increases to 10 and 100 epochs. For SimCLR, prolonging the style extraction time too much only improves the performance marginally. For SimSiam, increasing style extraction time to 100 epochs results in better performance. 
But a downside of continuing style extraction for a long period is that it drastically increases the computational requirement locally.

% \begin{figure}
%      \centering
%      \begin{subfigure}[b]{0.25\textwidth}
%          \centering
%          \includegraphics[width=\textwidth]{images/personalized.png}
%          \caption{}
%     \label{fig:ablation_stylized}
%      \end{subfigure}
%      \qquad
%      \begin{subfigure}[b]{0.35\textwidth}
%          \centering
%          \includegraphics[width=\textwidth]{images/adaptiope_sepochs.png}
%          \caption{}
%          \label{fig:sepochs}
%      \end{subfigure}
%      \caption{(a).Improvement of introducing stylized features for personal downstream tasks. (b).The effect of style extraction epochs $E_s$ on the performance.}
%      \vspace{-5mm}
% \end{figure}

\textbf{Local style-infused loss strength} ($\lambda$)
To investigate the effect of $\lambda$ on the learned representations, we run experiments with $\lambda=\{0.01,0.10,0.50,1.00,5.00,10.0\}$. We report average linear evaluations on Office Home @ 1 client and Adaptiope @ 1 client in Table \ref{tbl:lambda}. For the SimCLR method, as $\lambda$ is too large, personalizing with the generator $G$ fails and collapses due to the weak architecture regularization. In the case of Simsiam, both personalization and generalization deliver much lower accuracy when large $\lambda$ is applied. Simsiam still works to some extent under large $\lambda$ because it regularizes the output entropy of the model via stop-gradient operation while SimCLR only relies on architectural regularization. The results are consistent with our theoretical analysis in Section~\ref{sec:analysis}.

\begin{table}[h]
\caption{}
\begin{subtable}[h]{0.45\textwidth}
\centering
\resizebox{\textwidth}{!}{
\begin{tabular}{|c|cc|cc|}
\hline
          & \multicolumn{2}{c|}{Office Home SimCLR @1 Client}               & \multicolumn{2}{c|}{Adaptiope SimSiam @ 1 Client}               \\ \hline
$\lambda$ & \multicolumn{1}{c|}{Generalization Avg.} & Personalization Avg. & \multicolumn{1}{c|}{Generalization Avg.} & Personalization Avg. \\ \hline
0.01      & \multicolumn{1}{c|}{0.542$\pm$0.009}               & 0.518$\pm$0.013                & \multicolumn{1}{c|}{0.634$\pm$0.007}               & 0.647$\pm$0.011                \\ \hline
0.10      & \multicolumn{1}{c|}{0.540$\pm$0.010}               & 0.512$\pm$0.007                & \multicolumn{1}{c|}{0.635$\pm$0.009}               & 0.649$\pm$0.015                \\ \hline
0.50      & \multicolumn{1}{c|}{0.549$\pm$0.011}               & 0.535$\pm$0.012                & \multicolumn{1}{c|}{0.639$\pm$0.007}               & 0.651$\pm$0.011                \\ \hline
1.00      & \multicolumn{1}{c|}{0.546$\pm$0.010}               & 0.354$\pm$0.227                & \multicolumn{1}{c|}{0.624$\pm$0.007}               & 0.638$\pm$0.010                \\ \hline
5.00      & \multicolumn{1}{c|}{0.539$\pm$0.010}               & 0.022$\pm$0.002                & \multicolumn{1}{c|}{0.542$\pm$0.025}               & 0.566$\pm$0.024                \\ \hline
10.0     & \multicolumn{1}{c|}{0.534$\pm$0.012}               & 0.021$\pm$0.003                & \multicolumn{1}{c|}{0.552$\pm$0.040}               & 0.566$\pm$0.036                \\ \hline
\end{tabular}}
% \vspace{-3mm}
\caption{Effect of $\lambda$ on representations.}
\label{tbl:lambda}
\end{subtable}
\quad
\begin{subtable}[h]{0.45\textwidth}
    \centering
    \resizebox{\textwidth}{!}{
\begin{tabular}{|c|cc|cc|}
\hline
          & \multicolumn{2}{c|}{Office Home SimCLR @1 Client}               & \multicolumn{2}{c|}{Adaptiope SimSiam @ 1 Client}               \\ \hline
 $\mathbf{n}$ & \multicolumn{1}{c|}{Generalization Avg.} & Personalization Avg. & \multicolumn{1}{c|}{Generalization Avg.} & Personalization Avg. \\ \hline
1      & \multicolumn{1}{c|}{0.545$\pm$0.008}               & 0.026$\pm$0.002                & \multicolumn{1}{c|}{0.632$\pm$0.006}               & 0.008$\pm$0.001                \\ \hline
2      & \multicolumn{1}{c|}{0.549$\pm$0.011}               & 0.535$\pm$0.012                & \multicolumn{1}{c|}{0.639$\pm$0.007}               & 0.651$\pm$0.011                \\ \hline
3      & \multicolumn{1}{c|}{0.547$\pm$0.010}               & 0.516$\pm$0.013                & \multicolumn{1}{c|}{0.635$\pm$0.009}               & 0.664$\pm$0.008                \\ \hline
\end{tabular}}
% \vspace{-3mm}
\caption{Effect of number of layers of $G$ on representations.}
\label{tbl:layers}
\end{subtable}
\end{table}

\textbf{Generator structures} ($G$) is set to be a two-layer MLP to approximate the generation function ($F'$) in the latent space along with the projector ($g$), $g\circ{G}\approx{F'}$. We experiment with different numbers of layers ($\mathbf{n}$) to investigate the robustness of the generator to generate stylized content feature for personalization. In Table \ref{tbl:layers}, the MLP is robust in both generalization and personalization to different contrastive variations and different dataset when strong architectural constraint is imposed. Nonetheless, consistent with the analysis in Section\ref{sec:analysis}, the personalization collapses due to weaker architectural constraint when a one-layer linear MLP is used.

%% file: sec/6_conclusions.tex
\section{Conclusion}
In this paper, we propose FedStyle to address the generalization of the global model and personalization of the local models in a user-centered decentralized system with unlabeled data. We propose to extract local style information to improve the global generalization and stylize the content feature to encourage local personalization. We provide an analysis of the effectiveness of style infusion in enhancing the robustness of the learned representation. The proposed method extends to multiple contrastive learning frameworks. In various controlled decentralized settings, FedStyle addresses data heterogeneity caused by heterogeneous user preference and styles.

%% file: sec/X_suppl.tex
\section{Information on datasets} \label{appendix:datasets}
We run experiments on the three data types. Each data type consists of different styles of same objects. This simulates the distinctive user style in a distributed system. The detail of each data type is listed in the Table \ref{tbl:datasets}. To sample equal number of images for each style dataset in a data type, the average number of total training images in all style datasets is sampled (duplicate sampling if necessary).

\begin{table}[h]
\centering
\begin{tabular}{|c|c|c|c|}
\hline
Data types  & Style datasets                 & \# object classes & \# images                               \\ \hline
Digit       & MNIST/SVHN/USPS                & 10                & 70k/$\sim$100k/$\sim$10k                \\ \hline
Office Home & Art/Clipart/Product/Real World & 65                & $\sim$24k/$\sim$44k/$\sim$44k/$\sim$44k \\ \hline
Adaptiope   & Sythetic/Product/Real Life     & 123               & 12.3k/12.3k/12.3k                       \\ \hline
\end{tabular}
\caption{Dataset information}
\label{tbl:datasets}
\end{table}

\section{Non-IID sampling} \label{appendix:non-iid}
\begin{figure}[h]
\captionsetup[subfigure]{justification=centering}
\centering
    \begin{subfigure}[b]{0.3\textwidth}
          \centering
          \resizebox{\linewidth}{!}{\includegraphics{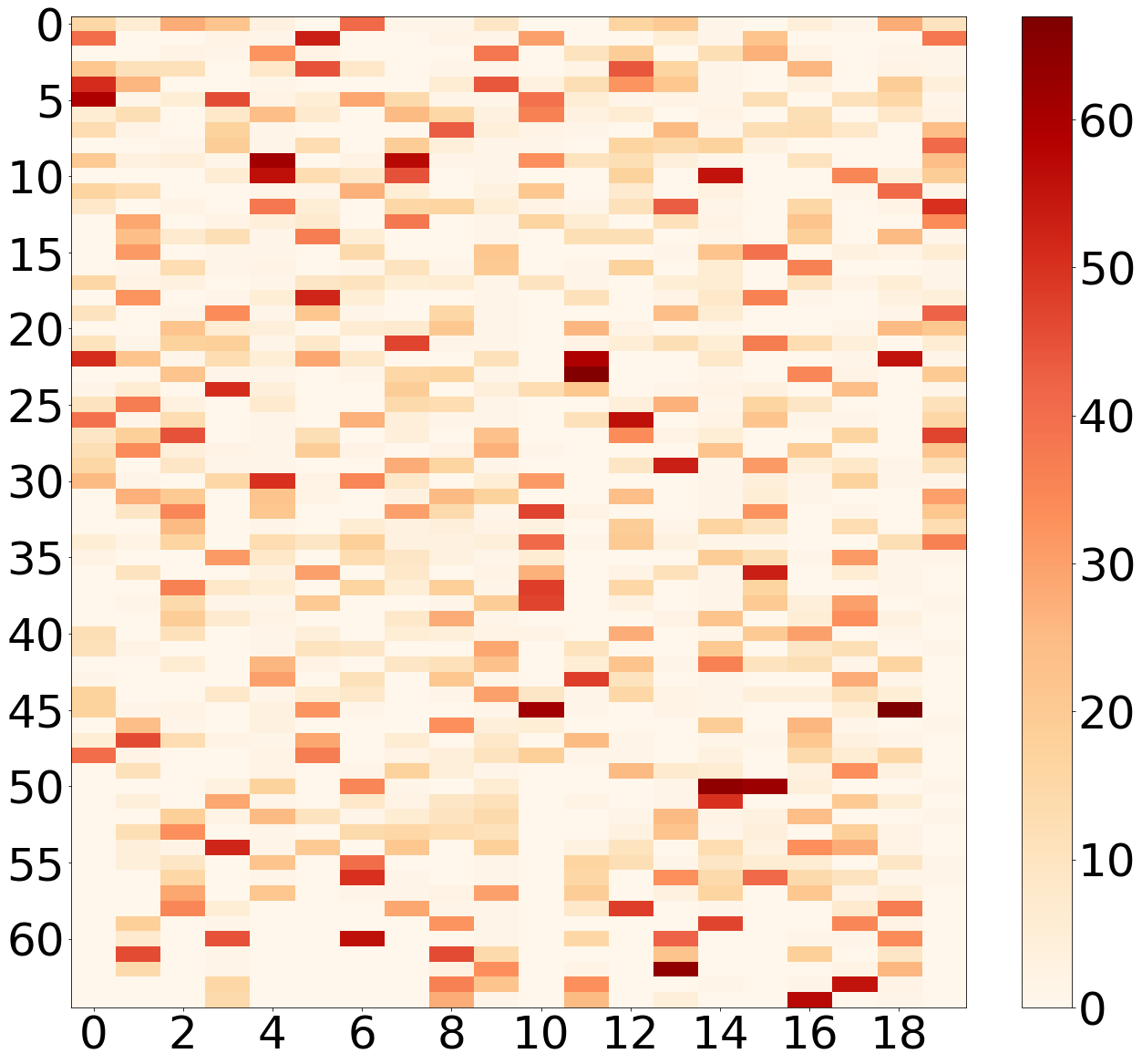}}  
          \caption{Office Home 5 clients in each style, $\beta=0.2$}
     \end{subfigure}
     \begin{subfigure}[b]{0.3\textwidth}
          \centering
          \resizebox{\linewidth}{!}{\includegraphics{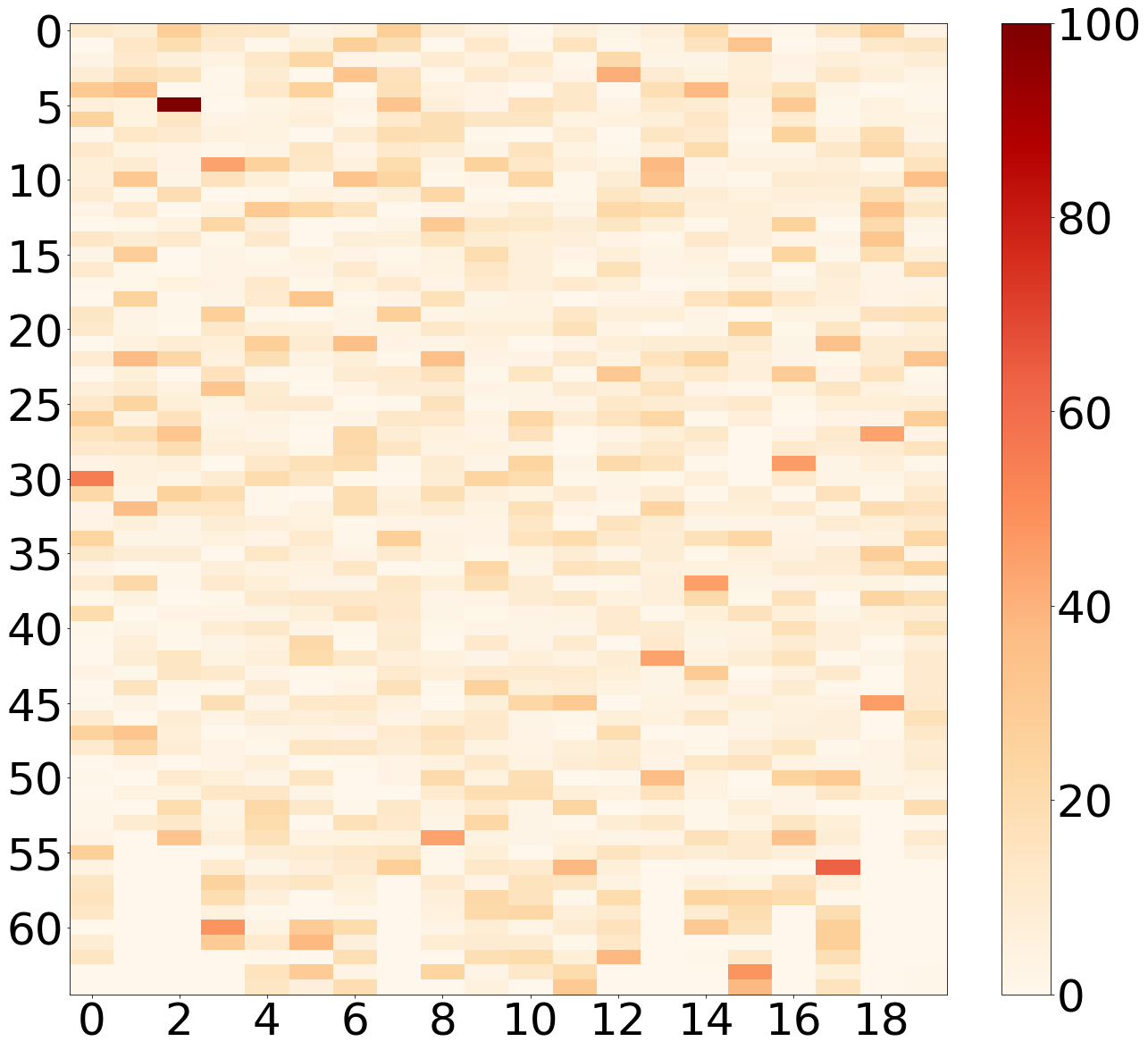}}  
          \caption{Office Home 5 clients in each style, $\beta=0.8$}
     \end{subfigure}
     \begin{subfigure}[b]{0.3\textwidth}
          \centering
          \resizebox{\linewidth}{!}{\includegraphics{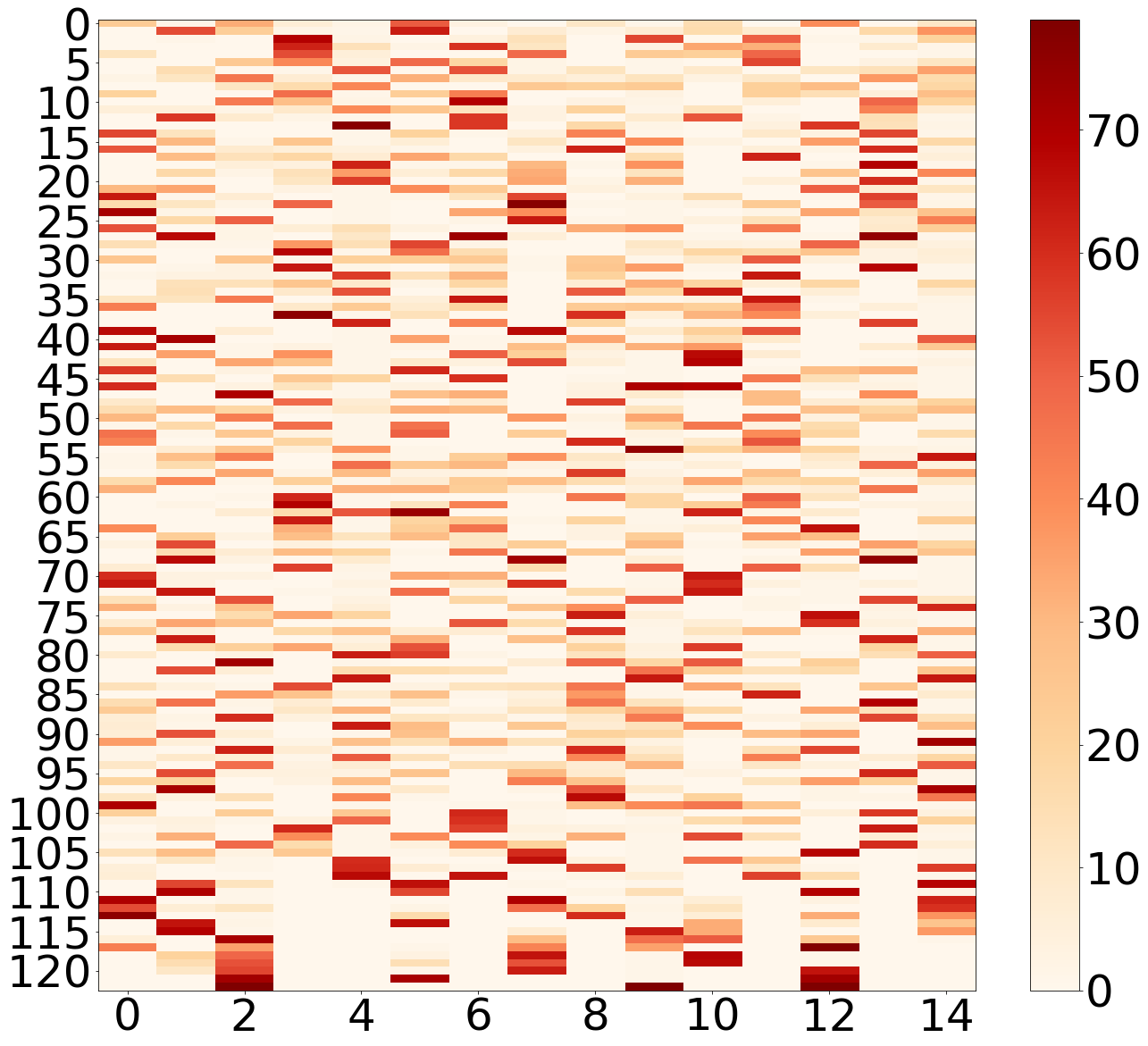}}  
          \caption{Adaptiope 5 clients in each style, $\beta=0.2$}
     \end{subfigure}
     \begin{subfigure}[b]{0.3\textwidth}
          \centering
          \resizebox{\linewidth}{!}{\includegraphics{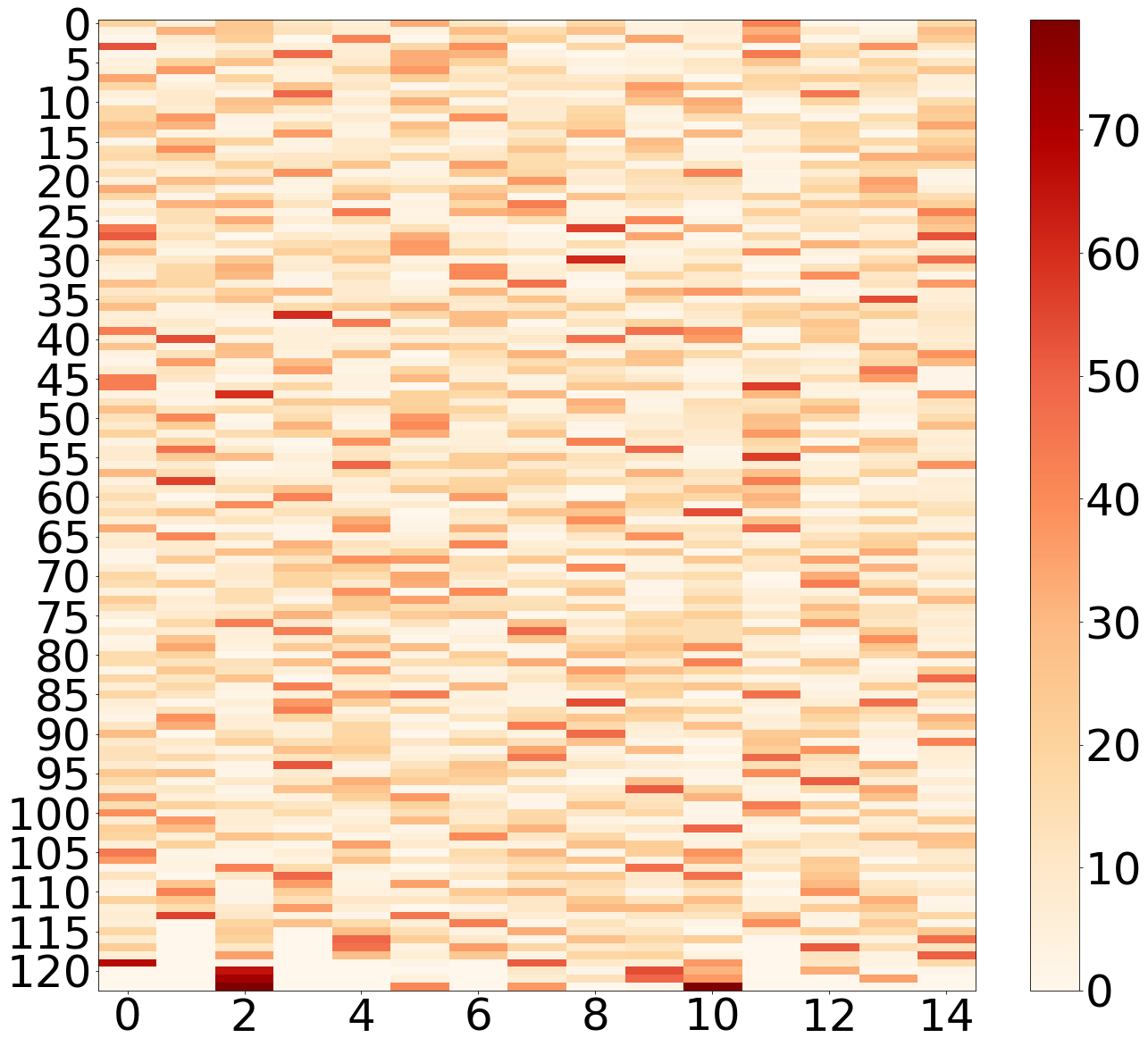}}  
          \caption{Adaptiope 5 clients in each style, $\beta=0.8$}
     \end{subfigure}
     \begin{subfigure}[b]{0.3\textwidth}
          \centering
          \resizebox{\linewidth}{!}{\includegraphics{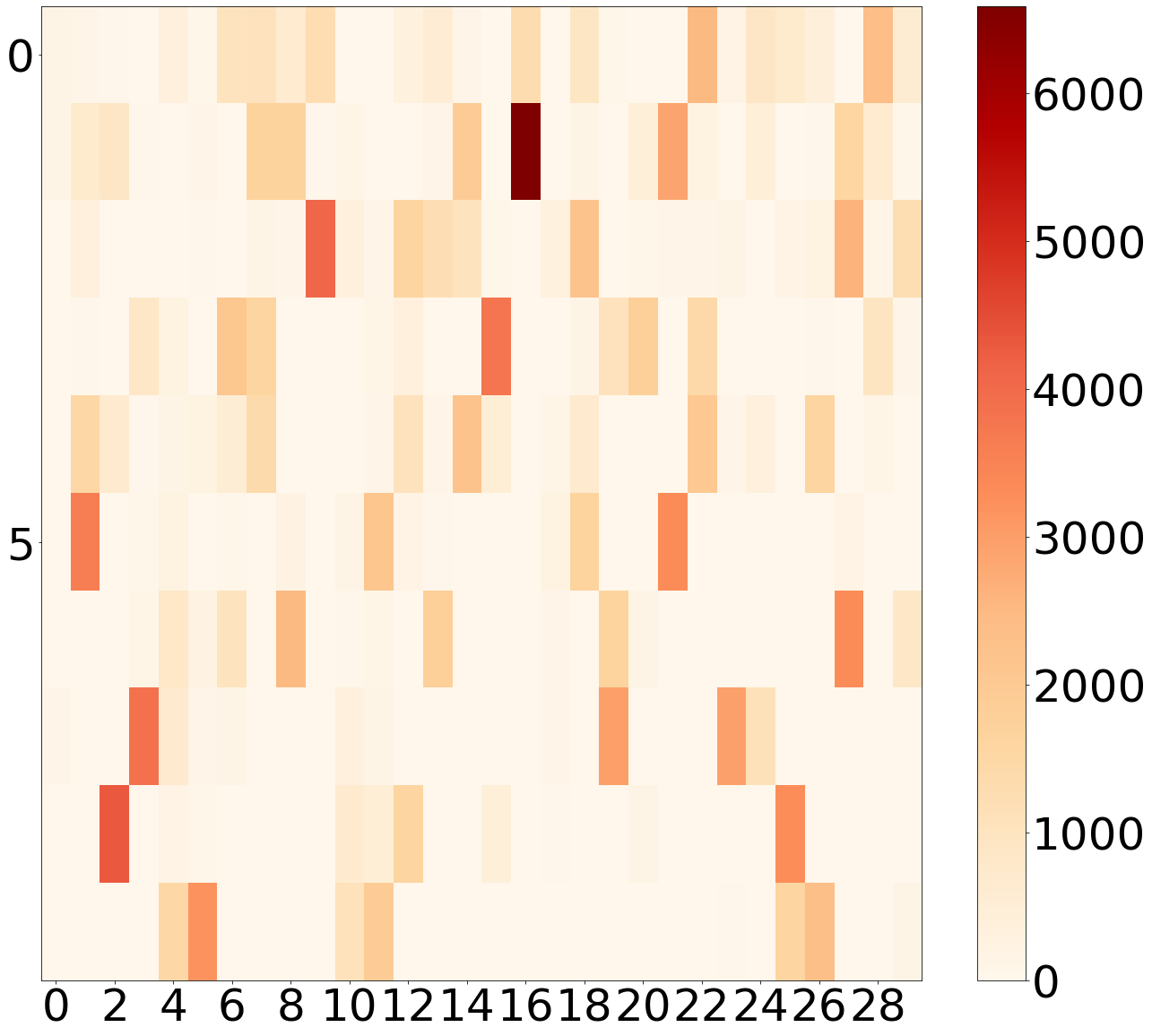}}  
          \caption{Digit 10 clients in each style, \\ $\beta=0.2$}
     \end{subfigure}
     \begin{subfigure}[b]{0.3\textwidth}
          \centering
          \resizebox{\linewidth}{!}{\includegraphics{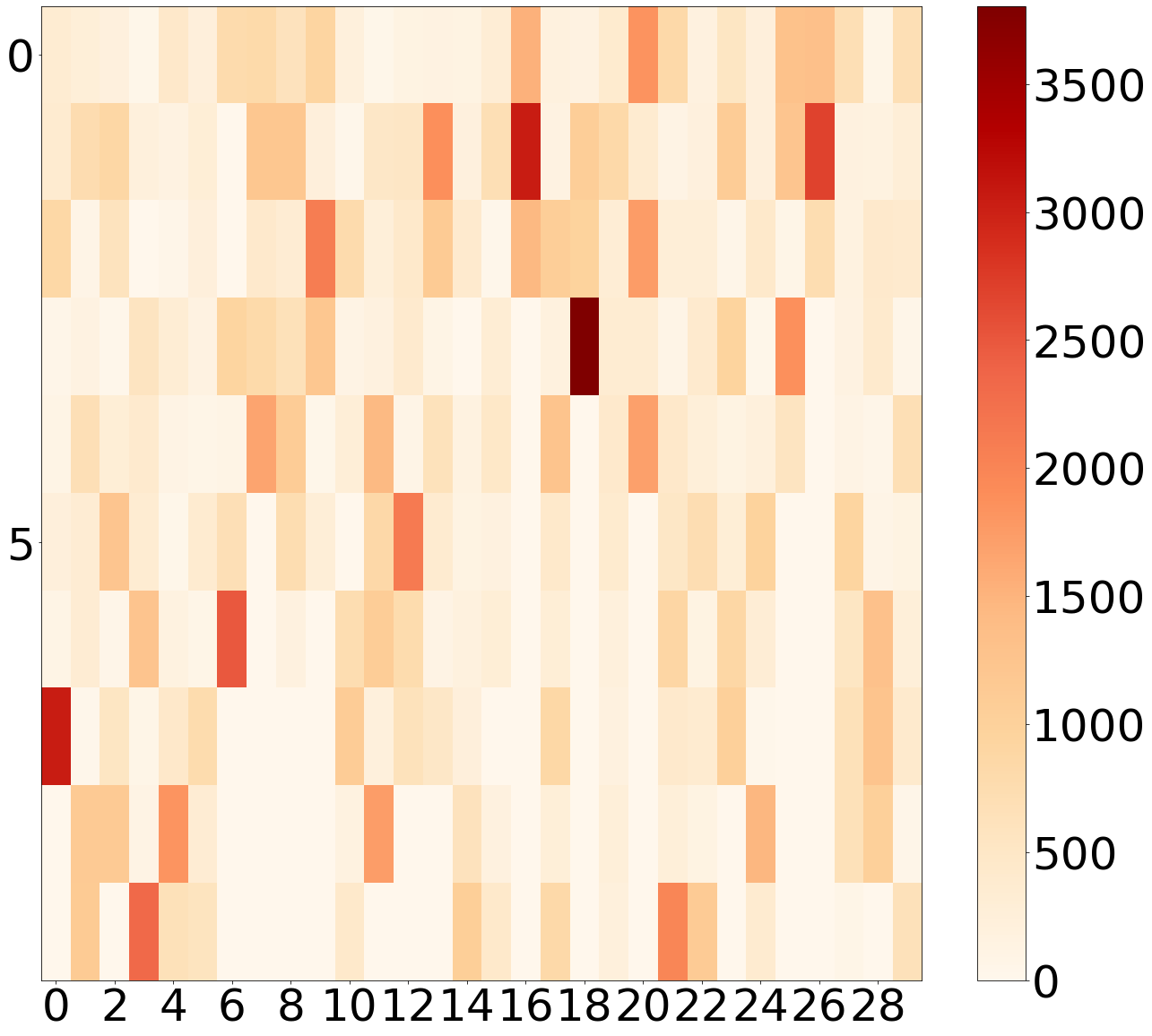}}  
          \caption{Digit 10 clients in each style, \\ $\beta=0.8$}
     \end{subfigure}
     \caption{The data distribution of each client using non-IID data partition. The color bar denotes the number of data samples. Each rectangle represents the number of data samples of a specific class in a client.}
     \label{fig:sampling}
\end{figure}

To sample different proportions of object classes in each client, we use a Dirichlet distribution with concentration $\beta$ to determine the number of each class to be sampled. A larger $\beta$ means a more even distribution among classes. Figure \ref{fig:sampling} shows examples of data distributions with different concentrations for different data types.

\section{Adapting personalized methods for federated unsupervised representation learning} \label{appendix:adaptation}

We adapt all baseline methods (FedAvg, FedProx, FedRep, Fedper, APFL, and Ditto) to the federated unsupervised learning since these methods are originally proposed to solve supervised problems. The workflow for all methods is as illustrated in Algorithm \ref{algo:furl}, but details in $args$, $LocalUpdate()$, and $LocalTraining$ are different and will be described in the following sections. Note that $w$ consists of weights in a feature extraction model $f$ and weights in a feature projector $g$. The default settings for parameters used in the experiments is listed in the [$args:$] for each method.

\begin{algorithm}[htbp]
\caption{Federated unsupervised representation learning algorithm. {\small $*$ denotes a federated learning method (FedAvg, FedRep, FedProx, FedPer, APFL, Ditto, FedStyle)}}
\label{algo:furl}
\textbf{Input}: {\small Server global model, $w$. Number of clients, $N$. Client local models, $w_k$. Local data distribution, $\mathcal{D}_k$. Communication rounds, $E_g$. Local training epochs, $E_l$. Client sample ratio, $\alpha$. Method arguments, $args$.}
\begin{algorithmic}[1] %[1] enables line numbers
\STATE Initialize $w$, $w_{k,k\in{[1:N]}}$.
\FOR {$round=1:E_g$}
\STATE Sample $K=\lfloor{\alpha\times{N}}\rfloor$ clients set $\{C\}$.
\FOR {$i\in \{C\}$}
\STATE \textit{$LocalUpdate_{*}$}($w$,$w_i$)
\FOR {$epoch=1:E_l$}
\STATE minimize $\mathcal{L}_i\leftarrow$\textit{$LocalTraining_{*}$}($w_i$,$\mathcal{D}_i$,$args_{*}$)
\ENDFOR
\ENDFOR
\STATE $w=\sum_{k\in{\{C\}}}\frac{1}{|\mathcal{D}_k|}w_k$ \COMMENT{global aggregate}
\ENDFOR
\end{algorithmic}
\end{algorithm}

\begin{algorithm}[htbp]
\caption*{$LocalUpdate_{*}(w,w_k)$}
% \label{algo:furl}
\textit{\underline{FedAvg}}
\begin{algorithmic}[1] %[1] enables line numbers
\STATE $w_k\leftarrow{w}$ 
\end{algorithmic}
\textit{\underline{FedRep, FedPer}}
\begin{algorithmic} [1] %[1] enables line numbers
\STATE $w^*_k\leftarrow{w^*}$ for $*\in{f}$ 
\end{algorithmic}
\textit{\underline{FedProx, APFL, Ditto}}
\begin{algorithmic}[1] %[1] enables line numbers
\STATE $w_{k,g} \leftarrow{w}$ \COMMENT{Store $w$ temporally as $w_{k,g}$}
\end{algorithmic}
\end{algorithm}

\begin{algorithm}[htbp]
\caption*{$LocalTraining_{*}(w_k,\mathcal{D}_k,args_{*})$}
% \label{algo:furl}
\textit{\underline{FedAvg, FedPer}} [$args$: None]
\begin{algorithmic}[1] %[1] enables line numbers
\STATE Initialize optimizer
\FOR{$\mathbf{x}_{\mathcal{B}}\sim{\mathcal{D}_k}$}
\FOR{$\mathbf{x}_i \sim \mathbf{x}_{\mathcal{B}}$}
\STATE $\mathbf{x}_i^1,\mathbf{x}_i^2 \leftarrow{Augmentations(\mathbf{x}_i)}$
\STATE $\mathbf{h}_{i}^1 \leftarrow{f_{w_k}(\mathbf{x}_i^1)},\mathbf{h}_{i}^2 \leftarrow{f_{w_k}(\mathbf{x}_i^2)}$
\STATE $\mathbf{z}_{i}^1 \leftarrow{g_{w_k}(\mathbf{h}_i^1)},\mathbf{z}_{i}^2 \leftarrow{g_{w_k}(\mathbf{h}_i^2)}$
\ENDFOR
\STATE {$w_k\leftarrow{backward(\mathcal{L}_{unsup}(\mathbf{z}_{\mathcal{B}}^1,\mathbf{z}_{\mathcal{B}}^2))}$}
\ENDFOR
\end{algorithmic}
\textit{\underline{FedRep}} [$args$:None]
\begin{algorithmic}[1] %[1] enables line numbers
\FOR{$epoch=1:E_l$}
\STATE Initialize optimizer, freeze $f_{w_k}$
\FOR{$\mathbf{x}_{\mathcal{B}}\sim{\mathcal{D}_k}$}
\FOR{$\mathbf{x}_i \sim \mathbf{x}_{\mathcal{B}}$}
\STATE $\mathbf{x}_i^1,\mathbf{x}_i^2 \leftarrow{Augmentations(\mathbf{x}_i)}$
\STATE $\mathbf{h}_{i}^1 \leftarrow{f_{w_k}(\mathbf{x}_i^1)},\mathbf{h}_{i}^2 \leftarrow{f_{w_k}(\mathbf{x}_i^2)}$
\STATE $\mathbf{z}_{i}^1 \leftarrow{g_{w_k}(\mathbf{h}_i^1)},\mathbf{z}_{i}^2 \leftarrow{g_{w_k}(\mathbf{h}_i^2)}$
\ENDFOR
\STATE {$g_{w_k}\leftarrow{backward(\mathcal{L}_{unsup}(\mathbf{z}_{\mathcal{B}}^1,\mathbf{z}_{\mathcal{B}}^2))}$}
\ENDFOR
\ENDFOR
\FOR{$epoch=1:E_l$}
\STATE Initialize optimizer, freeze $g_{w_k}$
\FOR{$\mathbf{x}_{\mathcal{B}}\sim{\mathcal{D}_k}$}
\STATE Repeat 4-8.
\STATE {$f_{w_k}\leftarrow{backward(\mathcal{L}_{unsup}(\mathbf{z}_{\mathcal{B}}^1,\mathbf{z}_{\mathcal{B}}^2))}$}
\ENDFOR
\ENDFOR
\end{algorithmic}
\end{algorithm}

\begin{algorithm}[htbp]
\textit{\underline{APFL}} [$args:\alpha=0.5,w_{k,g}$]
\begin{algorithmic}[1] %[1] enables line numbers
\STATE Initialize optimizers
\FOR{$\mathbf{x}_{\mathcal{B}}\sim{\mathcal{D}_k}$}
\FOR{$\mathbf{x}_i \sim \mathbf{x}_{\mathcal{B}}$}
\STATE $\mathbf{x}_i^1,\mathbf{x}_i^2 \leftarrow{Augmentations(\mathbf{x}_i)}$
\STATE $\mathbf{h}_{i,g}^1 \leftarrow{f_{w_{k,g}}(\mathbf{x}_i^1)},\mathbf{h}_{i,g}^2 \leftarrow{f_{w_{k,g}}(\mathbf{x}_i^2)}$
\STATE $\mathbf{z}_{i,g}^1 \leftarrow{g_{w_{k,g}}(\mathbf{h}_{i,g}^1)},\mathbf{z}_{i,g}^2 \leftarrow{g_{w_{k,g}}(\mathbf{h}_{i,g}^2)}$
\STATE $\mathbf{h}_{i}^1 \leftarrow{f_{w_{k}}(\mathbf{x}_i^1)},\mathbf{h}_{i}^2 \leftarrow{f_{w_k}(\mathbf{x}_i^2)}$
\STATE $\mathbf{z}_{i}^1 \leftarrow{g_{w_k}(\mathbf{h}_i^1)},\mathbf{z}_{i}^2 \leftarrow{g_{w_k}(\mathbf{h}_i^2)}$
\STATE $\mathbf{\hat{z}}_{i}^1 \leftarrow{\alpha\times{\mathbf{z}^1_{i,g}.detach()}+(1-\alpha)\times{\mathbf{z}_{i}^2}}$
\STATE $\mathbf{\hat{z}}_{i}^2 \leftarrow{\alpha\times{\mathbf{z}^2_{i,g}.detach()}+(1-\alpha)\times{\mathbf{z}_{i}^2}}$
\ENDFOR
\STATE {$w_{k,g}\leftarrow{backward(\mathcal{L}_{unsup}(\mathbf{z}_{\mathcal{B}}^1,\mathbf{z}_{\mathcal{B}}^2))}$}
\STATE {$w_{k}\leftarrow{backward(\mathcal{L}_{unsup}(\mathbf{\hat{z}}_{\mathcal{B}}^1,\mathbf{\hat{z}}_{\mathcal{B}}^2))}$}
\ENDFOR
\end{algorithmic}
\textit{\underline{FedProx}} [$args:\mu=0.2,w_{k,g}$]
\begin{algorithmic}[1]
\STATE Initialize optimizer
\FOR{$\mathbf{x}_{\mathcal{B}}\sim{\mathcal{D}_k}$}
\FOR{$\mathbf{x}_i \sim \mathbf{x}_{\mathcal{B}}$}
\STATE $\mathbf{x}_i^1,\mathbf{x}_i^2 \leftarrow{Augmentations(\mathbf{x}_i)}$
\STATE $\mathbf{h}_{i}^1 \leftarrow{f_{w_k}(\mathbf{x}_i^1)},\mathbf{h}_{i}^2 \leftarrow{f_{w_k}(\mathbf{x}_i^2)}$
\STATE $\mathbf{z}_{i}^1 \leftarrow{g_{w_k}(\mathbf{h}_i^1)},\mathbf{z}_{i}^2 \leftarrow{g_{w_k}(\mathbf{h}_i^2)}$
\ENDFOR
\STATE {$w_k\leftarrow{backward(\mathcal{L}_{unsup}(\mathbf{z}_{\mathcal{B}}^1,\mathbf{z}_{\mathcal{B}}^2)+\frac{\mu}{2}||w_k-w_{k,g}||^2)}$}
\ENDFOR
\end{algorithmic}
\textit{\underline{Ditto}} [$args:\mu=2,w_{k,g}$]
\begin{algorithmic}[1]
\FOR{$epoch=1:E_l$}
\STATE Initialize optimizer
\FOR{$\mathbf{x}_{\mathcal{B}}\sim{\mathcal{D}_k}$}
\FOR{$\mathbf{x}_i \sim \mathbf{x}_{\mathcal{B}}$}
\STATE $\mathbf{x}_i^1,\mathbf{x}_i^2 \leftarrow{Augmentations(\mathbf{x}_i)}$
\STATE $\mathbf{h}_{i,g}^1 \leftarrow{f_{w_{k,g}}(\mathbf{x}_i^1)},\mathbf{h}_{i,g}^2 \leftarrow{f_{w_{k,g}}(\mathbf{x}_i^2)}$
\STATE $\mathbf{z}_{i,g}^1 \leftarrow{g_{w_{k,g}}(\mathbf{h}_{i,g}^1)},\mathbf{z}_{i,g}^2 \leftarrow{g_{w_{k,g}}(\mathbf{h}_{i,g}^2)}$
\ENDFOR
\STATE {$w_{k,g}\leftarrow{backward(\mathcal{L}_{unsup}(\mathbf{z}_{\mathcal{B},g}^1,\mathbf{z}_{\mathcal{B},g}^2))}$}
\ENDFOR
\ENDFOR
\FOR{$epoch=1:E_l$}
\STATE Initialize optimizer
\FOR{$\mathbf{x}_{\mathcal{B}}\sim{\mathcal{D}_k}$}
\FOR{$\mathbf{x}_i \sim \mathbf{x}_{\mathcal{B}}$}
\STATE $\mathbf{x}_i^1,\mathbf{x}_i^2 \leftarrow{Augmentations(\mathbf{x}_i)}$
\STATE $\mathbf{h}_{i}^1 \leftarrow{f_{w_k}(\mathbf{x}_i^1)},\mathbf{h}_{i}^2 \leftarrow{f_{w_k}(\mathbf{x}_i^2)}$
\STATE $\mathbf{z}_{i}^1 \leftarrow{g_{w_k}(\mathbf{h}_i^1)},\mathbf{z}_{i}^2 \leftarrow{g_{w_k}(\mathbf{h}_i^2)}$
\ENDFOR
\STATE {$\begin{aligned}
    w_k\leftarrow backward(\mathcal{L}_{unsup}(&\mathbf{z}_{\mathcal{B}}^1,\mathbf{z}_{\mathcal{B}}^2) \\ &+\frac{\mu}{2}||w_k-w_{k,g}||^2)
\end{aligned}$}
\ENDFOR
\ENDFOR
\end{algorithmic}
\end{algorithm}

\section{Figures for experiment results}
\label{appendix:whole results}

\begin{figure}[h]
    \centering
    \includegraphics[scale=0.15]{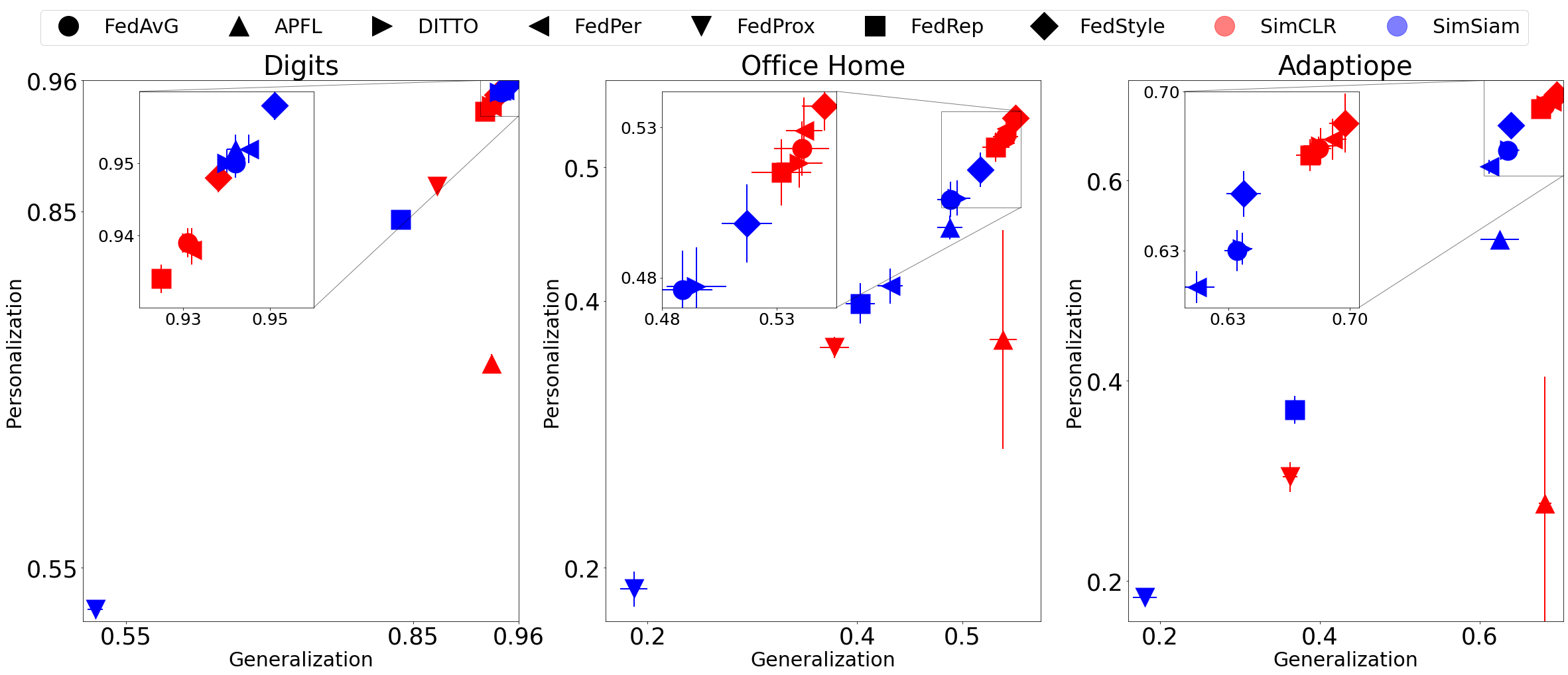}
    \caption{Generalization v.s. Personalization accuracy.}
    \label{fig:whole_single}
\end{figure}

\begin{figure}[htbp]
    \centering
    \includegraphics[scale=0.15]{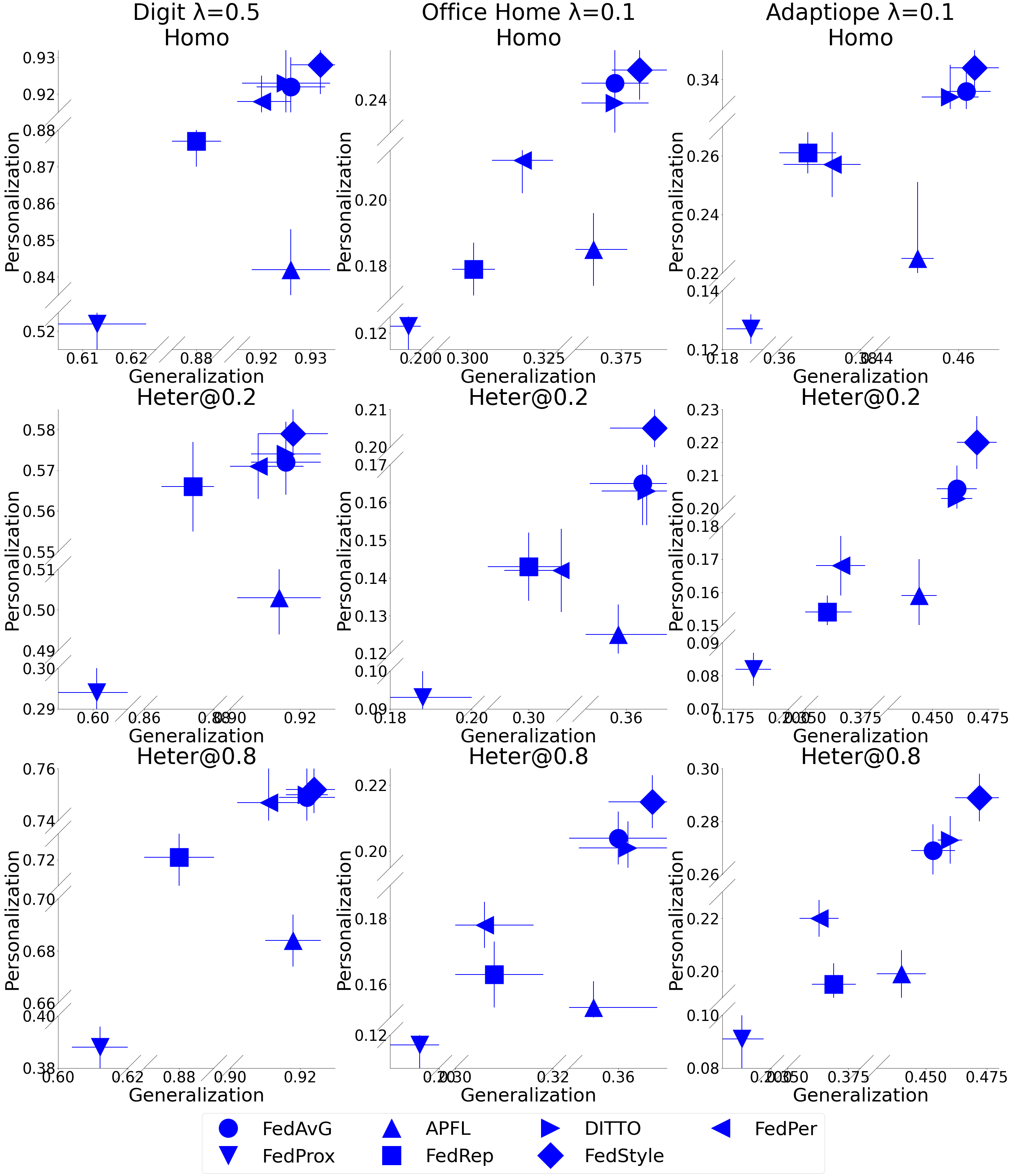}
    \caption{\textbf{SimSiam} \textit{Personalization} v.s. \textit{Generalization} accuracies for three distribution settings.}
    \label{fig:whole_simsiam}
\end{figure}